# ANALYZING AND IMPROVING STATISTICAL LANGUAGE MODELS FOR SPEECH RECOGNITION

by

Joerg Ueberla

B.Sc. Technische Universitaet Muenchen, 1988

M.Sc. Universite J. Fourier, Grenoble, France, 1990

A THESIS SUBMITTED IN PARTIAL FULFILLMENT
OF THE REQUIREMENTS FOR THE DEGREE OF
DOCTOR OF PHILOSOPHY
in the School
of
Computing Science

© Joerg Ueberla  1994

SIMON FRASER UNIVERSITY

May 1994

# APPROVAL

**Name:**              Joerg Ueberla

**Degree:**            Doctor of Philosophy

**Title of thesis:**   Analyzing and Improving Statistical Language Models for
                       Speech Recognition

**Examining Committee:**  Dr. Veronica Dahl
                          Chair

                          _______________________________________

                          Senior Supervisor: Dr. Fred Popowich

                          _______________________________________

                          Dr. Bob Hadley

                          _______________________________________

                          Dr. Tom Perry

                          _______________________________________

                          Dr. Binay Bhattacharya

                          _______________________________________

                          External Examiner: Dr. Renato De Mori

**Date Approved:**        _______________________________________

# Abstract

A speech recognizer is a device that translates speech into text. Many current speech recognizers contain two components, an acoustic model and a statistical language model. The acoustic model indicates how likely it is that a certain word corresponds to a part of the acoustic signal (e.g. the speech). The statistical language model indicates how likely it is that a certain word will be spoken next, given the words recognized so far. Even though the acoustic model might for example not be able to decide between the acoustically similar words "peach" and "teach", the statistical language model can indicate that the word "peach" is more likely if the previously recognized words are "He ate the".

Current speech recognizers perform well on constrained tasks, but the goal of continuous, speaker independent speech recognition in potentially noisy environments with a very large vocabulary has not been reached so far. How can statistical language models be improved so that more complex tasks can be tackled? This is the question addressed in this thesis.

Since the knowledge of the weaknesses of any theory often makes improving the theory easier, the central idea of this thesis is to analyze the weaknesses of existing statistical language models in order to subsequently improve them. To that end, we formally define a weakness of a statistical language model in terms of the logarithm of the total probability, $LTP$, a term closely related to the standard perplexity measure used to evaluate statistical language models. This definition is applicable to many probabilistic models, including almost all of the currently used statistical language models.

We apply our definition of a weakness to a frequently used statistical language model, called a bi-pos model. This results, for example, in a new modeling of unknown words which improves the performance of the model by 14% to 21%. Moreover, one of the identified weaknesses has prompted the development of our generalized $N$-pos language model, which is also outlined in this thesis. It can incorporate linguistic knowledge even if it extends over many words and this is not feasible in a traditional $N$-pos model. This leads to a discussion of what knowledge should be added to statistical language models in general and we give criteria for selecting potentially useful knowledge. These results show the usefulness of both our definition of a weakness and of performing an analysis of weaknesses of statistical language models in general.

To my parents

# Acknowledgements

I would like to express my deepest gratitude to my senior supervisor, Dr. F. Popowich. He was always there to help me deal with the multitude of problems I encountered. Without his great enthusiasm and continuous encouragement, this work would not have been possible. I would also like to thank the other members of my examining committee, Drs. B. Hadley, B. Bhattacharya and T. Perry for their valuable comments. Special thanks go to my external examiner, Dr. R. De Mori, for providing many valuable suggestions. Furthermore, I am indebted to E. Atwell, for enabling a very stimulating stay at the University of Leeds, and to Dr. R. Kuhn at CRIM, for his help and encouragement.

During my stay at SFU, I experienced many personally stimulating and enriching moments. I would like to thank all the people who contributed to them for sharing the mostly enjoyable life of a graduate student with me. In particular, I would like to sincerely thank Eli, Pp, Allanbb, Carl, Glenn, Katerina, Alicja, Pierre, Diana, Martin, Graham and Brigitte for making my stay at SFU such a memorable experience.

For the financial support that made my stay at SFU possible, I am indebted to Dr. F. Popowich, Dr. V. Dahl, the School of Computing Science and Simon Fraser University. Special thanks also go to the always friendly staff in Computing Science, in particular to Elma and Kersti.

# Contents

# Chapter 1

# Introduction

The study of speech recognition is of great importance because of the social and economic impact speech recognition will have on our society. We humans spend a large fraction of our lifetime speaking, listening, reading and writing. Already today, computers are involved in a large part of human communication, be it telephone switching, electronic mail, word processing, information retrieval or computer bulletin boards. At little extra cost, computers provide additional features improving the quality of these processes or making human labour more effective. The impact of computer technology on society has its good and bad sides (see [125], [129] and [38] for a discussion). But because humans will still want to communicate in the years to come and because computers are very likely to continue to become cheaper, it is very likely that an even bigger share of our spoken and written communication will be mediated by computers in the future. As computers continue to penetrate our society, being able to communicate with computers via speech is therefore of great social and economical importance.

Moreover, the study of speech recognition is interesting because of the intellectual challenge posed by a problem whose solution involves many different scientific disciplines. In the past, the field of speech recognition has benefited from sciences as diverse as biology, computer science, electrical engineering, linguistics, mathematics, philosophy, physics, psychology and statistics. Thus, the questions raised by speech recognition range from philosophical questions about the nature of mind to practical design and implementation issues. Motivated by the study of artificial intelligence, speech recognition can therefore serve as a testing ground, bringing many disciplines together in a concrete task, thus avoiding the dangers of a potentially introspective and subjective undertaking.

The fascinating interplay between different scientific disciplines and the great social and economic importance of speech recognition make it a very challenging, stimulating and exciting research field.

In the following, we [1] will briefly present the main difficulties of speech recognition and the approaches people have used to tackle them. We then give an overview of different methods of joining natural language processing and speech recognition and identify the topic of this thesis, statistical language models for speech recognition, as one of them. In the remainder of this thesis, we will use the term language model as a short hand for statistical language models for speech recognition. We conclude this chapter by giving an overview of our work. Most of the material of section 1.1, section 1.2 and of the previous three paragraphs is drawn from [100, p.10], [151, p.1-5] and [104].

## 1.1 The Difficulties of Speech Recognition

A speech recognizer is a device that translates speech intro written text and the problem of speech recognition has been studied actively since the 1950's. Enormous progress has been made, but many problems of speech recognition remain unsolved today. What makes speech recognition such a difficult task? Here are the main difficulties:

1) Each elementary sound, also called a phoneme, is modified according to its context, for example the immediately preceding and following phoneme. This is partly due to a property of our vocal apparatus called coarticulation: as one phoneme is pronounced, the pronunciation of the next phoneme is prepared by a movement of the vocal apparatus. Modification of a phoneme is also caused by the larger context such as its place in a sentence.

2) There is no separator, e.g. no silence, between words. This creates additional confusable words and phrases (e.g. "youth in Asia" and "euthanasia"). It also leads to more coarticulation, e.g. between words, and to poorer articulation (e.g. "did you" becomes "didja").

3) The variability of the speech signal for the same utterance is enormous. For example, there is intra-speaker variability due to the speaking mode (singing, shouting, with a cold, under stress, speaking rate, etc.), inter-speaker variability (sex, age etc.) and variability due to the environment (noise, lipsmacks etc.).

---

[1] I would like to clarify that even though I use the word "we" throughout this thesis, the work presented here is my own and thus qualifies for submission as a thesis.

4) Because of 1) and 2), it is necessary to process large sets of data in order to define what constitutes an elementary sound, despite the different contexts, speaking modes etc. For example, it is hard to decide that an "a" pronounced by a male adult is more similar to an "a" pronounced by a child in a different word and in a different environment than an "o" pronounced by the same male adult in the same environment [2].

5) Because the signal carries different types of information (sounds, syntactic structure, semantics, identity and mood of speaker etc.), a speech recognition system will have to differentiate between the information useful for its task and the remaining, irrelevant information.

6) There is no precise formalism that allows us to formalize the knowledge at all the different levels (e.g. acoustics, syntax, semantics etc.). However, recent trends suggest that a probabilistic framework might be used at many levels.

These six points are the main problems a speech recognizer has to face in general. However, concrete speech recognition tasks may vary greatly in the degree of difficulty they present. The following six dimensions can be used to classify a speech recognition task according to its difficulty:

1) Isolated (with pauses) or continuous speech. Continuous speech recognition is far more difficult because there are no word boundaries and because the variability of the signal is much greater.

2) Vocabulary size. As the vocabulary size increases (from small vocabularies of less than 500 words [3] to very large vocabularies of about 20,000 words), the task becomes more difficult because the number of acoustically confusable words increases and because more time is needed to evaluate all possible words.

---

[2]Nevertheless, even if it is the case that the "a" of the male adult is more similar to his "o" than to a child's "a", we do need to recognize his "o" as an "o", but the more different "a" of the child as an "a".

[3]It is important to know that "car" and "cars" are counted as two different words in a speech recognizer. Thus, a speech recognizer with 500 words has far fewer words in the usual sense than the number 500 might suggest.

3) Task and language constraints. The size of the vocabulary is not sufficient for determining the difficulty of a task because some words may not be allowed at a given time. For example, a task with 500 words, each of which can appear at any time, may be more difficult than a task with 700 words with strong restrictions on which words may follow other words.

4) Speaker dependence (for one speaker only) or speaker independence (for many speakers). A speaker independent task is much more difficult because of the additional inter-speaker variability.

5) Acoustic ambiguity. The acoustic confusability of words in the vocabulary also influences the difficulty of the task. For example, a task with 100 words that are highly confusable may be harder than a task with 200 words that are very dissimilar.

6) Environmental noise. A task in a very noisy environment is more difficult because the noise can lead to arbitrary distortions and modifications of the speech signal.

## 1.2   Different Approaches to Speech Recognition

Having seen the difficulties of speech recognition, how have researchers tried to tackle these problems? We can differentiate four different approaches – template-based, knowledge-based, stochastic and connectionist – and we will briefly present each one of them.

In the template-based approach, units of speech (e.g. words) are represented in the same form as the speech input itself. The input is compared to the templates using some distance metric thus identifying the best match. The problem of temporal variability is tackled by dynamic programming. For simple applications requiring minimal overhead, this approach has been quite successful.

In the knowledge based approach proposed in the 70's and early 80's, human knowledge is coded into expert systems. Rule-based systems had only limited success, but in more successful systems, the knowledge is integrated into a sound mathematical approach and this additional knowledge is found to improve the performance.

In the stochastic approach (e.g. using hidden Markov models or HMMs), a template pattern is represented at a higher level of abstraction by a reference *model* thus allowing some generalization. HMMs are based on a sound probabilistic framework that can model the uncertainty and variability inherent in speech recognition. Since HMMs simultaneously solve the segmentation and classification problem, they are particularly well suited for continuous speech recognition. Most successful large-vocabulary systems today use the stochastic approach.

The most recent development in speech recognition is the connectionist approach. This approach does not require some of the often incorrect assumptions underlying the stochastic approach. Even though no large scale, fully integrated connectionist system has been demonstrated, this approach holds considerable promise, especially in combination with the stochastic approach.

In the rest of this thesis, we will assume that the speech recognizer is built according to the widely used stochastic approach. Nevertheless, the ideas of language modeling presented in this thesis are also applicable to other approaches. A language model could, for example, be used to rescore hypotheses in a template-based or knowledge-based approach.

## 1.3 Incorporating Natural Language Constraints into Speech Recognition

Given any one of the approaches mentioned in the previous section, a speech recognizer can identify a set of candidate words, which are likely to correspond to a part of the signal. Suppose for example that we have recognized the words "He ate the" covering a certain part of the signal, how can we extend recognition by another word? Based on the acoustic properties of the words in the vocabulary, we can, for example, identify the words "peach" and "teach" as candidates, because they are very similar to the next stretch of the signal. However, we can not identify exactly which one of the candidate words was the one spoken. We can then use the linguistic context to identify the word that is more likely to appear next. In this example, it is clear that the word "peach" is far more likely than the word "teach". Using these constraints imposed by context is very important for speech recognition. This is for example pointed out in [117, p.33]: "We know that, in a real task, the importance of the language model is comparable to that of the acoustic module in determining the final performance". In general, this kind of reasoning involves constraints on the next word imposed by syntax, semantics or pragmatics. These constraints are part of the domains of natural language processing and linguistics. In the following, we will therefore look at different ways of incorporating natural language constraints into speech recognition.

There are many different ways of incorporating natural language constraints into a speech recognizer. Following roughly a classification suggested in [108], we will present four different approaches for combining a speech recognizer and a natural language processor. We will see how each approach acts in our example, e.g., how it chooses between "peach" and "teach" as possible continuations of the sentence fragment "He ate the".

1) Serial connection. In this approach, the natural language processor receives the most likely sentence from the speech recognizer and interprets it further. The advantage of this approach is that both systems have no additional computational burden from the "integration". The disadvantage is that there is almost no interaction between the two components. As a result, the natural language processor can not correct errors of the speech recognizer. This method is for example used in [132]. In our example sentence, the speech recognizer would have to choose between "teach" and "peach" independently of the natural language processor [4].

2) $N$-best sentence interface. The speech recognizer outputs the $N$ best scoring sentences (for $N = 1$, this is the serial interface). The natural language processor chooses the sentence that best satisfies the natural language constraints. The advantage is that this allows some interaction of the two, while adding only some additional computational burden. The size of $N$ determines the tradeoff between allowed interaction and additional burden. One disadvantage is that $N$ may be required to rise exponentially with the length of the input sentence. This approach is for example used in [132]. In our example, the speech recognizer could use "peach" in one of the $N$-best sentences and "teach" in another and this would allow the decision to be taken by the natural language processor.

3) Word lattice interface. the speech recognizer produces a graph with possible starting times, end times and recognition scores for any word of the vocabulary at any time. The natural language processor searches this graph for the most likely sentence that satisfies the natural language constraints. The advantage of this approach is the high degree of interaction between the two components. The disadvantage is, the additional computational burden for both systems. The speech recognizer has to keep track of and output many hypotheses, rather than concentrating on the best one. The natural language processor has to evaluate many more possible sentences. This method is used in [146] and [131]. For our example, the speech recognizer does not make the decision and both "teach" and "peach" will appear in the word lattice. The natural language processor will then make the final decision.

---

[4] However, the speech recognizer could use a very simple natural language processor to find the most likely sentence and this is further discussed after this classification.

4) Parallel connection. In this approach, the constraints provided by the natural language processor are used directly in the speech recognizer to reduce the search space. Within this category, we can further distinguish the approaches with respect to the complexity of the natural language processor. The natural language processor can be very complex, attempting to produce a parse and a semantic representation of the sentence. Or it can be very simplistic, attempting only to identify which words are likely to appear given the preceding word. We divide the whole spectrum into two classes, complex natural language processors and simplistic natural language processors. The line between the two classes can be drawn in many ways (e.g. whether the natural language processor attempts a parse or not), but for our purposes we don't need to specify exactly where we draw the line in order to continue.

a) Complex natural language processor. The advantage of this approach is that it allows considerable interaction between the two components. The constraints provided by the natural language processor are used directly during recognition to rule out some of the word candidates. One disadvantage is the amount of computation required to check the constraints provided by the natural language processor for all word candidates in the acoustic search. Another disadvantage is that very complex natural language processors can usually be built only for limited domains and this method is thus difficult to use for unrestricted speech. An example of use with a restricted domain can be found in [134]. Context free rules are derived automatically from sample sentences and then approximated by a probabilistic finite state machine. Another way of using this approach is presented in [108]. The natural language constraints are expressed in terms of a finite state machine, that is in turn used directly by the speech recognizer. However, since a typical natural language system will produce an enormous or even infinite number of states, only the parts that are currently searched by the speech recognizer are dynamically created by the natural language processor.

b) Simplistic natural language processor. The advantage of this approach is that

it allows considerable interaction between the two components. Moreover, since the natural language processor is very simplistic, it can efficiently score all the word candidates during the acoustic search. The disadvantage is that the natural language processor only captures very few constraints, thereby for example allowing ungrammatical sequences of words. We will see many examples of this approach later on.

In our example, the knowledge of the natural language processor and the speech recognizer are combined during the acoustic search to choose between the words "peach" and "teach".

For more information about this issue, the interested reader can refer to [52], [75], [108], [113], [116] and [144]. In this thesis, we will focus on the simplistic natural language processors of category 4b), also called language models, and we give four reasons for this choice. First, language models are used in many existing recognizers and this shows their great practical importance. Second, language models can be used even if the serial approaches of category 1) or 2) are chosen. In that case, the speech recognizer uses a language model to arrive at the $N$ most likely sentences (for example), which are then further processed by a more complex natural language processor. Third, if the task at hand does not require the *understanding* of the utterance, a parse may not be necessary and language models still provide a way of incorporating some natural language constraints into the speech recognizer. Fourth, more complex natural language processors (as in 4a) are mostly limited to one specific domain and they are thus of limited use for unconstrained speech recognition.

It is important to be aware of two different subtasks sometimes lumped together in the term speech recognition: speech understanding and speech recognition (proper). The goal of speech understanding is to understand spoken language and to react to it in a meaningful manner. Since this task is usually limited to a narrow domain, more complex natural language processors can be used. An example of a speech understanding task is that of understanding spoken queries to a database. Contrary to that, the task of speech recognition is only to transcribe speech into text. Because this task does not require understanding, it can and should deal with unrestricted text, not limited to a certain domain. Therefore, more simplistic natural language processors are commonly used. An example of the speech recognition task is the phonetic typewriter, a device that is able to output a printed version of a spoken conversation. Since our work focuses on the simplistic natural language processors, it is mostly relevant to speech recognition. But as pointed out above, even a speech understanding system with a complex natural language processor connected in a serial manner might use a simplistic language model during recognition.

Now that we have informally [5] presented the focus of our work, language models for speech recognition, we will give an overview of the rest of this thesis by giving a summary of each chapter.

## 1.4 Overview

### 1.4.1 Chapter 2: Language Modeling for Speech Recognition

In chapter two, we give an overview of the different components of a speech recognizer, describe their interaction, define the task of the component central to this thesis, the language model more formally, and review the most commonly used language models.

After introducing the different components of a speech recognizer, we define the task of a language model as the construction of one or more probability distributions over all the words of a vocabulary given the words that have been recognized so far. Intuitively, the speech recognizer uses this distribution to decide which words are likely to appear next, e.g. based on the probability distribution, it chooses the word "peach" over "teach" when the preceding sentence fragment is "He ate the". As this example illustrates, different words are more likely to appear in different contexts. Therefore, a language model usually has many different distributions, one for each context. In language modeling, context often means the two or three words preceding the word to predict. To show the usefulness of such a simple context, let us define the context as the immediately preceding word and consider what this entails for the following word. Only nouns and adjectives are likely to appear if the immediately preceding words is for example "the". Thus, even this simple definition of context can severely restrict the following word.

During speech recognition, the language model only has to choose a distribution according to the current context and to look up the probabilities of words in this distribution. The important task in constructing the language model is to determine, prior to recognition, the number of contexts it differentiates and to construct a probability distribution for each one of them.

---

[5]We will give a more formal description in the section defining the task of the model (section 2.3).

Language models are usually described in terms of frequency counts of their probability distributions, but for the purposes of this thesis, it is more appropriate to describe language models at the more abstract level of probability distributions and contexts. This is because we are more interested in the conceptually important aspects, e.g. the way in which a model defines context, rather than in the details of a particular technique of constructing probability distributions from frequency counts. Therefore, we only give a brief description of how probability distributions can be estimated from frequency data, often referred to as *training data*. The principle of this estimation is that of counting how often a certain event appears in a given context in the training data and of dividing this count by the number of overall occurrences of the context. For example, we can estimate the probability of having sunshine tomorrow given that it was raining today by dividing the number of times we had sunshine given that it was raining the previous day by the overall number of rainy days.

We then give a review of different language models in order to present the state of the art in language modeling and to set the stage for the remainder of this thesis. For each model, we give the definition of context it uses, the number of probabilities it needs to estimate and some of its advantages and disadvantages. In particular, we present the class based models or $N$-pos models, in which words are grouped into classes called parts of speech, which roughly resemble their grammatical function. In these class based models, the prediction of the next word is a two step process: first, the part of speech is predicted by one component, then the word given the part of speech is predicted by a second component.

### 1.4.2 Chapter 3: Analysing and Improving Language Models

In chapter three, which contains the central idea of this thesis, we propose to perform error analyses of language models in order to improve the models afterwards, define what we mean by "error or weakness of a language model" and present a method to identify the weaknesses of a given model.

We begin by noting that error analysis of existing theories about the world often leads to improvements of these theories. By analogy to this, we propose in this chapter to analyze errors of a language model in order to improve the model afterwards. But how can we define an error of a language model? The definition of an error should be related to the measure used to evaluate the performance of a language model. If definition and measure are not related, we may still identify and then remove an error, but this may not translate into an improvement in performance (since error and performance measure are not related).

Before defining an error, we therefore first introduce the standard measure used to evaluate language models, called perplexity.

Intuitively, the goal of a language model is to predict words in a given context. A good model should therefore assign a high probability to each word in a piece of text. Hence, the average probability assigned to words in a *testing text* — the geometric mean of probabilities to be more precise — is a good measure for the quality of a language model. Perplexity, the standard measure used to evaluate language models, is just the reciprocal of the geometric mean of probabilities. Besides explaining the perplexity intuitively, we also derive the perplexity using methods from information theory and further discuss its advantages and disadvantages.

Since the term *error* does not really apply to a language model, we prefer to use the term *weakness*. We now define a weakness [6] of a language model in terms of the logarithm of the total probability (LTP) of a sequence of words, a measure closely related to the perplexity. Moreover, for models with several components (for example the class based model we use), we develop the method of probability decomposition, which allows us to identify weaknesses of the different components separately.

The main idea of this chapter — applying error analysis to language models — applies to any probabilistic model whose performance is measured in terms of perplexity. We conclude this chapter by giving examples of models to which our method of error analysis applies.

### 1.4.3 Chapter 4: Analyzing and Improving a Bi-pos Language Model

In chapter four, we apply the central idea of this thesis, our technique of identifying weaknesses of a language model presented in the previous chapter, to a commonly used bi-pos language model and report the results.

In order to apply the technique of identifying weaknesses of a language model to a concrete model, we first choose a corpus (the Lancaster-Oslo-Bergen corpus), a model (the bi-pos model) and verify that the section of the corpus we use contains enough data to train our model. This work prompts an investigation into the issue of sample space, the set of all possible events considered by a model. We note that it is not meaningful to use the perplexity measure to compare language models that differ in their underlying sample spaces. Yet language models are usually compared with the perplexity measure, even though they sometimes differ in their underlying sample spaces, either due to different vocabularies or due to different ways of dealing with unknown words. We also discuss possible solutions to the problem of different sample spaces.

---

[6]We use the term weakness as a technical term and the intuitions, that still apply to our technical use of it are discussed on page 42.

We then apply our method of identifying weaknesses of a language model to our chosen bi-pos model and report three results of general interest. First, a very small number of classes [7] are identified as weaknesses of the model. We believe that these results are helpful for future efforts to improve the model, because we know on which classes we should concentrate our efforts. Second, unknown words are identified as weaknesses. This prompts the development of a new modeling of unknown words, which improves the performance by between 14% and 21%. Third, the word component of our bi-pos model is shown to be at least as important as the class component. This has interesting ramifications for using probabilistic context free grammars for language modeling, an approach that has recently received a lot of attention. Even though using probabilistic context free grammars may result in an improved prediction of a class (or part of speech), it is not likely to improve the prediction of the actual word given its class. We should therefore improve the word component of a class based model even if probabilistic context free grammars are being used. The additional insight gained from these results also show the usefulness of the central idea of this thesis, the identification and analysis of weaknesses of language models.

How can we go about improving the word component of a class based language model? This question leads us to develop our generalized $N$-pos model, which is shown to be a true generalization of the $N$-gram and the $N$-pos model. Moreover, it can incorporate any linguistic knowledge not restricted to the immediate context of the word to be predicted. This is exemplified by incorporating a very simple knowledge source into our generalized $N$-pos model. The results of this example also show that a considerable improvement (around 15%) is achieved in the prediction of words for which our generalized $N$-pos model actually differs from the original $N$-pos model. However, the overall improvement is negligible, because the cases in which our simple knowledge source can be used are very rare in our example. This leads to a general discussion of what knowledge we should add to a language model — an issue we address in the next chapter.

---

[7]For an intuitive explanation of the use of classes see the summary of chapter 2.

### 1.4.4 Chapter 5: Adding Linguistic Knowledge to Language Models

In chapter five, we motivate the addition of knowledge to language models, develop different criteria to identify useful knowledge, and present methods to combine knowledge in a language model.

We begin by pointing out three reasons for wanting to add knowledge to a language model. First, we would like to improve its performance. Second, if we apply current speech recognition technology to more complex tasks than the ones tackled today, the number of acoustically confusable hypotheses will increase, and we may well need a better language model in order to deal with the additional ambiguity. Third, adding knowledge is more satisfying than sticking to existing models on psychological grounds because humans seem to use knowledge to predict a word other than the knowledge used in current models, namely the immediately preceding two or three words. Hence, there is clearly a need for a language model which incorporates more linguistic knowledge.

Once we have decided to add knowledge to a language model, the following two questions come to mind. First, *what* knowledge should we add, and second, how can we *combine* different types of knowledge in a language model. We address both questions in turn.

Rather than trying to give a necessarily incomplete list of types of knowledge that we should add, we present four criteria that we think should be used to identify useful knowledge. First, the knowledge should restrict the number of possible words, otherwise it is not going to help in solving our task. Second, it should be applicable often enough to be of statistical significance. Third, it should be possible computationally to use this knowledge in real time speech recognition. Finally, we should be able to acquire and code this knowledge for use with unrestricted text.

We develop a classification of possibly useful knowledge and apply the criteria for identifying useful knowledge to one type of knowledge that promises to be useful for improving language models in general.

We then move on to the issue of combining different types of knowledge in a language model. We present three methods of combining knowledge and develop some of the advantages and disadvantages we see in each method. Following that, we conclude that it is very unlikely that we will have enough data to estimate distributions that depend on several knowledge sources directly, even with the availability of increasingly large corpora. Therefore, we think that methods that combine distributions from single knowledge sources in a meaningful fashion will be very useful and require further investigation. One such method shown to be very useful in recent work is the maximum entropy method, which will also be presented briefly in this chapter. It holds great promise for future work.

# Chapter 2

# Language Modeling for Speech Recognition

In the last chapter, we described intuitively how the topic of this thesis, language modeling for speech recognition, relates to speech recognition research in general. In this chapter, we will make this relationship more precise by introducing the different components of a speech recognition system (section 2.1), and, after introducing some notations in section 2.2, by defining the task of a language model more formally (section 2.3). Since a language model consists mainly of probability distributions, we present the method of constructing and smoothing probabilities distributions we use for our work (section 2.4). Given the notations introduced in section 2.2 and having seen the key issues of language modeling, we then give an overview of the state of the art in language modeling by reviewing existing language models (section 2.5).

## 2.1 The Components of a Speech Recognizer

A speech recognizer is a device that translates speech into written text. As input, it takes the acoustic signal recorded by a microphone. As output, it produces a string of words intended to correspond to the input. The mapping from acoustic signal to a string of words is a complex task and it involves several stages. To illustrate this mapping in a simplified way, we will present a set of stages that are very similar to the ones used in one of the first successful, large vocabulary, speaker independent, continuous speech recognizers, the SPHINX system ([95]).

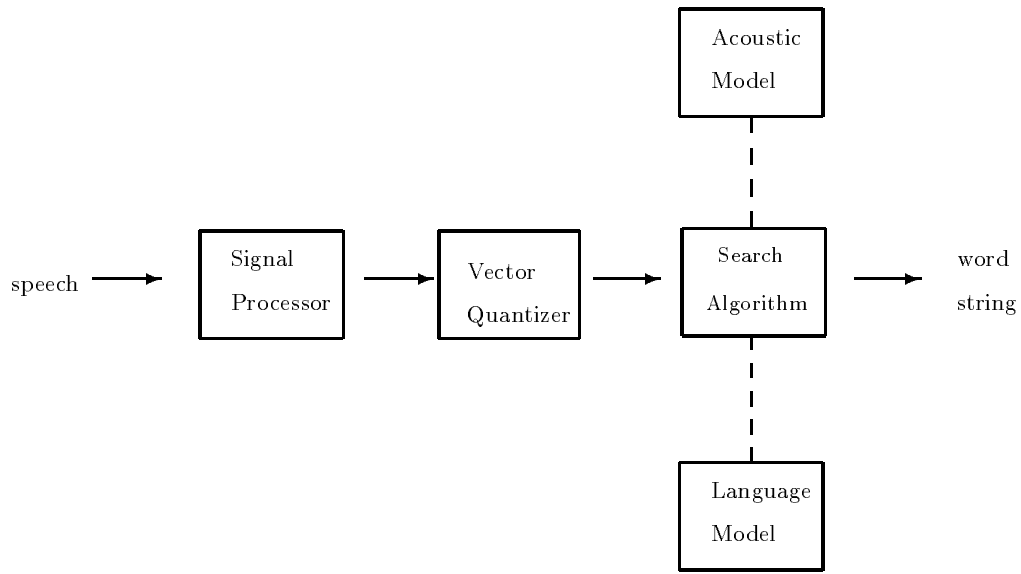

Figure 2.1: Components of the SPHINX system

As we can see from figure 2.1, the acoustic signal is first given to a signal processing component. This component performs several transformations, e.g. sampling the signal at fixed time intervals, reducing the noise etc. As output, it produces one 12 dimensional vector of floating point values per time interval. These vectors are the input to the next component, the vector quantizer [1], which compares each input vector to stored prototype vectors and outputs the index of the vector that is closest to the input vector. For a sequence of 12 dimensional input vectors, it thus produces a sequence of integers. This sequence constitutes the input to the next component, the search algorithm. It is a time synchronous algorithm (a Viterbi beam search, see [95]), that compares the likelihood of different sequences of units of speech. In order to calculate these likelihoods, the search algorithm uses the acoustic model and the language model. The acoustic model provides the algorithm with the likelihood that a unit of speech (phonemes and words in SPHINX) correspond to parts of the given sequence of integers. The language model provides the algorithm with the likelihood of occurrence of a unit of speech given the previously identified units. Based on these two components, the search algorithm identifies the most likely sequence of units of speech and this constitutes the recognized string of words.

---

[1] In recent years, the tendency has been to eliminate the vector quantizer by having continuous density hidden Markov Models.

As an example of the interplay of the acoustic model and the language model, suppose we have recognized the sequence "He ate the" so far. The acoustic model calculates the probabilities that the words "teach" corresponds to a prefix of the sequence of integers it receives as input. Similarly, it calculates the probabilities for the word "peach". The language model calculates probabilities for the words "teach" and "peach", given for example that the last recognized word is an article. Even if the acoustic model can not decide between "teach" and "peach" because they are very similar acoustically, the language model can have a clear preference for "peach" (because it is a noun and nouns are far more likely than verbs to follow articles). Based on the probabilities of the acoustic model and the language model together, we can then choose "peach" as the next recognized word.

This thesis is concerned with the language model component of a speech recognizer. Similar to the language model in SPHINX mentioned above, we can describe a language model in general as follows. A speech recognizer has recognized a sequence of words in the past and is now trying to extend this sequence by another word. Based on the acoustic signal it receives, it can identify a set of candidate words whose acoustic signal is very similar. However, based on the acoustic signal alone, it can't identify precisely which of these candidate words is the one that was spoken. It therefore uses a language model to pick the word that is more likely to appear in this context. This likelihood of appearance is formalized in terms of probabilities: each word of the vocabulary has a probability of appearing next in a given context. We can thus describe the task of the language modeling for speech recognition intuitively as the construction of a probability distribution over all the words of the vocabulary. As an example, consider the words "He ate the" and two candidate words "peach" and "teach". Humans can easily identify the word "peach" as the likely continuation of the sentence fragment. The probability distribution constructed by a language model should (ideally) give a higher probability to "peach" than to "teach", allowing it to make the correct choice.

## 2.2 Frequently Used Notations

In this section we will introduce some notations that we will use for the remainder of the thesis. Given these notations, we can then define the task of a language model more formally in section 2.3.

- $V = \{w_1, ..., w_m\}$ will denote the vocabulary of a speech recognizer

- $l, 1 \leq l \leq m$ will denote an index ranging over this vocabulary

- $W = w[1], ..., w[n]$ will denote a string of words of $V$. In other words: $\forall i : \exists l(i), 1 \leq l(i) \leq m : w[i] = w_{l(i)}$

- $i, 1 \leq i \leq n$ will denote an index ranging over the string of words

- $w[i1 : i2]$: if $i1 \leq i2$ it will be a short form for $w[i1], ...w[i2]$ , else it will denote the empty string

- $p(x|y)$ will denote the conditional probability of $x$ given $y$

- let $V^+$ denote one or more symbols of $V$; $argmax_W f(W)$ will denote the $W \in V^+$ for which $f(W)$ has the maximum value [2]

- $p(w[i] = w_l|c)$ will denote the probability of the $i^{th}$ word in the sequence being the word $w_l$ given the context $c$

- $p(w[i]|c)$ will denote the probability distribution over the vocabulary given the context $c$. In order for $p(w[i]|c)$ to be a probability distribution, it has to satisfy the following two constraints. Each probability must be between 0 and 1, and the sum of the probabilities must be 1:

  1. $\forall l : 0 \leq p(w[i] = w_l|c) \leq 1$
  2. $\sum_l p(w[i] = w_l|c) = 1$

- $A$ will denote the acoustic data given to the recognizer

- $G = \{g_1, ..., g_t\}$ is a set of classes or parts of speech (we will introduce the notion of word classes in section 2.5.3)

- $j, 1 \leq j \leq t$ will denote an index ranging over the set of classes

- $g(w)$ will denote the class of a word $w$

- $g(w[i1 : i2]), i1 \leq i2$ is a short hand for $g(w[i1]), ..., g(w[i2])$

---

[2]If there are several $W$'s for which $f(W)$ has the maximal value, $argmax_W f(W)$ will denote one of them that was picked randomly.

## 2.3   The Task of a Language Model

Given the notations introduced in the previous section, we can now derive the task of a language model more formally.

A good speech recognizer should choose the most likely string $W^*$, given the acoustic data $A$. This is expressed by the following formula

$$W^* = argmax_W \, p(W|A). \tag{2.1}$$

Based on Bayes' formula (see for example [41, p.150]), we can rewrite the probability from the right hand side of 2.1 according to the following equation:

$$p(W|A) = \frac{p(W) * p(A|W)}{p(A)}. \tag{2.2}$$

$p(W)$ is the probability that the word sequence $W$ is spoken, $p(A|W)$ is the probability that the acoustic signal $A$ is observed when $W$ is spoken and $p(A)$ is the probability of observing the acoustic signal $A$. Based on this formula, we can rewrite the maximization of equation 2.1 as

$$W^* = argmax_W \frac{p(W) * p(A|W)}{p(A)}. \tag{2.3}$$

Since $p(A)$ is the same for all $W$, the factor $p(A)$ does not influence the choice of $W$ and maximizing equation 2.3 is equivalent to maximizing

$$W^* = argmax_W \, p(W) * p(A|W). \tag{2.4}$$

The component of the speech recognizer that calculates $p(A|W)$ is called the acoustic model. The component calculating $p(W)$ is the language model.

Why is maximizing equation 2.4 easier than maximizing equation 2.1? Or, in other words, why did we use Bayes formula to rewrite equation 2.1? For equation 2.1, we would need to build a model for all possible acoustic signals $A$. For equation 2.4, we need a model for every possible word sequence $W$. Since $A$ is a continuous signal and $W$ is discrete, the latter is easier.

How can we calculate $p(W)$ for a given string $W$? Formally, we can decompose the probability of a sequence of words $p(W)$ as the product of probabilities of each word $w[i]$ given the preceding words $w[1:i-1]$:

$$p(W) = \prod_{i=1}^{i=n} p(w[i] = w_{l(i)} | w[1:i-1]) \tag{2.5}$$

This decomposition is appropriate for speech recognition for the following reason. It allows us to evaluate the probability of a prefix $w[1:k], 1 \leq k \leq n$ of $W$ as the product of probabilities of the $k$ words it contains:

$$p(w[1:k]) = \prod_{i=1}^{i=k} p(w[i] = w_{l(i)}|w[1:i-1]) \tag{2.6}$$

This is very useful when we try to perform the maximization in equation 2.4. Rather than having to construct a $W$ covering the entire signal before we can evaluate it with the language model, we can now evaluate partial strings covering only parts of the signal as they are constructed. We can thus prune the search space by never expanding or evaluating unlikely partial strings [3]. Using equation 2.6, we can now precisely define the task of a language model.

**Definition 1** *Given a set of contexts $C = \{c_1, ..., c_p\}$, the task of a language model is to provide a probability distribution $p(w[i]|c_k)$ for each context $c_k, 1 \leq k \leq p$ and a way of choosing a context given the words recognized so far.*

During recognition, all the language model has to do is to determine which is the current context and to look up the probabilities of words in the distribution for this context. This is fairly straight forward once the model has been constructed. The important issue, however, is to construct the language model prior to recognition. This requires the definition of the set of contexts and the estimation of a probability distribution for each context. These contexts can capture any information about the words spoken so far. However, the language model must be able to extract this information efficiently during recognition. An example of such information is whether the subject of the current sentence is animate or not. The language model must be able to decide efficiently whether the subject of the current sentence hypothesis is animate or not in order to determine the current context and therefore the distribution it is going to use.

As an example of a language model, consider a very simplistic model that constructs only one distribution, independent of context. The word "teach" for example will therefore be expected with the same probability, whether the previous words were "He likes to" or "He ate the". This is clearly not a very good model since the constraints on the following words vary significantly with context. A better language model would therefore have several distributions, one for each context it treats separately.

---

[3]Pruning has to be done with care because it can lead to the pruning of unlikely partial strings, that would become more likely given the later, yet unseen parts of the signal.

Here we can see two conflicting interests that influence the construction of a language model (see [62]). On the one hand, the more different contexts a language model can differentiate, the more distributions it has, and the better it can model a language. On the other hand, each distribution needs to be estimated from training data (see section 2.4). The more distributions it has, the more data it needs. In other words, given a fixed amount of training data, the more distributions a language model has, the less accurate the estimates will be. Trying to balance these conflicting goals is one of the difficulties of constructing language models and we will encounter this problem again in section 2.5.

## 2.4 Estimation and Smoothing of Probability Distributions from Frequency Data

Because we will be using many probability distributions throughout the rest of this thesis, we need to examine the estimation and smoothing [4] of probability distributions based on frequency data. All of the probability distributions are produced with similar techniques and once we have dealt with these issues here, we won't need to address them separately for each probability distribution we use. This way, we can describe different language models on the more abstract level of probability distributions, rather than having to describe them on the level of frequency data, requiring many lengthy formulas.

### 2.4.1 Estimation of Probability Distributions from Frequency Data

How can we estimate a probability distribution? As a simple example, consider the tossing of a coin. We would like to know with what probability it comes up head or tails and this will be its probability distribution. Intuitively, we can estimate this distribution in the following manner. Throw the coin $N$ times, count the number of times it comes up heads and tails, and denote these numbers with $H$ and $T$ respectively. We can then estimate the probability of the coin coming up heads or tails as $\frac{H}{N}$ or $\frac{T}{N}$ respectively.

How can we extend this to the more general issue of estimating probability distributions of events in certain contexts? A context, in the case of the coin tossing, could be the outcome of the previous toss or the fact, that the coin is lying head or tails up in our hand when we throw it. Let $E = \{e_1, ..., e_p\}$ denote a set of events and let $C = \{c_1, ..., c_r\}$ denote a set of contexts. As in section 2.2, we will denote the probability that event $e_l, 1 \leq l \leq p$ occurs in context $c_k, 1 \leq k \leq r$ as $p(E = e_l | C = c_k)$. Furthermore, we will denote the probability distribution over all events in a given context $c_k$ as $p(E | C = c_k)$. Our goal is to estimate a probability distribution $p(E | C = c_k)$ for all $c_k, 1 \leq k \leq r$.

---

[4]Smoothing attempts to make the probabilities depend less on the particularities of the training data and to avoid zero probabilities for events that were never seen. We will see smoothing in more detail later on.

If we follow the example of the coin, we simply make a large number of trials $N$ and count the number of times each event occurs in each context, denoted by $O(E = e_l | C = c_k)$. The occurrence counts $O(E = e_l | C = c_k)$ are often referred to as the *training data*. We can then calculate the number of times context $c_k$ occurs, denoted as $O(C = c_k)$, as the sum of the number of times each event occurs in that context:

$$O(C = c_k) = \sum_{l=1}^{l=p} O(E = e_l | C = c_k). \qquad (2.7)$$

The sum of the number of occurrences of each context will be the total number of trials:

$$N = \sum_{k=1}^{k=r} O(C = c_k). \qquad (2.8)$$

As in the example of the coin, we can then get an estimate of $p(E = e_l | C = c_k)$ by dividing the number of times event $e_l$ occurred in context $c_k$ by the total number of times context $c_k$ occurred:

$$p(E = e_l | C = c_k) = \frac{O(E = e_l | C = c_k)}{O(C = c_k)} = \frac{O(E = e_l | C = c_k)}{\sum_{l=1}^{l=p} O(E = e_l | C = c_k)}. \qquad (2.9)$$

From all possible values we can estimate for $p(E = e_l | C = c_k)$, the value estimated above is the one that has the highest likelihood of producing the observed data [5]. This method of estimation is therefore called maximum likelihood estimation (MLE). As pointed out in [140], the principle of maximum likelihood estimation was first proposed by Sir R. A. Fisher in 1926 (see for example [37]). Maximum likelihood estimation is a very simple method that can be used for a wide range of problems. Even though more sophisticated methods are available (see for example [20]), we use the maximum likelihood estimation for reasons of simplicity for our work.

---

[5]Strictly speaking, it is the value that has the highest likelihood of producing the observed data given some additional assumptions about the distribution of probability values.

From now on, we will denote the quotient of the occurrences in equation 2.9 by the frequency $f(E = e_l | C = c_k)$:

$$f(E = e_l | C = c_k) = \frac{O(E = e_l | C = c_k)}{O(C = c_k)} = \frac{O(E = e_l | C = c_k)}{\sum_{l=1}^{l=p} O(E = e_l | C = c_k)}. \quad (2.10)$$

### 2.4.2 Smoothing of Probability Distributions

In the case of a language model, it is important to avoid zero probabilities for events that never occurred in the data. The reason for this is that the speech recognizer should correctly decode what the user said. Since we cannot prevent the user from saying nonsensical or ungrammatical words at some point in the sentence, the language model should not give a zero probability to any of the words at any time. If it did, such a word could not be recognized even if it was the one said by the user. If we use the maximum likelihood estimation for events that never occurred, they will receive a probability estimate of zero because their occurrence count is zero. In order to avoid this, the probability estimates have to be smoothed. This results in giving a small probability to unseen events and in reducing the probability of other events.

The techniques often used for smoothing are the addition of a small constant probability (see for example [85]), deleted interpolation ([66]), backing off ([73]), different discounting methods ([110]), Good-Turing formula ([48]) or enhanced Good-Turing formula ([20]). In choosing one of these methods for our work, we had two criteria. First, since smoothing is not an issue of particular interest to our work, we would like the method to be fairly simple. Second, in order to ensure that the method is acceptable to other researchers, the method should be used by other researchers in similar language models. Adding a small constant probability satisfies both criteria and this is the method we chose. We will present it in the following.

Suppose we want to estimate $p(E = e_l | C = c_k)$ based on the frequency counts $f(E = e_l | C = c_k)$. In a first approach, we can simply use

$$p(E = e_l | C = c_k) = f(E = e_l | C = c_k). \quad (2.11)$$

However, some of the events in $E$ may never have been observed in a certain context, for example event $e_p$ in context $c_k$. Thus, we would obtain $p(E = e_p | C = c_k) = 0$. As pointed out in the preceding paragraph, we should avoid zero probabilities in language models. A simple way to avoid zero probabilities is to add a small constant value $v_1$ to all probabilities:

$$p(E = e_l | C = c_k) = f(E = e_l | C = c_k) + v_1. \tag{2.12}$$

Adding $v_1$ indeed avoids zero probabilities, but the sum of probabilities of all possible events is now bigger than 1 and the resulting distribution is not a probability distribution any more (see section 2.2):

$$\sum_{e_l \in E} (f(E = e_l | C = c_k) + v_1) = \sum_{e_l \in E} f(E = e_l | C = c_k) + |E| * v_1 = 1 + |E| * v_1. \tag{2.13}$$

In order to compensate for the extra probability mass of $|E| * v_1$, we will simply multiply the frequencies $f(E = e_l | C = c_k)$ with the constant value $v_2 = 1 - |E| * v_1$:

$$p(E = e_l | C = c_k) = v_2 * f(E = e_l | C = c_k) + v_1. \tag{2.14}$$

We can verify that the sum of the probabilities of all words now adds up to 1:

$$\sum_{e_l \in E} (v_2 * f(E = e_l | C = c_k) + v_1) = v_2 * \sum_{e_l \in E} f(E = e_l | C = c_k) + |E| * v_1 \tag{2.15}$$

$$= v_2 * 1 + |E| * v_1 = 1 - |E| * v_1 + |E| * v_1 = 1. \tag{2.16}$$

Thus, the smoothed estimate in equation 2.14 constitutes a true probability distribution and avoids zero probabilities.

### 2.4.3 Assumptions about Probability Distributions in our Work

In the rest of the thesis, we will often refer to probability distributions and all of these can be estimated using the methods presented above. For example, we will denote with $p(w[i] = w_l | w[i - 1])$ the probability that the $i^{th}$ word of the sequence is $w_l$, given that the previous word $w[i - 1]$ is $w_{l(i-1)}$ (see section 2.2). It is understood that this probability is estimated approximately as presented above. Summing up, this estimation roughly works as follows: count the number of times $w_{l(i-1)}$ occurs in the training data, count the number of times it is followed by the word $w_l$, and then estimate the probability as the quotient of these two numbers. Moreover, to ensure that our results are easily reproducible, we will give the complete formulas, including the smoothing, for the probability distributions we actually implemented.

## 2.5    Review of Existing Language Models

Having introduced the key issues of language modeling, we will now review many commonly used language models. However, rather than giving the often complicated formula in terms of frequency counts, we will describe each model on a more abstract level in terms of its probability distributions. This way, we don't have to consider estimation and smoothing issues, but can focus on the following conceptually important issues:

1) We saw in section 2.3 that the task of a language model is to provide a probability distribution for a set of suitably defined contexts. How is the context defined or, in other words, on what does the probability distribution depend in each model? This is a crucial point of each model because it shows which linguistic regularities (e.g. those that involve only the preceding two words) it can capture. We will provide an intuitive description of the contexts as well as the formula each model uses to instantiate the general $p(w[i] = w_l|c)$ with a specific context $c$.

2) How many probabilities have to be estimated in each model? This is important because it determines the amount of data needed to train the model. This in turn determines the situations and tasks in which each model can be used.

### 2.5.1    Context Independent Models

Context independent models have only one probability distribution. This distribution is used to assign probabilities to words, independent of the current context.

The most simplistic, context independent model is the model that treats all words as being equiprobable. Given the vocabulary $V$, this results in the following formula:

$$p(w[i] = w_l|c) = \frac{1}{|V|} \tag{2.17}$$

This model does not have any probabilities to estimate and therefore does not need any training data. Even though this model satisfies the requirements of a language model, it is of no use to a speech recognizer because all words receive the same probability. It will therefore have no influence on the ranking of the words.

A more sensible way to construct a context independent model is to estimate the probability of each word according to its frequency, but independent of context. This leads to the following model:

$$p(w[i] = w_l|c) = p(w[i] = w_l). \tag{2.18}$$

This model also has only one distribution, but it has $|V|$ probabilities to estimate. Even though this model is very simple, it is actually being used in a commercial speech recognizer ([56]). Because it is a special case ($N = 1$) of the $N$-gram model (see the next section), it is sometimes referred to as the uni-gram model.

The advantage of the uni-gram model is that it requires only very little training data. Its disadvantage is that the probability of a word will always be the same, independent of the context.

### 2.5.2  $N$-gram Models

The previous models had only one distribution, independent of context. From now on, we will make the probability distributions depend on context. The different models we will see will mostly differ in the kind of context they consider.

Before looking at the general $N$-gram model, we will consider the special case of the bi-gram model, where $N = 2$. When we look at a fragment of a sentence, e.g. "He ate the", it is quite clear that certain words are not valid continuations of the sentence. For example, if the last word in our sentence fragment, namely 'the', is being followed by a verb, it will not lead to a grammatical sentence [6]. It is therefore intuitively appealing to make the probability distribution depend on the previous word. The context is therefore simply the preceding word. Even though this captures only a very small amount of context, it does capture the restrictions in the above example. Moreover, there is considerable empirical evidence from corpus linguistics that the immediate context of many words is very predictable (see the discussion in section 5.2.3). This is especially true for fixed word order languages like English (see [23, p.32]), where the local constraints are quite powerful. Making the probability distribution depend on the previous word leads to the following formula:

$$p(w[i] = w_l | c) = p(w[i] = w_l | w[i - 1]).\qquad(2.19)$$

In the more general form of the model, the so called $N$-gram model, the probability of the $i^{th}$ word of an input sentence is made dependent on the preceding $N - 1$ words. The context is therefore defined by the preceding $N - 1$ words:

$$p(w[i] = w_l | c) = p(w[i] = w_l | w[i - N + 1 : i - 1]).\qquad(2.20)$$

---

[6] Even if a word can be a noun and a verb, each *occurrence* of a word has only one grammatical function. Thus, if the article is followed by an occurrence of a verb, it will not lead to a grammatical sentence.

For each $N-1$ tuple of words of the vocabulary, the $N$-gram model has a separate probability distribution and, at a given point in a sentence, the distribution it chooses is determined by the previous $N-1$ words. There are $|V|^{N-1}$ different $N-1$ tuples and this is the number of distributions of the $N$-gram model. For each distribution, we have to estimate $|V|$ probabilities. This gives a total of $|V|^N$ probabilities to estimate. For a vocabulary size of 10,000 words, the number of probabilities that need to be estimated increases dramatically (exponentially) with $N$. For example, for $N=3$, we have $10^{12}$ probabilities. Therefore, in practice, $N$ is usually taken to be two (see for example [79], [34]) or three ([5], [13], [63], [27], [67], [65]).

The advantage of the $N$-gram model is that it captures all the information provided by the preceding $N-1$ words. Judging from its success, this is quite an important source of information, especially for fixed word order languages like English. Its disadvantage is the enormous amount of training data needed to train all the probabilities. For example, in [13], several hundred millions of words are used for training.

As pointed out in [63], fifteen years after the first use of a tri-gram model in large vocabulary speech recognition ([5]), the tri-gram model is still considered one of the best performing models and it is used as a component in many other models.

### 2.5.3  $N$-pos Models

The major problem with the $N$-gram models is the amount of data required for training. Moreover, one can argue that some of the local constraints depend less on the identity of the previous words than on their grammatical function. This leads to the idea of grouping words together in classes and making the probabilities depend on these classes. Traditionally, these classes are called parts of speech (pos) in linguistics which explains the name of the $N$-pos models. Within the class of $N$-pos models, there are many different variants. In the following, after starting with a very basic model, we will present two modifications that lead to the model used in our implementation.

Let $G = \{g_1, ..., g_j, ...g_t\}$ denote the set of classes, let $g(w)$ denote the class of a given word $w$ and let $g(w[i1 : i2]), 1 \leq i1 \leq i2 \leq n$ be a short form for $g(w[i1]), g(w[i1+1]), ..., g(w[i2])$ [7]. In the $N$-pos model, the probabilities depend on the *classes* of the previous $N-1$ words.

---

[7] All of these notations are also mentioned in the section on notations (2.2).

Therefore, the context is defined by the preceding $N - 1$ classes:

$$p(w[i] = w_l|c) = p(w[i] = w_l|g(w[i - N + 1 : i - 1])).$$ (2.21)

This model has $|G|^{N-1}$ distributions and requires the estimation of $|G|^{N-1} * |V|$ probabilities. For common values of $|G| = 200, |V| = 10,000$ and $N = 3$, the $N$-pos (e.g. tri-pos) model has $8 * 10^{10}$ probabilities. This is a significant reduction with respect to the tri-gram model.

Furthermore, one can argue that the class of the previous word mostly restricts the *class* of the next word, but not its identity. Hence, we can derive probabilities in a two-step process. First, we predict the class of the next word based on the classes of the previous $N - 1$ words. Then, we predict the actual word given its class, but independent of preceding classes. This leads to the following formula

$$p(w[i] = w_l|c) = p(g(w[i])|g(w[i - N + 1 : i - 1])) * p(w[i] = w_l|g(w[i])).$$ (2.22)

This model has the same definition of a context, but it only has $|G|^{N-1} * |G| + |G| * |V|$ free parameters. For the same values of $|G|$ and $|V|$, this corresponds to $10^8$ probabilities, a further reduction compared to the previous formula.

The above model, used for example in [15], [29], [30], [60], [78], [112], [156] and [155], requires disjoint classes. However, one word can belong to several classes. For example, the word 'light' can be a noun, verb or adjective. Hence, the probability of seeing the word 'light' is the probability of seeing it as a noun plus the probability of seeing it as a verb plus the probability of seeing it as an adjective. This leads to the following formula, where the probabilities are summed over all possible classes in $G$:

$$p(w[i] = w_l|c) = \sum_{g_j \in G} p(g(w[i]) = g_j|g(w[i-N+1:i-1])) * p(w[i] = w_l|g(w[i]) = g_j).$$ (2.23)

This is equivalent to summing over all classes $w[i]$ can actually belong to since the second term in the formula will be zero for classes that do not contain $w[i]$.

In order to relate the $N$-pos model to the $N$-gram model, it is quite revealing to look at the extreme cases of $N$-pos models, e.g. at a model with only one class and at a model with one class per word (see Figure 2.2). If a $N$-pos model has only one class, then knowing the classes of the $N - 1$ previous words does not contain any information about the context because the last $N - 1$ words always belong to the same single class. Similarly, the class of the word to predict does not contain any information about the word to predict or the context, because all words belong to this class. Thus, since both factors in equation 2.23 are independent of the context, the prediction of the next word will be independent of the context. We therefore obtain a context independent model with only one distribution. This model is identical to the uni-gram model of section 2.5.1. At the other extreme, we have a model with a separate class per word. If a model has one class per word, then predicting the word given its class becomes trivial, because each class contains only one word. In this case, defining the context in terms of the classes of the previous $N - 1$ words actually means defining the context in terms of the identity of the previous $N - 1$ words. Thus, the second factor in equation 2.23 will always be equal to one and the first factor will be the prediction of the next word given the previous $N - 1$ words. In other words, we obtain the $N$-gram model from section 2.5.2. From these observations, we can see that the $N$-pos model is somewhere between the unigram and the $N$-gram model, depending on the number of classes it uses.

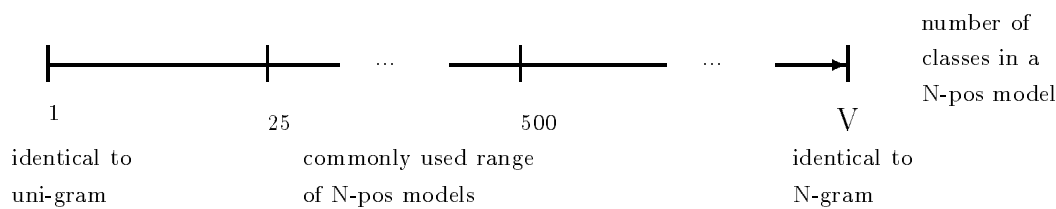

Figure 2.2: The relationship between $N$-gram and $N$-pos model

This model, used for example in [85], [32], [62] and [147], has the same number of distributions and parameters as the previous model – the difference is that the classes it uses have to be disjoint.

The advantage of the $N$-pos model is that it requires far less training data than the $N$-gram model, while still considering the class information of the previous $N - 1$ words. Its disadvantage is that its distributions depend on classes, and not on particular words. As an example, suppose that the class ARTICLE contains the singular article "a" as well as other articles like "the". In the case of a bi-pos model, we will have one distribution, given that the last word was an article. However, if we knew that the last word was the article "a", the distribution would be significantly different since it would not contain plural nouns. We will come back to this in section 4.4.2, page 84. In general, the performance of the $N$-pos model is not as good as an $N$-gram model trained on sufficient data, but it is better than an $N$-gram model trained with insufficient data. Here, we can see again the conflicting interests in constructing a language model that we saw on page 21.

### 2.5.4 Decision Tree Based Models

In all previous models, the number of distributions is fixed independently of the particularities of the training data. For example, in the tri-pos model, there is a separate distribution for each pair of preceding classes. This is done for all pairs, even though some of the resulting distributions may be very similar. As an example, the distribution in the contexts [verb, article] and [preposition, article] might be very similar to each other and even to the distribution in the context [article]. This is a serious disadvantage because it leads to the construction of very similar distributions, which do not result in improved performance.

The statistical technique of decision trees can avoid this problem. It has been used recently for different tasks in statistical natural language processing ([7], [9], [150], [148], [87], [10]). A good introduction to a specific method for constructing decision trees, called CART, is given in [11]. Other recent algorithms are presented in [45] and [16]. More on decision trees in general can be found in [46]. In the following, we will briefly outline the basic idea and its application to language modeling.

A decision tree contains the probability distributions of a language model and a method of identifying the distribution that should be used in the current context. In such a tree, each leaf contains exactly one distribution and each internal node contains exactly one question about the context. The following method is used to find the distribution (or the leaf) that should be used in the current context. Starting at the root node, we look at the question contained in the current node. Based on the answer to this question, we move to one of the children, making it the current node. This process continues until we arrive at a leaf. We then use the probability distribution associated with this leaf.

To construct a decision tree, we start with only one node, the root, containing only one probability distribution. Given a predefined set of possible questions about the context, we choose one that maximizes some criterion, e.g. performance of the tree on test data. This question is then placed at the root node, the children are created, and separate distributions are estimated for each child. This process continues recursively until some stopping criterion, e.g. none of the questions lead to improvements, is met. At each leaf, the distribution is estimated in the following manner. Our training data consists of a set of data points, e.g.

the occurrence of some word in a given context. Each of these data points starts out at the root node and then, by answering the questions at each internal node it encounters, ends up at some leaf node. Once all data points have reached the leaves, we estimate a probability distribution for each leaf based on the frequencies of events in the data it contains.

As an example, suppose we have a set of training data containing the identity of a word $w[i]$, given the previous word $w[i-1]$. One such data point could be (w[i-1],w[i])=(the, weather). Suppose further, that the set of possible questions we can ask is "Is the part of speech of the previous word $g_j$?", where $g_j$ can be any part of speech. Our goal now is to construct a language model based on this training data and on this set of questions. We will start out with one node containing only one distribution. As for every other leaf, this distribution is estimated from the frequencies of events in its data. In this case, the distribution will correspond to the uni-gram (see section 2.5.1), giving the relative frequency of each word. For each question we can ask, we construct the children, their separate distributions and measure the performance of the resulting language model on testing data. We then choose the question that leads to the best performing model and actually put this question in the root node. We then create two children, send each data point to the right or left child, depending on its answer to the question we just chose, and construct separate distributions for the children based on the frequency of events in their data. We can then apply this process recursively to each child until some stopping criterion is met, e.g. none of the questions leads to a further improvement. This terminates the process and the resulting decision tree is our language model.

All the models we have seen so far can be represented in terms of a decision tree. Decision trees are therefore more general language models. Their advantage is that the number of distributions is not fixed in advance, but it is determined by the training data. The number of distributions is therefore in general more appropriate than if one of the previous models is used. Its disadvantage is that the task of constructing the tree is computationally very expensive. And even though the resulting model frequently has less distributions to store, the improvement in performance is often relatively small (see [7], [150]).

## 2.5.5 Dynamic, Adaptive and Cache-Based Models

In all of the models we have seen so far, the probability distributions are estimated from the training data and do not change further when the model is used on different texts or on different portions of a text. For this reason, they are called static language models ([67]).

However, intuitively, it is very clear that some words are very "bursty" of nature: they do not occur in a large portion of the text, but then occur frequently in one small section. There is also strong empirical evidence to support this intuition. In [68] and [69], it is shown for three different corpora that frequencies of words vary greatly between different types of text. Language models that try to capture this short term fluctuation of the overall frequencies are called dynamic, adaptive or cache-based language models.

So far, there has been very little work on dynamic language models in the literature. The idea was first proposed by R. Kuhn in [84], then developed in [85], [86] and [88]. It was further tested in [67], [79] and [27]. Since all these approaches have a considerable degree of similarity, we will only present one of them in more detail.

In [67], the occurrences of the $N$ most recent words $w[i - N : i - 1]$ (e.g. $N = 1000$) are considered as separate training data. Based on this data, separate unigram, bi-gram and tri-gram frequency distributions are estimated. They are combined using one of the smoothing methods to obtain a single dynamic tri-gram estimate denoted by $p_{dyn}(w[i]|w[i-2:i-1])$. This distribution assigns a non-zero probability to all the words that occurred within the last $N$ words. To avoid zero probabilities of the remaining words, the model is combined with the static tri-gram model $p_{sta}(w[i]|w[i-1:i-1])$ to give the combined model $p_{com}(w[i]|w[i-2:i-1])$. This combination is performed by a linear interpolation:

$$p_{com}(w[i]|w[i-2:i-1]) = \lambda * p_{dyn}(w[i]|w[i-2:i-1]) + (1-\lambda) * p_{sta}(w[i]|w[i-2:i-1]).$$
(2.24)

A well known estimation algorithm, the forward-backward algorithm ([6]), is used to estimate the interpolation parameter $\lambda$. $\lambda$ ranges from 0.07 to 0.28 depending on the static tri-gram model used and on the cache size $N$.

Using the combined model, the improvement in performance of the language model ranges between 8% and 23%. As reported in the same paper, for an isolated speech recognizer, this leads to a reduction in error rates ranging from 5% for shorter documents to 24% for larger documents. This is because the cache starts out empty at the beginning of each document and it takes some time before its estimates accurately reflect the particular document. Improvements of about the same size are reported in [85], [86] and [79].

## 2.6  Summary

In this chapter, we gave an overview of the different components of a speech recognizer, defined the task of the component central to this thesis, the language model, and reviewed the most commonly used language models.

We began by giving an overview of the different components of a speech recognizer designed according to the stochastic approach (see section 1.2). We briefly explained the tasks of the signal processor, the vector quantizer, the acoustic model and the language model and how they interact to perform the mapping from the acoustic signal to a string of words. Moreover, we introduced many of the notations used throughout this thesis.

After having described the task of the language model intuitively, we then defined the task of the language model more formally as follows. Given a set of contexts, the task of a language model is to provide a probability distribution for each context and to provide a way of choosing a context given the words recognized so far.

Since probability distribution are used frequently in this thesis, we explained how probability distributions can be estimated from frequency data using the maximum likelihood criterion. Furthermore, we briefly addressed the issue of smoothing probability distributions and presented a very simple smoothing technique, the addition of a small, constant baseline probability. The estimation and smoothing methods that have been presented here are the ones we use for our work.

Having seen the major issues in constructing language models, we reviewed many existing language models (context independent, $N$-gram, $N$-pos, decision tree based and adaptive models). For each language model presented, we focussed on two conceptually important issues: how does the model define the context $c$ for its probability distributions $p(w[i]|c)$ and how many probabilities does the model have to estimate. The first point is important because it determines which linguistic regularities (e.g. the ones involving only the two preceding words) the language model can capture. The second point is important because it determines the amount of data needed to train the model and therefore the situations in which the model can be used.

# Chapter 3

# Analyzing and Improving Language Models

In the last chapter, we reviewed many existing language models for speech recognition. Even though some of these models may achieve good performance, the type of speech recognition tasks we can tackle with existing speech recognition technology, including the language models, is still limited. How can we improve the language models so that we can tackle more complex speech recognition tasks? In this chapter, we propose the central idea of this thesis, namely trying to analyze errors of existing language models in order to subsequently improve the models. We first present some intuitions on improving language models (section 3.1) followed by a derivation of the standard measure used to evaluate language models (section 3.2). Given a term closely related to this perplexity measure, the logarithm of the total probability $LTP$, we then define a weakness of a language model in terms of $LTP$ (section 3.3). For models with several components (e.g. the class based models), we develop the method of probability decomposition (section 3.4) which allows us to analyze the weaknesses of the components separately. We conclude this chapter by showing that the idea of analyzing weaknesses is applicable to many probabilistic models (section 3.5).

## 3.1 Intuitions on Analysing and Improving Language Models

Generally speaking, theories about the world remain valid as long as they correctly predict the observed empirical data. But when contradictory evidence is found, new theories are sometimes found naturally by identifying and analyzing the errors of the old theory. This can be taken as a very simple, intuitive model of scientific progress. By analogy to this model of progress, we propose in this chapter the central idea of this thesis, namely trying to analyze errors or weaknesses of a language model in order to subsequently improve the model.

Before developing this line of thought further, we should try to find out whether something similar has been tried before. In the proceedings of the recent major conference in North America ([57]) and one of the main European conferences ([35]), we did not find one paper that attempts an error analysis of a language model. In all of the typical language modeling literature (e.g. proceedings of previous years, workshops, etc.), we have not come across a paper that tries to tackle the problem from this angle. Therefore, the current literature on language modeling shows an apparent lack of interest in error analysis. This is very surprising, especially when the recent increase in work on the topic is taken into account.

In order to perform an error analysis of a language model, we first have to define what constitutes an error. How do we define an error of a language model? Rather than trying to define an error at this point, we observe that the definition of an error should be related to the performance measure used to evaluate a language model. If they are not related, we can still identify and remove an error, but by doing so, we may not improve the performance of the model because the error is not related to the performance measure. Before defining an error, we therefore introduce the standard measure used to evaluate the performance of a language model.

## 3.2 Evaluating a Language Model

### 3.2.1 A Simple Mathematical Measure for the Quality of a Language Model

Since the task of the model is to predict words, it seems natural to evaluate a model by looking at the probabilities it gives to words in a sample of text. This text is referred to as a *testing text*. The geometric mean of these probabilities (the "average value") therefore seems like a good measure for the ability of the model to predict words. Perplexity (see [62]), the standard yardstick for comparing performances of language models, is just the reciprocal of the geometric mean.

We now present the measure of perplexity in more detail. As introduced in section 2.2, $W = w[1], ..., w[i], ..., w[n]$ denotes a sequence of words. Here, the sequence is the sequence of words in the testing text. Let $c_{k(i)}$ be the context the language model chooses for the prediction of the word $w[i]$ (see section 2.3). Furthermore, $p(w[i] = w_{l(i)}|c_{k(i)})$ denotes the probability assigned to the $i^{th}$ word by the model. The total probability $TP$ of the sequence is

$$TP = p(W) = p(w[1:n]) = \prod_{i=1}^{i=n} p(w[i] = w_{l(i)}|c_{k(i)}).$$ (3.1)

The perplexity $PP$ we just described intuitively as the geometric mean of the probabilities is then

$$PP = (TP)^{-\frac{1}{n}}.$$ (3.2)

For a large sample of text, the total probability $TP$ can get extremely small. Therefore, from a practical point of view, it is more convenient to use the logarithm of the total probability $LTP$

$$LTP = log_2(TP) = \sum_{i=1}^{i=n} log_2(p(w[i] = w_{l(i)}|c_{k(i)})$$ (3.3)

and the logarithm of the perplexity $LP$ [1]

$$LP = log_2(PP) = -\frac{1}{n} log_2(TP) = -\frac{1}{n} LTP.$$ (3.4)

---

[1] By analogy to $TP$ and $LTP$, we would prefer to use the term $LPP$ instead of $LP$. However, since $LP$ is the term used by many other researchers, we will be using it as well.

### 3.2.2 Information, Entropy and Perplexity from an Information Theoretic Point of View

In this section, we will derive the logarithm of the probability LP and the perplexity PP from an information theoretic point of view as measures for the quality of a language model. We will also derive the term of entropy, which we will use later in section 4.2.1, page 64. Most of the material here is taken from [62, pp.472], [122, p.6,p.54] and [135].

Information theory is concerned with sources of information. In simple terms, a source of information is a device that outputs symbols chosen from a finite set $V = \{x_1, ..., x_l\}$ known to the observer. The symbols are chosen according to a statistical law underlying the device. We will write the probability of observing symbol $x_i$ as $p(x_i)$. When an information source outputs a symbol, it provides information by removing the uncertainty about the identity of that symbol. Thus, a source provides more information if the uncertainty about the next symbol is greater. How can we measure the amount of uncertainty we have about the next symbol? If there is such a measure, say $H(p(x_1, ..., x_l))$, it is reasonable to require the following properties:

1) $H(p(x_1, ..., x_l))$ should be continuous in the $p(x_i)$.

2) If all the $p(x_i)$ are equal (e.g. $p(x_i) = \frac{1}{l}$), then $H(p(x_1, ..., x_l))$ should be a monotonic increasing function on $l$. In other words, if all symbols are equiprobable, then there is more uncertainty if there are more symbols.

3) If the choice of the next symbol is broken down into two successive choices, the original $H(p(x_1, ..., x_l))$ should be the weighted sum of the $H$ values of each choice. The meaning of this is illustrated in figure 3.1. In case a), we have three possibilities with probabilities $p(x_1) = \frac{1}{2}, p(x_2) = \frac{1}{3}$ and $p(x_3) = \frac{1}{6}$. In case b), we first choose between two possibilities each of which has the probability $\frac{1}{2}$, and if we have picked the second possibility, we will make another choice between two possibilities with probabilities $\frac{2}{3}$ and $\frac{1}{3}$. We require that

$$H(\frac{1}{2}, \frac{1}{3}, \frac{1}{6}) = H(\frac{1}{2}, \frac{1}{2}) + \frac{1}{2} * H(\frac{2}{3}, \frac{1}{3}).\tag{3.5}$$

It is shown in [135, p.116], that the only $H(p(x_1, ..., x_l))$ satisfying the three requirements is of the form

$$H = -k \sum p(x_i) * log p(x_i).\tag{3.6}$$

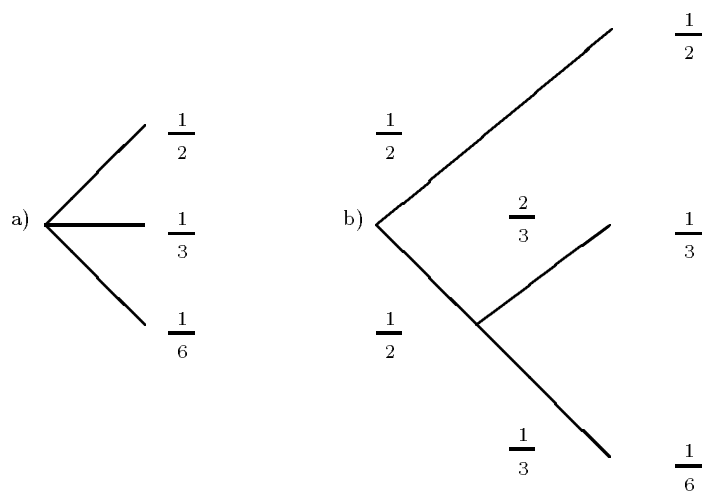

Figure 3.1: Breaking one choice into two successive choices

The constant $k$ merely determines the choice of a unit of measure. Quantities of the form

$$H = -\sum p(x_i) * log p(x_i) \tag{3.7}$$

play a central role in information theory as measures of information, choice and uncertainty. $H$ is also used as entropy in statistical mechanics, where $p(x_i)$ is the probability of a system being in cell $i$ of its phase space (see for example [145]).

One way of understanding intuitively why the logarithm is used is to look at the information provided by a source with $l$ equiprobable symbols. According to equation 3.7, the information content of such a source is

$$H(X) = \sum_{i=1}^{i=l} \frac{1}{l} log l = \frac{1}{l} * l * log l = log l. \tag{3.8}$$

If the source outputs two symbols in a row, we should get twice as much information. However, outputting two symbols is equivalent to a source outputting one of $l^2$ symbols independently and with equal probability. The information content of the second source should therefore be twice the information content of the first. Indeed, because we use the logarithm, we get

$$log l^2 = 2 * log l \tag{3.9}$$

The logarithm therefore conforms to our intuitions about the quantities of information.

Another way of understanding equation 3.7 is to rewrite it as

$$H = -\sum p(x_i) * log p(x_i) = \sum p(x_i) * log \frac{1}{p(x_i)}. \tag{3.10}$$

If $X$ denotes a random variable (our source) over the set $V = \{x_1, ..., x_l\}$, then $H$ is in fact the expected value of $log\frac{1}{p(x_i)}$, where $\frac{1}{p(x_i)}$ is the uncertainty associated with symbol $x_i$. If $x_i$ is very unlikely, then $\frac{1}{p(x_i)}$ is very big, thereby agreeing with our intuition that unlikely events carry a great degree of uncertainty.

As one example of this definition, consider the entropy of a variable that can only take one value, of course with probability one:

$$H(X) = 1 * log\frac{1}{1} = 0. \tag{3.11}$$

Since the outcome is absolutely certain, no information is provided by the source.

The logarithms in equation 3.7 are usually taken to the base two and in this case, the information is measured in units of binary symbols (bits). For example, the information provided by a uniform source with two symbols is one bit:

$$I = log_2 2 = 1. \tag{3.12}$$

The fundamental theory of information theory states (see [135, p.59, Theorem 9]) that on the average, it takes $H$ bits to represent a symbol put out by a source of entropy $H$. Furthermore a source of entropy $H$ provides as much information as a source that chooses it symbols independently, with equal probability, from a vocabulary size of

$$l = 2^H. \tag{3.13}$$

This is because, according to equation 3.8, the entropy $H'$ of the latter source is

$$H' = log2^H = H. \tag{3.14}$$

What about information sources that do not choose their symbols independently of previous symbols? Let $x[i]$ denote the $i^{th}$ symbol output by the source and let $x[i:j], i \leq j$ be a short hand for $x[i], x[i+1], ..., x[j]$. For this more general case, the entropy $H$ is defined as

$$H = -lim_{n \rightarrow \infty} \frac{1}{n} \sum_{x[1:n] \in V^n} p(x[1:n])logp(x[1:n]) \tag{3.15}$$

There is a groups of sources, called *ergodic* sources, for which we can simplify equation 3.15. Even though a rigorous definition of ergodicity [2] is quite complex, the general idea is simple.

_________________________

[2] The interested reader can find more on ergodicity in [135, p.47] and a more rigorous definition in [40].

"In an ergodic process every sequence produced by the process is the same in statistical properties. Thus the letter frequencies, digram frequencies, etc., obtained from particular sequences, will, as the lengths of the sequences increase, approach definite limits independent of the particular sequence. Actually, this is not true of every sequence but the set for which it is false has probability zero. Roughly the ergodic property means statistical homogeneity" ([135, p.45,46])

In the case of ergodic sources, equation 3.15 reduces to

$$H = -lim_{n \to \infty} \frac{1}{n} log p(x[1:n]).$$  (3.16)

In other words, we can estimate the entropy $H$ from long sequences of symbols that were generated by the source.

How can we apply information theory to language models? Language can be seen as an information source whose output symbols are words from the vocabulary $V = \{w_1, ..., w_m\}$ (see section 2.2). We can use formula 3.16 to estimate the content of information per word in a large corpus of text:

$$H = \frac{1}{n} log p(x[1:n]).$$  (3.17)

But how can we get the probabilities of sequences of words $w[1:n]$ of the language, that we need for equation 3.17? We can approximate them with the probabilities $\hat{p}(w[1:n])$ given by the language model. If we replace the true probabilities $p(w[1:n])$ of equation 3.17, with their approximations $\hat{p}(w[1:n])$ given by the language model, we obtain the logarithm of the probability (logprob) $LP$ that we saw in section 3.2:

$$LP = -\frac{1}{n} log \hat{p}(w[1:n]).$$  (3.18)

Intuitively, $LP$ is a measure for the entropy of our model for the language. As pointed out in [62, p.474], we can show that $LP \geq H$ if we assume proper ergodic behavior of the source generating the text. This is clear intuitively, because our model of the language can at most be as good as the language itself. From the view of the speech recognizer, $LP$ measures the difficulty in recognizing speech that was generated by the same source that generated the corpus. Thus, $LP$ is a very appropriate measure for the quality of a language model.

Similar to equation 3.15, we can say that the difficulty of a speech recognition task is also given by the perplexity $PP$ (see section 3.2):

$$PP = 2^{LP} = \hat{p}(w[1:n]).$$  (3.19)

Thus, the speech recognition task with a language model of logprob $LP$ can be thought of as being as difficult as the recognition of a language with $PP$ equally likely words.

### 3.2.3 Discussion of the Standard Perplexity Measure

The ultimate measure for the performance of a speech recognition system is its recognition accuracy. Why then do we want to measure the quality of a language model separately? First, because it allows us to measure the quality of one component of the speech recognizer, the language model, independently of the characteristics of the other components of the particular speech recognition system at hand. Not only does this make language models of different speech recognizers directly comparable, but it also allows researchers to work on the two subtasks separately, thus following the well known "divide and conquer" approach. Second, we can measure the quality of language models that are built for different tasks, e.g. word disambiguation and spelling correction or text encoding.

What do we expect of a measure of the quality of a language model, in particular with respect to speech recognition? Suppose we have two language models, LM1 and LM2, and – according to our performance measure – LM1 is better than LM2. We expect that in general, the recognition accuracy of a speech recognizer that uses LM1 will decrease if it uses LM2 instead. In other words, the measure of the quality of the language model should be highly correlated with the accuracy of any speech recognition system.

The perplexity measure from the previous sections has been shown to correlate well with the recognition accuracy many times. Moreover, it is a theoretically sound measure for the amount of choice in a text generated by the language model. It is therefore a very appropriate measure to use for the evaluating of language models. However, it also has the following problems:

1) The perplexity measure does not take the acoustic similarity of the words into account. Thus, there is no *perfect* correlation between perplexity and recognition accuracy. There have been cases reported in the literature (see for example [21]), where a language model LM1 with a higher perplexity than a model LM2 leads to better recognition accuracy.

2) The quality of the language model depends on the testing text. If we choose a testing text that is very different from the text used to train the model, the model will perform very poorly. However, this does not really mean that the model is bad, but that the testing text is very different from the training text. In fact, the language model may have learned the statistical properties of the training text very well.

3) The language model will ultimately be used to discriminate between likely and unlikely words. It seems that for this purpose, the *difference* in probability between likely and unlikely words is more important than the absolute *value* of the probabilities. In general, negative information is useful in language learning tasks (see for example [42] and [43]). It therefore seems appropriate for the language model to make use of a "false" text (for example a sequence of words chosen at random from a vocabulary or a permutation of an existing text). We could then for example measure the difference in perplexity on the real and the "false" text.

Even though there are problems with the perplexity measure, we choose to use it for our work for the following reasons. First, problem 1) is very rare and in the large majority of cases, the perplexity and the recognition accuracy *are* highly correlated. Second, problem 2) is less severe if we choose a testing text that is quite similar to the training text (and to the task the language model is used for in the end). This is common practice. Third, the amount of work required to investigate problem 3) is beyond the scope of our work here. Last, but not least, perplexity is still the only widely accepted measure for the quality of a language model.

## 3.3 Defining and Identifying Weaknesses of Language Models

Now that we have seen the standard measure used to evaluate language models, we can proceed with our endeavour of defining errors of language models. Since the language model constructs a probability distribution, we do not think that the word *error* is appropriate. Instead, we prefer to use the term weakness, because the intuitive notion of the term *weakness* as something that should be improved is what we want to express. However, in order to avoid a future misunderstanding, let us point out that we use the term weakness in a special, technical sense for the following reason. The commonly used notion of weakness implies that the weak part *can* be improved. In our usage, this might not always be possible. Even an intuitively "perfect" language model will not be able to predict the correct next word with a probability of one [3]. Since our definition of a weakness will be related to the information theoretic term of perplexity, only a model that can predict the next word with absolute certainty (e.g. with a probability of 1) would be perfect. Thus, any model that does not achieve this will still have weaknesses in the information theoretic sense. So even the intuitively "perfect" language model will still have a weakness according to our definition. Even though this might be a theoretical drawback of linking the definition of a weakness to information theory, the disadvantage only appears if the language model approaches the intuitively "perfect" model. However, current $N$-gram models have arguably not yet reached this state (see [63] for a brief comparison between the performance of a tri-gram model and a human guess).

We can now describe a weakness of a language model in terms of the logarithm of the total probability $LTP$, a term very closely related to the standard performance measure for language models (see section 3.2). Intuitively, a weakness of a language model is any part of the model that causes a large fraction of the $LTP$. In the following, we will formalize this intuitive description.

Let us begin by again pointing out that the performance of a language model depends on the testing text used to evaluate the model. This is one of the drawbacks of the standard performance measure and it was already mentioned in section 3.2.3. As we saw in the beginning of this chapter, our measure of weakness should be related to the performance measure. It is thus very clear that our weakness measure will also depend on the testing text. In other words, when we speak of a weakness of a language model, this will always be relative to a given testing text. As pointed out in section 3.2.3, the idea is of course that the testing text should be very similar to the actual data the language model will ultimately be used for. If this is the case, the above mentioned drawback, both for the performance measure and for our weakness measure, will be less severe.

A language model is evaluated on a testing text $W = w[1]...w[i]...w[n]$ (see section 3.2) and the probability the language model assigns to word $w[i]$ is denoted by $p(w[i] = w_{l(i)}|c_{k(i)})$. Furthermore, the $LTP$ of $W$ is calculated as (see equation 3.3)

$$LTP = log_2(TP) = \sum_{i=1}^{i=n} log_2(p(w[i] = w_{l(i)}|c_{k(i)})). \qquad (3.20)$$

In the following, we will denote the testing text $W$ by its index set $I_W = \{1, ..., n\}$. This way, we can denote any subset $W1$ of words of $W$ by giving the subset of indices $I_{W1} \subseteq I_W$. For a given subset $W1$, we can easily determine the $LTP$ it causes ($LTP_{W1}$) by summing up the logarithm of the probabilities of all the words in $W1$:

$$LTP_{W1} = \sum_{i \in W1} log_2(p(w[i] = w_{l(i)}|c_{k(i)})). \qquad (3.21)$$

---

[3]Unless we assume an all-knowing oracle and deny the existence of a free will of the speaker.

Given $LTP_{W1}$, we can then calculate the fraction of $LTP$ caused by $W1$ ($f_{W1}$) as

$$f_{W1} = \frac{LTP_{W1}}{LTP}. \qquad (3.22)$$

**Definition 2** *The impact of a subset $W1 \subseteq W$ is the fraction $f_{W1}$ of LTP the subset $W1$ causes.*

The intuitive idea behind this definition is that we need to improve the language model's prediction of the words that have a big impact, if we want to improve the overall performance significantly. Similarly, if a subset has a low impact, improving the language model's prediction of the words in this subset will not lead to a significant improvement in performance.

Given the impact of a subset $W1 \subseteq W$, we will now derive the part of the language model used in calculating the probability of $W1$ and the impact of this part. A language model contains many probability distributions and each probability distribution contains many probabilities. We therefore say that a language model is made up of a set of probabilities $S = \{p_1, ..., p_l\}$. Furthermore, we will call any subset $S1 \subseteq S$ a part of the model. In order to calculate the probabilities of a subset $W1$ of words (e.g. $p(w[i] = w_{l(i)}|c_{k(i)}), i \in I_{W1}$), the language model will use a subset $S_{W1} \subseteq S$ of its probabilities. Given a subset $W1$, we can then define the part $S_{W1}$ of the model as the subset of probabilities used to calculate the probabilities of words in $W1$. The impact of a part $S_{W1} \subseteq S$ of a language model $S$ is then given by $f_{W1}$, the fraction of $LTP$ that $W1$ causes. Finally, we can now define a weakness of a language model.

**Definition 3** *A part $S_{W1}$ of a language model $S$, defined by a subset $W1$ of the testing text $W$, is called a weakness, if $S_{W1}$ has a great impact.*

The intuitive idea behind this definition is as follows. If subset $W1$ causes a large fraction of $LTP$, then improving it is very important. This conforms to our intuitive meaning of a weakness as something that should be improved.

In the next chapter, we will see what useful results we obtain with this definition. For now, we demonstrate the usefulness of this definition through an example. We define a simple language, a simple model of the language, evaluate the model on a simple testing text and perform the analysis of its weaknesses. We can then verify that, in this case, the weaknesses identified by our definition correspond to what we would like to be identified intuitively as weaknesses.

Consider the language consisting of sequences of the four symbols $\{a, b, c, d\}$. Suppose that we know nothing about this language in general and the frequencies of each symbol in particular. We therefore choose the most simplistic model introduced in section 2.5.1 as a model of the language. This model treats all symbols as equiprobable and therefore assigns the probability $\frac{1}{4}$ to each of the symbols, independent of context. As testing text, we use the string "bddad". When we evaluate this model (see section 3.2), we obtain for its total probability ($TP$), perplexity ($PP$), logarithm of total probability ($LTP$) and logarithm of perplexity ($LP$):

$$TP = p(W) = p(w[1:5]) = \prod_{i=1}^{i=5} p(w[i]) \tag{3.23}$$

$$= p(b) * p(d) * p(d) * p(a) * p(d) = \frac{1}{4} * \frac{1}{4} * \frac{1}{4} * \frac{1}{4} * \frac{1}{4} \tag{3.24}$$

$$= (\frac{1}{4})^5 \approx 0.00098 \tag{3.25}$$

$$PP = (TP)^{-\frac{1}{5}} = 4 \tag{3.26}$$

$$LTP = log_2(TP) = log_2(\frac{1}{4}) + log_2(\frac{1}{4}) + log_2(\frac{1}{4}) + log_2(\frac{1}{4}) + log_2(\frac{1}{4}) \tag{3.27}$$

$$= 5 * (-2) = -10 \tag{3.28}$$

$$LP = log_2(PP) = -\frac{1}{5}LTP = 2 \tag{3.29}$$

$$\tag{3.30}$$

According to the preceding discussion, we can now try to identify the weaknesses of the model. Since there are four basic events distinguished by the model, it seems natural to look at the importance of the subsets $A, B, C$ and $D$ of $W$ that are defined by all occurrences of each symbol in the testing text. For the subset A, we thus obtain the index set $I_A = \{4\}$, the set containing the index of the only occurrence of the symbol "a". Similarly, we obtain $I_B = \{1\}, I_C = \{\}, I_D = \{2, 3, 5\}$. The $LTP$ caused by subset $A$ is simply the sum of the logarithms of the probabilities of all occurrences of "a":

$$LTP_A = \sum_{i \in I_A} log_2(p(w[i])) = log_2(p(w[4])) = log_2(p(a)) = log_2(\frac{1}{4}) = -2 \qquad (3.31)$$

Similarly, we obtain $LTP_B, LTP_C$ and $LTP_D$:

$$
\begin{aligned}
LTP_B &= \sum_{i \in I_B} log_2(p(w[i]) \\
&= log_2(p(w[1])) = log_2(p(b)) = log_2(\frac{1}{4}) = -2 \\
LTP_C &= \sum_{i \in I_C} log_2(p(w[i])) = 0 \\
LTP_D &= \sum_{i \in I_D} log_2(p(w[i])) = log_2(p(w[2])) + log_2(p(w[3])) + log_2(p(w[5])) \\
&= 3 * log_2(p(b)) = 3 * log_2(\frac{1}{4}) = -6.
\end{aligned}
$$

We then divide the $LTP$ caused by each subset by the overall $LTP = -10$ to obtain the fraction of $LTP$ caused by each subset:

$$
\begin{aligned}
f_A &= \frac{LTP_A}{LTP} = \frac{-2}{-10} = 0.2 \\
f_B &= \frac{LTP_B}{LTP} = \frac{-2}{-10} = 0.2 \\
f_C &= \frac{LTP_C}{LTP} = \frac{0}{-10} = 0 \\
f_D &= \frac{LTP_D}{LTP} = \frac{-6}{-10} = 0.6
\end{aligned}
$$

Because all occurrences of the symbol "d" cause 60% of the total $LTP$ (e.g. $f_D = 0.60$), we can see that the prediction of the symbol "d" is the most important one for the performance of our model on this testing text. But what does this result imply for the model? We will now identify the part of the model that is used in calculating the probabilities of the subset $D$. In this example, this will be very straight forward, but as we will see in section 3.4, this is not always the case.

Our model has only one probability distribution containing four probabilities. We thus write our model $S$ as $S = \{p_a, p_b, p_c, p_d\}$. The only probability used in calculating the probabilities of words in $D$ is $p_d$. We thus obtain the part of the model $S_D = \{p_d\}$.

This part causes 60% of the total $LTP$ (e.g. $f_D = 0.6$) and is therefore identified as a weakness of the model. Intuitively, if we have the chance to increase one of the probabilities in $S = \{p_a, p_b, p_c, p_d\}$ (and decrease others in return), we would choose $p_d$, if we want to significantly improve the overall performance of the model. This is also the part that is identified as a weakness using our definition and the results of our analysis therefore correspond well to our intuition.

## 3.4 Probability Decomposition

In the last section, we defined a weakness of a language model as a part of the model that has a great impact on the overall performance. A part of the model was in turn defined as all the probabilities of the language model that are used in calculating the probabilities of a subset of words from the testing text. In some language models (e.g. the $N$-gram models), the probability $p(w[i] = w_l|c_{k(i)})$ of the $i^{th}$ word is just the probability value of the probability distribution for context $c_{k(i)}$. In such models, we can thus measure the impact of a single probability value of the model and, by considering any set of these probability values, the impact of any subset of the model. This allows a very fine-grained analysis of the language model.

However, there are language models in which each probability $p(w[i] = w_l|c_{k(i)})$ is calculated from several probability values of the language model. An example of such a model is the last class based model we saw in section 2.5.3. In this model, the probability of seeing the word 'light' is the probability of seeing it as a noun plus the probability of seeing it as a verb plus the probability of seeing it as an adjective. The exact formula of the model is

$$p(w[i] = w_l|w[i-1]) = \sum_{g_j \in G} p(g(w[i]) = g_j|g(w[i-N+1:i-1])) * p(w[i] = w_l|g(w[i]) = g_j).$$
(3.32)

We can see that the probability of the $i^{th}$ word is calculated as a sum of terms, where each term is the product of two probabilities. The subset of the model, defined by just one word $w[i]$ therefore contains many probability values (as opposed to just one in the $N$-gram model, for example). The parts of the model identified as weaknesses thus tend to be large. But if we want to improve such a large part of the model, which of its probability values are really important? Is it for example the probability values that predict the next class (e.g. $p(g(w[i]) = g_j|g(w[i-1]))$) or is it the probability values that predict the word given the class (e.g. $p(w[i] = w_l | g(w[i]) = g_j)$)? In order to answer this question, we will now develop a method called probability decomposition. It will allow us to divide the probability of the $i^{th}$ word into two parts, the part used to predict the next class and the part used to predict the next word given its class.

To start with, we are given the sum of the form $S = \sum_i a_i * b_i$ representing the bi-pos formula (equation 3.32), where $S$ corresponds to $p(w[i] = w_l | w[i-1])$, $a_i$ to $p(g(w[i]) = g_j | g(w[i-1]))$ and $b_i$ to $p(w[i] = w_l | g(w[i]) = g_j)$. We therefore have $0 < a_i, b_i \leq 1$. In our overall analysis of the language model, $S$ will cause a certain percentage $f_S$ of the total $LTP$. Our goal is to split $f_S$ up into two parts $f_A$ and $f_B$, the fractions of $LTP$ caused by the $a_i$'s and $b_i$'s respectively. This will allow us to concentrate our efforts to improve the model on the $a_i$'s or $b_i$'s, depending on which one has a bigger impact on the overall performance. In order to split $f_S$ into $f_A$ and $f_B$, we need to know the percentage of $S$ that is given by the $a_i$'s and $b_i$'s. How can we calculate that percentage? We start with the most simple case and suppose that the sum is only over one term, e.g. $S = a_1 * b_1$ with $a_1 = \frac{1}{2}, b_1 = \frac{1}{4}, S = \frac{1}{2} * \frac{1}{4} = \frac{1}{8}$. What percentage $p_A$ of $S$ is given by $a_1$? To answer this question, we first need to take a closer look at our intuitive notion of percentage. Suppose we have a sum $S = a_1 + ... + a_n$. When we say $a_1$ is y*100% of S, e.g. $\frac{a_1}{S} = y$, y is a measure of how many $a_1$'s make up the total S with respect to the operator '+'. In fact

$$a_1 + ... + a_1(\frac{1}{y} times) = a_1 * \frac{1}{y} = a_1 * \frac{S}{a_1} = S. \quad (3.33)$$

The same should hold for a product. Suppose we have $S = a_1 * ... * a_n$. If we say $a_1$ is y*100% of P, we mean

$$a_1 * ... * a_1(\frac{1}{y} times) = a_1^{\frac{1}{y}} = P. \quad (3.34)$$

By solving [4] the last equation for $y$, we get the percentage of $P$ given by $a_1$:

$$log_2(a_1^{\frac{1}{y}}) = log_2(P) \quad (3.35)$$

$$\frac{1}{y} * log_2(a_1) = log_2(P) \quad (3.36)$$

$$y = \frac{log_2(a_1)}{log_2(P)}. \quad (3.37)$$

---

[4] We used the logarithm to the base two to solve the equation. But because $\frac{log_a(x)}{log_a(y)} = \frac{log_b(x)}{log_b(y)}$, the notion of percentage does not depend on the base used.

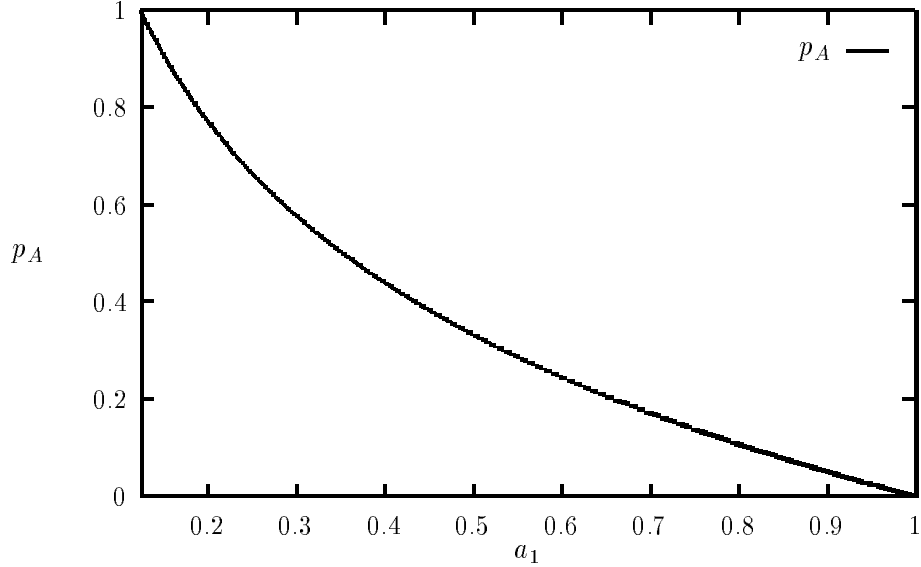

Figure 3.2: The graph of $p_A$

Thus, in our example, the percentage $p_A$ of $S = \frac{1}{8}$ given by $a_1 = \frac{1}{2}$ is

$$p_A = \frac{log_2(\frac{1}{2})}{log_2(\frac{1}{8})} = \frac{1}{3}. \tag{3.38}$$

To get a better intuitive understanding of $p_A$, we plot in figure 3.2 the relationship between $p_A$ and $a_1$ for $S = \frac{1}{8}, \frac{1}{8} < a_1 \leq 1$. We limit $a_1$ to the range $\frac{1}{8} < a_1 \leq 1$ because only in this range we can find a $b_1$ in the range $0 < b_1 \leq 1$ such that $S = a_1 * b_1$. Because the curve is falling, we can see that a smaller $a_1$ corresponds to a higher percentage. Thus if we have $S = \frac{1}{8} = a_1 * b_1$ with $a_1 = \frac{1}{4}, b_1 = \frac{1}{2}$ and $a'_1 = \frac{1}{8}, b'_1 = 1$, then $a'_1$ causes a higher percentage of $S$ than $a_1$.

How can we calculate the percentage $p_A$ of $S$ given by $a_i$ if the sum is over several terms? As an example, consider $S = a_1 * b_1 + a_2 * b_2$ with $a_1 = \frac{1}{2}, b_1 = \frac{1}{4}, a_2 = \frac{1}{2}, b_2 = \frac{1}{2}, S = \frac{1}{2} * \frac{1}{4} + \frac{1}{2} * \frac{1}{2} = \frac{3}{8}$. We can first determine the percentage $p_{A1}$ of $a_1 * b_1$ given by $a_1$ (as above) and the percentage $p_{A2}$ of $a_2 * b_2$ given by $a_2$. Given $p_{A1}$ and $p_{A2}$, we could obtain the overall percentage $p_A$ by simply taking the average, e.g.

$$p_A = \frac{1}{2} * (p_{A1} + p_{A2}). \tag{3.39}$$

However, since $a_1 * b_1$ and $a_2 * b_2$ make up different portions of the total sum $S$, it is more fair to weigh $p_{A1}$ and $p_{A2}$ according to the portion of $S$ represented by $a_1 * b_1$ and $a_2 * b_2$ respectively. This gives the weighted average

$$p_A = \frac{a_1 b_1}{S} * p_{A1} + \frac{a_2 b_2}{S} * p_{A2}. \tag{3.40}$$

Similarly, we can calculate the percentage $p_B$ of $S$ given by $b_i$ as

$$p_B = \frac{a_1 b_1}{S} * p_{B1} + \frac{a_2 b_2}{S} * p_{B2}. \tag{3.41}$$

In our example, this gives

$$p_A = \frac{a_1 * b_1}{S} * \frac{log_2(a_1)}{log_2(a_1 b_1)} + \frac{a_2 * b_2}{S} * \frac{log_2(a_2)}{log_2(a_2 b_2)} \tag{3.42}$$

$$= \frac{\frac{1}{8}}{\frac{3}{8}} * \frac{log_2(\frac{1}{2})}{log_2(\frac{1}{8})} + \frac{\frac{1}{4}}{\frac{3}{8}} * \frac{log_2(\frac{1}{2})}{log_2(\frac{1}{4})} \tag{3.43}$$

$$= \frac{1}{3} * \frac{1}{3} + \frac{2}{3} * \frac{1}{2} \tag{3.44}$$

$$= \frac{4}{9} \tag{3.45}$$

$$p_B = \frac{a_1 * b_1}{S} * \frac{log_2(b_1)}{log_2(a_1 b_1)} + \frac{a_2 * b_2}{S} * \frac{log_2(b_2)}{log_2(a_2 b_2)} \tag{3.46}$$

$$= \frac{\frac{1}{8}}{\frac{3}{8}} * \frac{log_2(\frac{1}{4})}{log_2(\frac{1}{8})} + \frac{\frac{1}{4}}{\frac{3}{8}} * \frac{log_2(\frac{1}{2})}{log_2(\frac{1}{4})} \tag{3.47}$$

$$= \frac{1}{3} * \frac{2}{3} + \frac{2}{3} * \frac{1}{2} \tag{3.48}$$

$$= \frac{5}{9} \tag{3.49}$$

$$\tag{3.50}$$

Knowing that $p_A$ and $p_B$ are the percentages of $S$ given by $a_i$ and $b_i$, we can then simply write $S = A * B$ with

$$A = S^{p_A}, B = S^{p_B}. \tag{3.51}$$

In our example, we thus get

$$A = S^{p_A} = (\frac{3}{8})^{\frac{4}{9}} \approx 0.6466 \tag{3.52}$$

$$B = S^{p_B} = (\frac{3}{8})^{\frac{5}{9}} \approx 0.5799 \tag{3.53}$$

$$A * B = (\frac{3}{8})^{\frac{4}{9}} * (\frac{3}{8})^{\frac{5}{9}} = \frac{3}{8}. \tag{3.54}$$

$$\tag{3.55}$$

We have now decomposed the value of $S = \frac{3}{8}$ into $S = A * B$ with $A \approx 0.6466$ and $B \approx 0.5799$. What is so special about these values, among all the possible values for $A$ and $B$? They are special because the percentage of $S$ given by $A$ is $\frac{4}{9}$, e.g. $p_A = \frac{log_2(A)}{log_2(B)} = \frac{4}{9}$, the weighted average of the percentages of $a_i * b_i$ caused by $a_i$.

We will now extend this method to the general case. Given $S = \sum a_i * b_i$, we will decompose $S$ as follows:

$$S \quad = A * B \tag{3.56}$$

$$A \quad = \quad S^{p_A} \tag{3.57}$$

$$B \quad = \quad S^{p_B} \tag{3.58}$$

$$p_A \quad = \quad \sum_i \frac{a_i * b_i}{S} * \frac{log_2(a_i)}{log_2(a_i b_i)} \tag{3.59}$$

$$p_B \quad = \quad \sum_i \frac{a_i * b_i}{S} * \frac{log_2(b_i)}{log_2(a_i b_i)}. \tag{3.60}$$

We can verify that for this choice of $A$ and $B$, the percentage of $S$ caused by $A$ and $B$ is indeed $p_A$ and $p_B$ respectively and that the multiplication of $A$ and $B$ indeed gives $S$:

$$\frac{log_2(A)}{log_2(AB)} \quad = \quad \frac{log_2(S^{p_A})}{log_2(S)} = p_A \tag{3.61}$$

$$\frac{log_2(B)}{log_2(AB)} \quad = \quad \frac{log_2(S^{p_B})}{log_2(S)} = p_B \tag{3.62}$$

$$A * B \quad = \quad S^{p_A} * S^{p_B} = S^{\frac{log_2(A)}{log_2(AB)} + \frac{log_2(B)}{log_2(AB)}} = S. \tag{3.63}$$

$$\tag{3.64}$$

Given this method of probability decomposition, we can now replace the probability $p(w[i] = w_l | c_{k(i)})$ of the $i^{th}$ word with

$$p(w[i] = w_l | c_{k(i)}) = S = A_i * B_i \tag{3.65}$$

For one, this will allow us to look at the fraction of $LTP$ caused by different contexts. Moreover, by only looking at all the $A_i$'s (or $B_i$'s), we can now treat each component as a separate model, and this way, we can analyze the weaknesses of each component separately.

Please note that we can extend the method of probability decomposition to models that have more than two components. Suppose for example that a language model calculates the probability of the $i^{th}$ word as

$$p(w[i] = w_l | c_{k(i)}) = S = \sum a_i * b_i * c_i \tag{3.66}$$

We can then simply write

$$S \quad = A * B * C \tag{3.67}$$

$$A \quad = \quad S^{p_A} \tag{3.68}$$

$$B \quad = \quad S^{p_B} \tag{3.69}$$

$$C \quad = \quad S^{p_C} \tag{3.70}$$

$$p_A \quad = \quad \sum_i \frac{a_i * b_i * c_i}{S} * \frac{log_2(a_i)}{log_2(a_i b_i c_i)} \tag{3.71}$$

$$p_B \quad = \quad \sum_i \frac{a_i * b_i * c_i}{S} * \frac{log_2(b_i)}{log_2(a_i b_i c_i)} \tag{3.72}$$

$$p_C \quad = \quad \sum_i \frac{a_i * b_i * c_i}{S} * \frac{log_2(c_i)}{log_2(a_i b_i) c_i}. \tag{3.73}$$

$$\tag{3.74}$$

## 3.5  Applicability

The idea of identifying weaknesses of probabilistic models by measuring the amount of $LTP$ caused by different events is very general. It applies to all probabilistic models that derive a score for a sequence of tokens by multiplying the probabilities of individual tokens and that are evaluated using the perplexity measure.

Examples of models to which our idea of identifying weaknesses is applicable are all the models reviewed in section (2.5), ranging from N-gram over N-pos to models based on decision trees. Furthermore, the idea is also directly applicable to models that are based on units different from words, such as syllable based language models ([105]) and phone based language models ([96], [130]), or to models like [47], where the language model is made dependent on the state of a LR parser.

In general, speech recognition systems have been based on phonemes ([142]), diphones ([103], [22], [1], [133]), syllables ([55], [154], [44], demi-syllables ([127], [124]), and disyllables ([137]). Language models can be built on all of these levels and the idea proposed here is applicable to all of them. As an example, we will show how the idea of identifying weaknesses can be applied to a syllable based language model.

The basic linguistic unit in Japanese sentences is the syllable (see [105]), corresponding roughly to a consonant-vowel unit. The syllable therefore constitutes a convenient unit for recognition of Japanese speech and is, for example, used in the Japanese phonetic typewriter (see [74]). Given the syllable $s_1, ..., s_{i-1}$ recognized so far, the acoustic model can decide which syllables are likely to correspond to the next stretch of the acoustic signal. However, as in word based recognition, the acoustic model may not be able to decide based on the signal alone which one of the likely syllables was the one spoken. The phonetic typewriter therefore uses a syllable based trigram language model to decide which syllables are likely to appear next, given that the syllables recognized so far are $s_1, ..., s_{i-1}$. As usual, this syllable trigram language model is trained on large amounts of text and its performance is measured on a testing text. How can we improve such an existing syllable based language model? Following the idea proposed here, we can first identify the weaknesses of the syllable based language model. We therefore should identify the syllables whose predictions account for a large fraction of the $LTP$. Once these syllables have been identified, we can look at why they account for such a large fraction of $LTP$ and, more importantly, how we can improve the language model to avoid this weakness.

An example of a model to which our definition of weakness is not applicable is a probabilistic context free grammar (see for example [157]). This is because the probability of a sequence of words is not obtained by multiplying the probabilities of each word.

Besides language modeling for speech recognition, $N$-gram based probabilistic models are also used in the area of optical and handwritten character recognition ([54], [123] and [138]). However, according to a recent survey on optical character recognition ([58, p.11]), the $N$-gram model does not come under the name of language model, but is referred to as contextual processing and is one part of the postprocessing in optical character recognition. Nevertheless, the principles are very similar to language models in speech recognition. For example, in [138], the Viterbi algorithm ([149]) is used as in speech recognition, to find the best sequence of letters according to the probabilistic scores provided by two models (corresponding to the acoustic model and the language model in speech recognition). The model that corresponds to the language model is in fact a letter bi-gram language model. In other words, it calculates the probability of a sequence of letters $z_1, ..., z_n$ by multiplying the probabilities of each letter $z_i$, which only depends on the previous letter. This is expressed in the following formula:

$$p(z_1, ..., z_n) = \sum_{i=1}^{i=n} p(z_i | z_{i-1}). \tag{3.75}$$

The probabilities $p(z_i | z_{i-1})$ of letter bi-grams are estimated from the Brown corpus. As with the syllable-based Japanese language model we just saw, we can apply our analysis of weaknesses to this letter based language model in order to subsequently improve the model.

## 3.6 Summary

In this chapter, which contains the central idea of this thesis, we proposed to perform analyses of weaknesses of language models in order to improve the models afterwards, defined what we meant by "weakness of a language model" and presented a method to identify the weaknesses of a given model.

We began by noting the widely accepted idea that improving any kind of model or theory is usually easier once its shortcomings are known. Moreover, in a very simple, intuitive model of scientific progress, the knowledge of the errors of an existing theory is crucial. In analogy to this model of progress, we proposed in this chapter to analyze weaknesses of a language model in order to subsequently improve the model.

Since the measure of weakness of a language model should be related to the performance measure used to evaluate the model, we turned to the standard perplexity measure used for evaluating language models. We introduced perplexity intuitively as the reciprocal of the geometric mean of probabilities assigned to the words in the testing text, derived the measure from an information theoretic point of view and discussed its advantages and shortcomings.

Given the perplexity and the closely related logarithm of the total probability $LTP$, we defined a weakness of a language model in terms of $LTP$ [5]. A part of a language model defined by a subset of words of the testing text is called a weakness if it causes a large fraction of the $LTP$. This conforms to our intuitive meaning of a weakness as something that should be improved, because if we want to improve the model significantly, it is very important to improve the parts of the model that cause a large fraction of the $LTP$. For models with separate components (e.g. the class based models), we also developed the method of probability decomposition, which allows us to analyze the weaknesses of each component separately.

After having defined a weakness of a language model, we noted that this definition is applicable to any probabilistic model that derives a score for a sequence of symbols by multiplying the probabilities of each symbol and that is evaluated with the perplexity measure. We can thus apply our definition to almost all of the commonly used language models, including models based on units different from words (e.g. phonemes, syllables). As an example, we showed briefly how we can apply our definition to a Japanese syllable tri-gram model and how this could help in improving the model. Besides language modeling for speech recognition, our idea of analyzing weaknesses is also applicable to other areas that use $N$-gram statistics, e.g. handwriting and optical character recognition.

---

[5]For a discussion of which of the commonly associated intuitions of the term weakness also apply to our technical use of the term, see page 42.

# Chapter 4

# Analyzing and Improving a Bi-pos Language Model

In the last chapter, we presented the main idea of this thesis, namely to perform an analysis of the weaknesses of language models. Moreover, we defined what we mean by a weakness of a language model. In this chapter, we apply this definition and the idea of analyzing weaknesses to a concrete language model. For that purpose, we first choose a training and testing corpus and a language model (section 4.1). We verify that the training data we use is sufficient to train our model and this leads to a discussion of the issue of sample space (section 4.2). We then proceed with the analysis of the weaknesses of our model and present the results (section 4.3). Trying to improve one of the identified weaknesses leads to the development of the generalized $N$-pos model (section 4.4).

## 4.1   Choosing a Corpus and a Language Model

The corpus used to train and test a language model is the primary source of information used in the model. It is crucial to the overall undertaking that the corpus contains sufficient data to train the model. We could of course choose a language model and then a corpus, but we would have no guarantee that the corpus contains enough training data. Since the corpus size determines the complexity of the language model that can be adequately trained with this amount of data, we will first choose a corpus and then a model.

### 4.1.1 Choosing a Corpus

A corpus is a collection of text in machine readable format, often annotated with additional information. These annotations can supply information related to, for example, the parts of speech of each word, the parse tree of each sentence or prosodic information.

Before focusing on a specific corpus for our work, let us consider the wide range of existing corpora. According to a recent review of corpora research and construction (see [31]), the three most commonly used corpora are the Brown corpus (see [82], [39], [81]), the Lancaster-Oslo-Bergen (LOB) corpus (see [71], [70], [97]), and the London-Lund corpus (see [143]). But a large variety of other corpora exist: the Lancaster Spoken English Corpus (SEC) (see [80]), the British National Corpus (see [119]), the Wall Street Journal Corpus (available from the ACL/DCI), and the International Corpus of English (see [49]) – just to name a few. Most of these corpora are available through institutions and initiatives, which have been created recently to oversee the collection of linguistic data, e.g. the data collection initiative of the Association for Computational Linguistics (see [99], [17], [152], [153]), the European Corpus Initiative of the Association for Computational Linguistics and the Linguistic Data Consortium.

In choosing one of the available corpora for our work, we will consider two criteria. First, the corpus should be used for language modeling by other researchers. This makes the results more widely acceptable and reproducible. Second, we prefer a "small" corpus (e.g. less than a million words) for the following three reasons.

First, many current speech recognizers are intended for a specific application domain (e.g. medical texts). As pointed out in [117], the performance of a language model in such a specific domain of application is often better if we train the language model on a small corpus task from the domain of application than if the language model is trained on a large corpus not specific to the domain. We therefore need to train language models on small, domain specific corpora [1] .

Second, if we find a technique to improve a language model trained on a small corpus, it is likely that this technique is also applicable to language models trained on large corpora. On the other hand, if we find a technique to improve a language model trained on a large corpus, this technique might require the availability of a large amount of training data

---

[1] Domain specific corpora are most likely very small, because corpora are expensive to produce. As an example consider the TI Digit Corpus ([98]), a collection of a large set of spoken digits. It required an estimated $300,000 to $400,000 for its construction.

and the technique might therefore not be applicable to language models trained on small corpora.

Finally, from a practical point of view, it is usually easier to understand a complex problem by looking at simple instances of the problem. In the case of a language model, a simple instance is a simple model, requiring little data to be stored, handled and analyzed.

Using these criteria, the LOB corpus is an adequate choice. It contains about one million words and is small compared to, for example, the Wall Street Journal Corpus (50 million words) or the British National Corpus (100 million words). Moreover, many researchers working on language modeling also use the LOB ([85], [109], [34], [106], [78], [110]).

Even though we are building language models for speech recognition, the corpus is constructed from written text. This is common practice, mainly for practical reasons. Large quantities of written text are already in a format that can be used for a corpus, whereas the transcription of spoken text is time consuming, tedious and expensive.

The LOB corpus is divided into 500 samples of text. Each sample contains slightly more than 2000 words and each word is tagged with one of 153 possible syntactic classes. These syntactic classes correspond to the parts of speech (pos) that we mentioned when we introduce the $N$-pos models (see section 2.5.3, page 27). The samples are grouped into 15 different categories, depending on the source of the text. Table 4.1 shows the 15 different categories and the number of samples in each. We see that the corpus covers a wide range of English prose. An example of the corpus material can be found in Appendix A. We use the first 50,000 words of sections A1-A34 as training text and roughly 25,000 words (sections A35-A44) as testing text. In section 4.2.3, we justify why 50,000 words constitutes enough training data.

It has been reported in the literature (for example [60], [78]) that the number of classes or tags a language model uses influences the performance of the model. In order to make our results less dependent on the number of tags provided with the corpus, we therefore decided to use more than one set of tags. We could have produced different tagsets automatically, as suggested in [62], [60], [78] and [111] . However, since this is not the main issue addressed in this thesis, we used the following simple heuristic to construct four different tagsets. The original tagset contains, for example, four different tags for adverbial nouns (NR, NR$, NRS, NRS$) and twelve different tags for proper nouns. We construct smaller tagsets by, for example, merging all four tags for adverbial nouns into one tag (NR), or by merging all twelve tags for proper nouns into one tag (NP). We can then construct an even smaller tagset by merging the set we just constructed for adverbial nouns (NR) and proper nouns (NP) into one tag for nouns (N). Thus, starting from the original tagset, we construct three other tagsets by merging tags with the same prefix. The resulting three tagsets have 88, 42 and 24 tags respectively. All four tagsets, together with examples for each tag are shown in appendix B.

| Category | Description | number of samples |
|----------|-------------|-------------------|
| A | Press: Reportage | 44 |
| B | Press: Editorials | 27 |
| C | Press: Reviews | 17 |
| D | Religion | 17 |
| E | Skills, Trades and Hobbies | 38 |
| F | Popular Lore | 44 |
| G | Belles Lettres, Biography and Essays | 77 |
| H | Miscellaneous | 30 |
| J | Learned and Scientific Writings | 80 |
| K | General Fiction | 29 |
| L | Mystery and Detective Fiction | 24 |
| M | Science Fiction | 6 |
| N | Adventure and Western Fiction | 29 |
| P | Romance and Love Story | 29 |
| R | Humour | 9 |

Table 4.1: The different categories of the LOB corpus

### 4.1.2 Choosing a Language Model

In recent years, a great number of different language models have been developed (see section 2.5). Most commonly used are bi-gram, tri-gram, bi-pos and tri-pos models. They differ significantly in their complexity and in the amount of training data needed. Which of the models should we use for our work? As was pointed out in the last section, we use a rather small corpus. Furthermore, the argument in favor of a simpler model making a complex process easier to understand is also valid for the choice of language model. We therefore chose the model requiring the least amount of training data, the bi-pos model. As mentioned in section 2.5.3, existing N-pos models furthermore differ in the fact that the classes the models use are overlapping or that they are mutually exclusive. Since the classes in the LOB are overlapping, we chose a model allowing multiple class membership. For example, the word 'light' can be a noun, verb or adjective depending on the context in which it is used.

Now that we have decided which of the models reviewed in section 2.5 we are going to use for our work, we will present the chosen model in more detail, describing the probability distributions in terms of smoothed frequency counts. But first, let us recall the model in terms of its probability distributions, as it was presented in section 2.5.3. The probability that the $i^{th}$ word $w[i]$ is the word $w_l$ is calculated as the sum (over all classes $g_j$) of the probabilities that the $i^{th}$ word $w[i]$ is $w_l$, where $w_l$ appears with the particular class $g_j$. The probability of the word $w_l$ appearing with a particular tag $g_j$ is the probability of having this tag $g_j$ ($g(w[i]) = g_j$) given the tag $g(w[i-1])$ of the previous word ($p(g(w[i]) = g_j|g(w[i-1]))$) times the probability of having word $w_l$ given the tag $g_j$ ($p(w[i] = w_l|g(w[i]) = g_j)$). This is expressed more precisely by the following formula (see section 2.5.3):

$$p(w[i] = w_l|w[i-1])$$
$$= \sum_{g_j \in G} p(g(w[i]) = g_j|g(w[i-1])) * p(w[i] = w_l|g(w[i]) = g_j).$$

As shown in section 2.4, we can estimate the probability distributions $p(g(w[i]) = g_j|g(w[i-1]))$ and $p(w[i] = w_l|g(w[i]) = g_j)$ in terms of frequencies $f(g(w[i]) = g_j|g(w[i-1]))$ and $f(w[i] = w_l|g(w[i]) = g_j)$ of events:

$$p(w[i] = w_l|w[i-1])$$
$$= \sum_{g_j \in G} f(g(w[i]) = g_j|g(w[i-1])) * f(w[i] = w_l|g(w[i]) = g_j).$$

In order to avoid zero probabilities (see section 2.4), we have to ensure that at least for one tag $g_0 \in G$, both frequencies in formula 4.1 are different from zero. If we suppose that every word $w_l$ of our vocabulary occurred at least once in our training text, then the occurrence of word $w_l$ has one tag $g_0$ associated with it and the factor $f(w[i] = w_l|g(w[i]) = g_0)$ in formula 4.1 is different from zero. However, if $g_0$ never occurred after the previous tag $g(w[i-1])$, the factor $f(g(w[i]) = g_0|g(w[i-1]))$ in formula 4.1 is zero. In order to avoid the second factor to be zero, we therefore smooth the second distribution. As suggested in [85], and as explained in section 2.2, we add a small constant probability value of $c_2$ and then use a matching constant $c_1$ to ensure that the sum over the probabilities of all the words is one:

$$p(w[i] = w_l|w[i-1])$$
$$= \sum_{g_j \in G} f(g(w[i]) = g_j|g(w[i-1])) * (c_1 * f(w[i] = w_l|g(w[i]) = g_j) + c_2).$$

The language model is evaluated on a testing text, which may contain words that are not part of the vocabulary $V$. We therefore have to adjust formula 4.1 to deal with these so called unknown words. We again adopt the approach taken in [85] and treat every unknown word as the occurrence of one special symbol, say $unknown$. We give a constant probability value of $d$ to occurrence of this symbol and have to multiply all other probabilities by $(1 - d)$ in order to ensure that the sum of probabilities of all words in the vocabulary plus the probability of the symbol $unknown$ sums up to one:

$$P(w[i] = w_l | w[i-1])$$
$$= \begin{cases} (1-d) \sum_{g_j \in G}[(c_1 * f(g(w[i]) = g_j | g(w[i-1]))) + c_2) \\ * f(w[i] = w_l | g(w[i]) = g_j)] & \text{if } w_l \in V \\ d & \text{otherwise .} \end{cases}$$

Following [85], we estimate the value of $d$ by Turing's formula as the number of unique words in the training text divided by the total number of words in the training text. $d$ decreases when the amount of training data increases and for our training text of 50,000 words, we obtain $d = \frac{4637}{50000} \approx 0.093$. For $c_2$, we chose the arbitrary value of $10^{-4}$ as suggested in [85]. As explained in section 2.4, we have to choose $c_1 = 1 - |G| * c_2$, which gives $c_2 = 0.9958$ for the tagset with 42 tags.

In formula 4.1, we need to know the tag of the word $w[i-1]$ in order to calculate the probability of word $w[i]$, even if word $w[i-1]$ is an unknown word. We could simply use the tags provided with our tagged testing text, but this would be 'cheating', because we would be using a source of information external [2] to the language model to make its task easier. Following [85], our model uses a heuristic in order to assign a tag to each word and the following three cases can occur:

1) the word is unknown, e.g. it did not occur in the training text and is not part of the vocabulary.

2) the word always occurred with the same tag in the training text.

3) the word occurred with several tags in the training text.

The model deals with these cases as follows:

---

[2]The tags were assigned to words manually or by a separate program (a tagger), and they are therefore not part of the information the language model can use.

1) the model chooses the tag $g_j$ which has the highest value of $f(g(w[i]) = g_j|g(w[i-1]))$, e.g. the tag that is most likely to follow the preceding tag.

2) the model uses the unique tag of the word.

3) the model chooses the tag $g_j$ which contributes the largest probability to the sum over all possible tags in equation 4.1.

## 4.2 Differences in Sample Spaces

In section 4.1, we chose a corpus and a language model for our work. Before we can proceed with the analysis of the weaknesses of the model, it is important to make sure that there is sufficient data in the corpus to train the model. Otherwise, the results would not be significant. As we will see in section 4.2.2, the analysis of the influence of the amount of training data on the model produces a counter-intuitive result: as the amount of training data increases, the performance of the model decreases. In order to understand this behaviour, it is useful to look at the underlying statistical issue, the issue of sample space.

In statistics ([41]), the word 'experiment' is used in a very wide sense, and it refers to any process of observation or measurement. The results obtained from an experiment are called its outcomes. The set of all possible outcomes for each experiment is called the sample space. Probabilities are derived as the ratio of successful outcomes to all possible outcomes and the sum of the probabilities of all events in the sample space has to be one. In a language model, the experiment is the observation of the identity of the word that occurs next in a given context. One outcome is the occurrence of one particular word and the sample space is the set of all words that can be observed.

### 4.2.1 Differences Due to the Modeling of Unknown Words

As we have seen on page 61, a language model has to deal with unknown words. In the literature, we can find several ways of approaching the issue of unknown words and we will present four different models, M1 to M4, in the following paragraphs. For our purpose, it is not crucial how probabilities are derived exactly for these unknown words, but rather what the underlying sample space is. We will therefore only describe the sample space for each of the four models.

We will denote the sample space by $S$ and the number of words in the training and testing text by $n_{train}$ and $n_{test}$ respectively. Let V(train[i]) denote the vocabulary derived from the first i words of training text and let V(train) be the shorthand notation for V(train[$n_{train}$]). Similarly, we define $V(test[i])$ and $V(test)$. Models, that change vocabularies during the testing will have a sequence of vocabularies denoted by $V_0, V_1, ..., V_{test_n}$. Furthermore, as introduced on page 61, we will denote the unknown symbol with *unknown*. The four models are:

**M1** Every occurrence of an unknown word is treated as the occurrence of the symbol *unknown*. This gives the sample space $S =$ V(train) $\cup$ {*unknown*}. This model was used by R. Kuhn and R. de Mori in [85].

**M2** All unknown words are again treated as one special *unknown* symbol. However, after an unknown word occurs, it is added to the vocabulary. We therefore get a sequence of vocabularies $V_0 =$ V(train), $V_1 =$ V(train) $\cup$ V(test[1]), ..., $V_{n_{test}} =$ V(train) $\cup$ V(test). This corresponds to a sequence of sample spaces $S_i = V_i \cup$ {*unknown*}. $S_0$ is equal to the sample space of model M1 and $S_{n_{test}}$ is equal to the sample space of model M3. This adaptive model was proposed by Jelinek et. al in [65].

**M3** The model that looks at the testing text in advance ([83]). All the words that would be unknown are added to the vocabulary before the testing begins, giving the sample space $S =$ V(train) $\cup$ V(test).

**M4** The model that derives a probability for an unknown word based on a character by character probability. It thus distinguishes between all possible unknown words. This corresponds to a sample space $S =$ V(train) $\cup\{(s_1, ..., s_k)\}$ for all $k \geq 1$ where each $s_i$ is one of the 95 printable ASCII characters. $S$ therefore has an infinite number of elements. This model was proposed by Brown et al. in [13].

Recall from section 3.2.1 that the quality of a language model is measured by the perplexity, the reciprocal of the geometric mean of the probabilities assigned to words in the testing text. We think that it is not meaningful to compare probabilities that are based on different sample spaces and we will illustrate this point in three different ways.

First, consider the extreme case of M1 trained on zero words of text. This gives a sample space that only contains one element, the *unknown* symbol. Since there are no other elements, the model gives the probability of 1 to this symbol. All words of the testing text will be treated as occurrences of *unknown* and the text will be reduced to a repetition of this symbol. It is clear that the model assigning the probability of 1 to this symbol will 'achieve' a perplexity of 1, which would imply that this is a particularly good language model!

Second, consider the number of different words that are distinguished in the above sample spaces, but that are not part of V(train). The last model, M4, has a countable infinite number of these words, whereas the first model, M1, has only one. If we have a fixed amount of probability to allocate to these unknown words, then it is clear that in models with many unknown words, each one will 'receive' a very small probability. To conclude from the high perplexity of such a model that it is worse at modeling the language is not really correct, because it solves a different task.

Third, consider all the unknown words in the testing text, that the model M1 treats as the one symbol *unknown*. If we treat all these as separate words, then the sum of probabilities of all words will be more than one. The model therefore does not construct a probability distribution and we therefore can't compare these 'probabilities' with probabilities of models.

All these examples illustrate the point that models with different sample spaces are in fact solving different tasks and can't be compared using the standard perplexity measure.

The amount by which the perplexity results are distorted depends on exactly how different the sample spaces are and how much of the total perplexity is caused by these unknown words. For example, if the model has a very big vocabulary, it has a higher coverage of words in the testing text and the perplexity caused by unknown words is very small. In the experiments reported in [13], the unknown words only account for roughly 5% of the total entropy (see section 3.2.2) . If we reduce the sample space by using M1, M2 or M3 (instead of M4), the entropy could at most be reduced by these 5%. The distortion is so small (at most 5%), because this model was trained on approximately 583 million words, it has a vocabulary of roughly 293,000 entries and only about 1% of the tokens in the testing text are unknown. In order to see how big the distortion is in a model trained on less data, we implemented and ran M1 and M4 on our 50,000 words of training text. We chose M1 and M4 because they differ the most in the size of their sample spaces and hence the distortion should be bigger than between other models. We ran our bi-pos model (see section 4.1.2) with M1 or M4 as models of unknown words, using our testing text with about 14% of unknown words. The results are shown in figure 4.2. The overall perplexity was $2.6 * 10^2$ when using M1 and $4.6 * 10^4$ when using M4 (see section 4.3.2). This huge difference can be understood by looking at the geometric mean of the probabilities assigned by the two models to unknown words. The first model had a geometric mean of approximately $7.4 * 10^{-2}$, the second model of approximately $9.39 * 10^{-17}$. The impact on the overall perplexity of this difference in probability is so much bigger in our experiment because the unknown words account for roughly 51% of the perplexity when M4 is used (see section 4.3.2).

| Model | perplexity | avg. prob of unknown words |
|-------|------------|---------------------------|
| M1 | $2.6 * 10^2$ | $7.4 * 10^{-2}$ |
| M4 | $4.6 * 10^4$ | $9.4 * 10^{-17}$ |

Table 4.2: Comparison of models M1 and M4

We do not try to decide here what the "correct" sample space should be. This should be done by the research community in general and, as one of the reviewers of our paper [147] pointed out, researchers seem to favor models like M4 because these models do not require researchers to agree on a fixed vocabulary. However, as long as different models have different sample spaces, one should keep in mind the distortion this can cause to the perplexity results when comparing language models with this measure.

The focus of this work is the analysis of the main part of the model, not the prediction of unknown words. For the remainder of the work, we therefore want to choose the model in which the prediction of unknown words plays the least important role. Hence, we will use model M1 for the remainder of this thesis.

### 4.2.2 Differences Due to Different Amounts of Training Text

In our implementation of M1 (see equation 4.1 for the exact equation), we made the following observation: as the training text increases in size, the performance [3] of the model decreases. This can be seen in Figure 4.1, which shows the logarithm of the total probability $LTP$ (see section 3.2.1) of the testing text for different sizes of training texts and a set of 42 tags. $LTP$ decreases as the training size increases. This behaviour is counter-intuitive because, as the training text increases, the model should get better at predicting the testing text and the $LTP$ should increase. In order to find the reason for this behaviour, we look at the parts of $LTP$, that are caused by non-vocabulary words, $LTP(unknown)$, and by vocabulary words,

---

[3]As pointed out in section 3.2.1, it is more convenient to use $LTP$ instead of $PP$. Moreover, since $LTP$ and $PP$ measure the same property, this does not influence the results.

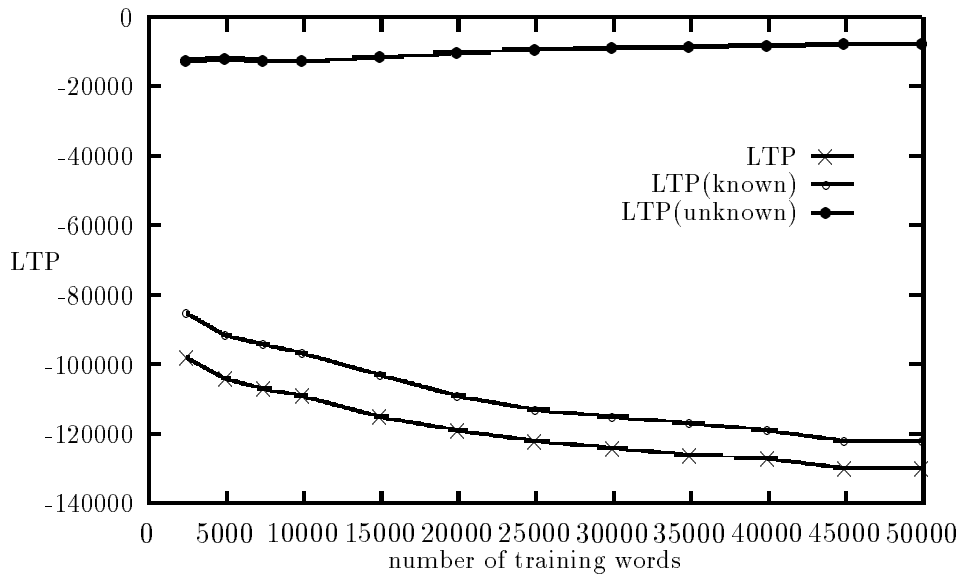

Figure 4.1: The graphs for LTP, LTP(known) and LTP(unknown)

$LTP(known)$ separately:

$$LTP = \sum_{i,s.t.w[i] \in V} log_2(P(w[i] = w_{l(i)}|w[i-1])) + \sum_{i,s.t.w[i] \notin V} log_2(P(w[i] = w_{l(i)}|w[i-1]))$$
$$= LTP(known) + LTP(unknown)$$

Both are also shown in Figure 4.1. $LTP(unknown)$ increases and $LTP(known)$ decreases as the training text grows. We explain this as follows. As the training size increases, more and more words that are unknown when a small training text is used, become known. Each of these words that is unknown and receives the probability of $d$ when a small training text is used, will receive the probability value according to the bi-pos formula when a larger training text is used. It turns out that $LTP(known)$ decreases more than $LTP(unknown)$ increases. This happens because the probability $d$ our model assigns to unknown words is higher than the average probability assigned to words in the vocabulary.

This behavior is again due to the fact that models trained on different amounts of text have different underlying sample spaces. We already mentioned this problem in section 4.2.1, when we considered a model trained on zero words of text, which would have a perplexity of one. In above example, the difference in sample spaces leads to a distortion of the probability measure to the extent that models trained on less data perform 'better'.

One way to solve this problem is to adjust for some of the distortion caused by different sample spaces. As we mentioned earlier, by giving the probability of $d$ to every unknown word, the probabilities sum up to more than one. For example, if we have $r$ different unknown words in the testing text and distinguish between them, the sum of the probabilities of the possible words will add up to $1 + (r-1)*d$, $1-d$ for the words in the vocabulary, and an extra $r*d$ for all the unknown words. Once the testing has been completed, we can adjust for this in the following way. Suppose a language model is tested on a text that contains $s$ occurrences of these $r$ different unknown words. If we suppose a uniform distribution of the unknown words, for example, as a rough approximation, we can divide the probabilities of the unknown words by $r$. We call this adjustment the adjusted logarithm of the total probability, $ALTP$, and similarly, the adjusted perplexity, $APP$:

$$ALTP \quad = \quad ALTP(known) + ALTP(unknown) \tag{4.1}$$

$$= \quad \sum_{i, s.t. w[i] \in V} log_2(p(w[i] = w_{l(i)} | w[i-1])) \tag{4.2}$$

$$+ \quad \sum_{i, s.t. w[i] \notin V} log_2(\frac{p(w[i] = w_{l(i)} | w[i-1])}{r}) \tag{4.3}$$

$$= \quad LTP(known) + LTP(unknown) + s * log_2(\frac{1}{r}) \tag{4.4}$$

$$= \quad LTP - s * log_2(r) \tag{4.5}$$

$$APP \quad = \quad (2^{ALTP})^{-1/n} \tag{4.6}$$

This will ensure that the probabilities sum up to one, but it will not change the fact that $d$ is 'allocated' to all unknown words. In order to calculate $APP$, we just calculate $LTP$ as before and keep a counter for $s$ and $r$. When we reach the end of the testing text, we calculate $ALTP$ and $APP$ from $LTP$ by adding the factor $-s * log_2(r)$ (see equation 4.5).

Figure 4.2 shows the adjusted logarithm of the total probability, $ALTP$, and its decomposition into $ALTP(known)$ and $ALTP(unknown)$. We can see that the $ALTP$ increases as the training data increases. The difference $LTP - ALTP$ exactly quantifies how much a language model is 'cheating' by allocating $d$ to every unknown word.

Let us summarize the advantages and disadvantages of the adjusted perplexity $APP$. First, we can approximately compare language models with different vocabularies. The $APP$ ensures that the models agree in parts of their sample spaces, namely, that the models distinguish between all the words that occur in the testing text. Second, from a practical point of view, we can write a language model without having to know the testing text, which should make it easier to run a language model on any testing text and to compare existing language models. Third, we can now quantify by how much a model is 'cheating' due to its modeling of unknown words. One disadvantage of this method is that it can not be used to adjust the probabilities of words as one goes through the testing text (e.g. during speech recognition). The adjustment can only be done once the complete testing text has been seen. However, it is sufficient if one wants to approximately compare two language models with different vocabularies. Another disadvantage is that models trained on different texts can still differ in their underlying sample spaces. The $APP$ measure only ensures that they distinguish between all the words in the testing text, but can of course not ensure that their sample spaces are identical. In other words, the two models can still try to solve different tasks.

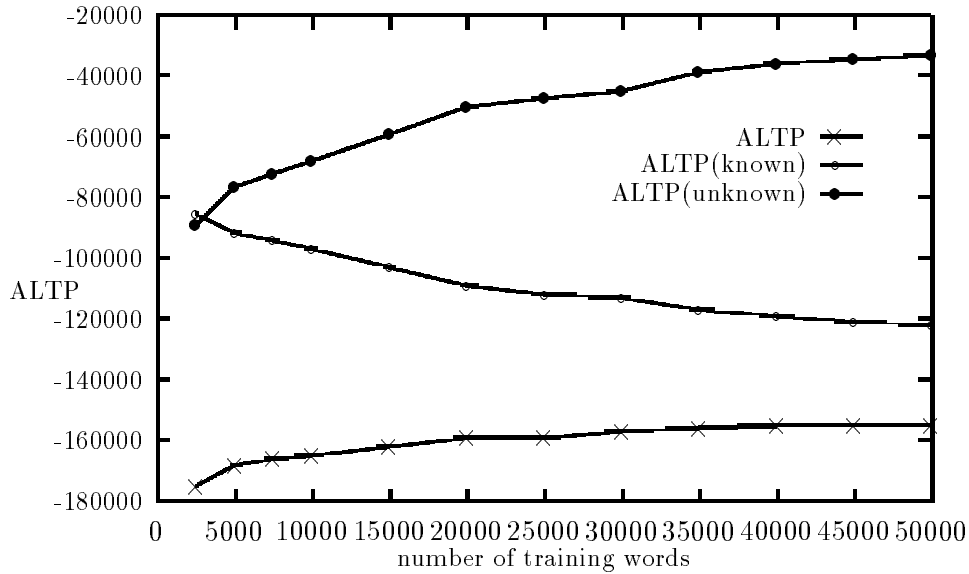

Figure 4.2: The graphs for ALTP, ALTP(known) and ALTP(unknown)

Another way of solving the problem of different sample spaces is to fix the vocabulary independently of the training text. This ensures that the underlying sample spaces are identical and that the perplexity measure is used to evaluate two models that are trying to solve the same task. In order to fix the vocabulary independently of the training text, we need to modify our model slightly. If the vocabulary is fixed in advance, it may contain words that were never seen in the training text and we will call these words *unseen*. What probabilities should we assign to *unseen* words? We change the model so that it gives a small, arbitrarily chosen value $d_1$ to every unseen word, which leads to the following model:

$$p(w[i] = w_l | w[i-1])$$
$$= \begin{cases} (1 - u * d_1 - d_2) \sum_{g_j \in G} [(c_1 * f(g(w[i]) = g_j | g(w[i-1]))) + c_2) \\ * f(w[i] = w_l | g(w[i]) = g_j)) & \text{if } w_l \in V \text{ and } w_l \text{ was seen} \\ d_1 & \text{if } w_l \in V \text{ but } w_l \text{ unseen} \\ d_2 & \text{if } w_l \notin V \end{cases}$$

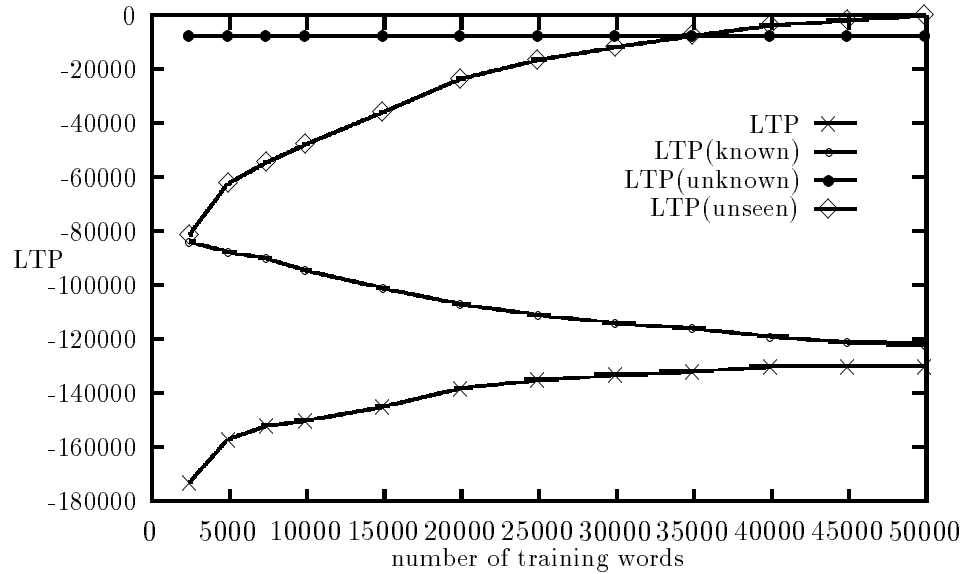

Figure 4.3: The graphs for LTP, LTP(known), LTP(unknown) and LTP(unseen) with fixed vocabulary

All the constants that are part of the standard bi-pos model, for example $c_1, c_2$ and $d_2$ (which corresponds to the former $d$), are determined as mentioned in section 4.1.2. For the new constant $d_1$ corresponding to the probability of *unseen* words we arbitrarily chose $d_1 = 10^{-6}$. The constant $u$ is set to the number of unseen words in the vocabulary and the second term $u * d_1$ is necessary to ensure that the probabilities of all words sum up to one.

Figure 4.3 shows the $LTP$ of the modified model. We can see that the model improves as the size of the training text increases. The modified model therefore conforms to our intuition. Moreover, since the vocabulary is fixed in advance, the $LTP$ caused by unknown words does not change when the training text is changed. However, we can also see that $LTP$ has $LTP(unseen)$, the entropy caused by unseen words, as an additional component. As the the size of the training text increases, more and more words that were previously *unseen* and received the probability of $d_1$ become part of the vocabulary and receive a probability according to the bi-pos formula. This explains why $LTP(unseen)$ increases and $LTP(known)$ decreases as the training text gets bigger. Since $LTP(unseen)$ increases more than $LTP(known)$ decreases, the overall effect on $LTP$ is an increase, which corresponds to an improvement in the model.

Summing up, we have now seen two ways of dealing with differences in sample spaces caused by different amounts of training data. We can alleviate the problem of different sample spaces by using the adjusted perplexity measure or we can avoid the problem of different sample spaces entirely by fixing the vocabulary independently of the training text. If it is not possible to agree on a common vocabulary (e.g. because different researchers working in different locations do not agree), the $APP$ is a flexible way of making the results comparable, without ensuring identical sample spaces. However, from a theoretical point of view, fixing the vocabulary in advance is more satisfying than using the $APP$. First, it ensures that the sample spaces are identical for all models. Second, it makes sense that the sample space should be fixed before one starts to compare probabilities. After all, if one wants to compare probabilities taken from probability distributions, the distributions should be constructed over the same sample space. Because it is preferable from a theoretical point of view to fix the vocabulary in advance and because we have no problem in agreeing on a common vocabulary, we will fix the vocabulary based on the training text in the rest of this thesis.

### 4.2.3 Influence of the Amount of Training Data on the Performance of our Language Model

As mentioned in the beginning of section 4.2, we wanted to ensure that our corpus contains enough data to train our bi-pos model. When we tried to measure the influence of the amount of training data on the performance of the model, the surprising results prompted a discussion of the underlying issue of sample spaces. Now that we have chosen to fix the vocabulary independently of the training text, we have a meaningful way to measure the influence of the amount of training data on the performance of the model. Figure 4.4 shows the perplexity of the model for different amounts of training data and for four different tagsets (as mentioned in section 4.1.1). First, we can see that the models do not continue to improve significantly when the training data is increased from 30,000 to 50,000 words. We thus assume that all the models are well trained after 50,000 words. This justifies why we only use 50,000 words of training text in the remainder of this thesis. Second, we can see that the amount of training data should influence the choice of tagsets. If, in this case, the available training data contained only 15,000 words, then a smaller tagset (e.g. the 24 tags) would lead to better results than a larger tagset and this confirms our intuition.

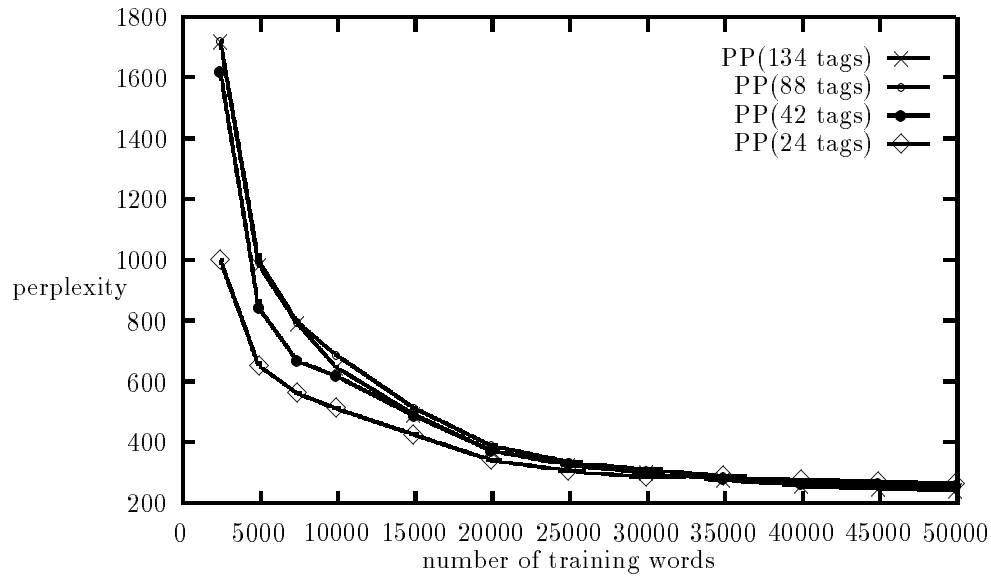

Figure 4.4: The perplexity for different amounts of training data and different tagsets

## 4.3  Weaknesses of the Bi-pos Model

In the following, we will apply the method of identifying weaknesses presented in chapter 3 to the chosen bi-pos model (see page 69).

The first weakness we will identify is the prediction of the next word in a very small number of contexts. We perform a very detailed analysis of these contexts in order to understand why they constitute a weakness. Even though we do not proceed in trying to improve this weakness, the information we uncover by analyzing weaknesses is already helpful in showing where we should concentrate future efforts.

The second weakness we will identify is the prediction of unknown words. As with the previous weakness, this is helpful for future work. Moreover, we actually try to improve this particular weakness and we develop a new modeling of unknown words. This results in a reduction of the perplexity ranging between 14% and 21%.

The third weakness we will identify is the second factor in our bi-pos formula (section 2.5.3, equation 2.23), the prediction of the next word given its class. This is important by itself, because it again identifies a weakness that needs to be improved for many different language models, even for the recently used probabilistic context free grammars. Trying to actually improve this weakness leads us to the development of a new generalized $N$-pos model that we will present at the end of this chapter.

### 4.3.1   Different Contexts

In the bi-pos model we use (see page 69), the current context consists simply of the tag of the preceding word. To recall the effect this definition of context has, suppose $w_1$ and $w_2$ are the last words of two sequences of words $\alpha$ and $\beta$. Further suppose that $w_1$ and $w_2$ have the same tag (or set of tags). The probability with which the model will expect to see a certain word next will be the same in both cases, whether the preceding sequence was $\alpha$ or $\beta$. In other words, the model distinguishes between as many contexts as it has classes or tags and it has a separate distribution for each of these contexts.

When we analyze this model with respect to its weaknesses, it seems natural to ask what fraction of the $LTP$ is caused by each context (or tag $g$). For that purpose, we group the elements of the sum in equation 3.3 , page 36, according to the preceding tag $g$ , produce a separate sum for each ($LTP_g$), and determine what fraction of the total $LTP$ each tag represents. We can thus calculate the fraction of $LTP$ caused by each context or by each preceding tag. In table 4.3, we give the ten tags that account for the biggest part of the $LTP$ when the tagset with 42 tags were used. In order to find out why these are the tags causing most of the $LTP$, we performed a more detailed analysis of the first three tags. For each tag $g$, we looked at the number $n_g$ of times $g$ occurred, the $LTP$ caused by the prediction of the next word given that the last tag was $g$ ($LTP_g$) and the average $LTP$ per word given that the last tag was $g$ ($avg_g$). The results are shown in the first three columns of table 4.4. Moreover, we used the method of probability decomposition to split up $LTP_g$ into the fraction caused by the prediction of the next tag ($f_{tag}$), the prediction of the next word given its tag ($f_{word}$) and the rest ($f_{rest}$). $f_{rest}$ contains for example the prediction of unknown words. These three values are shown in column four, five and six in table 4.4.

| Tag | Description | Fraction of $LTP$ |
|-----|-------------|-------------------|
| N | noun | 0.16 |
| AT | article | 0.13 |
| IN | preposition | 0.12 |
| V | verb | 0.08 |
| P | pronoun | 0.07 |
| NP | proper noun | 0.05 |
| , | comma | 0.05 |
| JJ | adjective | 0.05 |
| . | period | 0.05 |
| BE | forms of to be | 0.04 |

Table 4.3: The ten tags of the preceding word causing the biggest fraction of the $LTP$ when the tagset with 42 tags is used

| $g$ | $n_g$ | $LTP_g$ | $avg_g$ | $f_{tag}$ | $f_{word}$ | $f_{rest}$ |
|-----|-------|---------|---------|-----------|------------|------------|
| N | 4229 | $-2.1 * 10^4$ | -4.9 | 0.48 | 0.46 | 0.06 |
| AT | 2448 | $-1.7 * 10^4$ | -6.9 | 0.29 | 0.61 | 0.10 |
| IN | 2925 | $-1.6 * 10^4$ | -5.3 | 0.43 | 0.48 | 0.09 |

Table 4.4: A detailed analysis of the three tags causing the highest fraction of $LTP$

We can see that the tag $N$ causes the largest fraction of $LTP$ because it occurs very often. Even though it is relatively easy to predict the next word given that the last tag was $N$ ($avg_N$ is the highest), this is more than compensated by its frequency of occurrence. When predicting the next word, about the same fraction of $LTP$ is coming from the prediction of the next tag ($f_{tag}$) and the prediction of the next word given its tag ($f_{word}$).

The tag $AT$ occurs far less frequently, but predicting the next word knowing that the last tag was $AT$ is very difficult ($avg_{AT}$ is the lowest). Moreover, we can see from columns four and five that it is the prediction of the next word given its tag ($f_{word}$) that accounts for most of the $LTP$ (61%). This is because articles are often followed by so called open class words. Open class words are words like verbs or nouns which belong to a class with a very large, almost unlimited number of members. This contrasts with closed class words like articles, which belong to a class with a very small, predetermined number of members. Because the prediction of the actual word given its class is very difficult for open class words, they account for a large fraction of $LTP$ and the same is true of tags who can often be followed by open class words (because predicting the next word is very hard, if the next word belongs to an open class).

| $g$ | $f_{LTP}$ | $n_g$ | $LTP_g$ | $avg_g$ |
|-----|-----------|-------|---------|---------|
| N   | 0.24      | 517   | $-5.0 * 10^3$ | -9.7 |
| IN  | 0.24      | 1479  | $-5.0 * 103$ | -3.4 |
| V   | 0.11      | 222   | $-2.3 * 10^3$ | -9.6 |
| ,   | 0.05      | 447   | $-1.0 * 10^3$ | -2.37 |
| .   | 0.05      | 520   | $-1.0 * 10^3$ | -2.0 |

Table 4.5: The five tags causing the highest fraction of $LTP$ given that the last tag was $N$

The tag $IN$ is more similar to the tag $N$ in its behavior. It accounts for a large fraction of the $LTP$ because it occurs quite frequently (more often than the tag $AT$), but the prediction of the next word given that the last tag is $IN$ is easier than for the tag $AT$ ($avg_{IN}$ is higher).

Moreover, we can see from table 4.4 that $f_{tag}$ is highest if the last word was a noun. Why is it so difficult to predict the next tag in this particular context? In order to answer this question, we looked at the tags that follow $N$ and the five tags causing most of the $LTP$ are shown in table 4.5. For each tag $g$, we give the fraction of $LTP$ ($f_{LTP}$), given that the last tag was $N$, it causes, the number $n_g$ of times $g$ occurred after $N$, the $LTP$ caused by the prediction of the word with tag $g$ given that the last tag was $N$ ($LTP_g$) and the average $LTP$ per word given that the last tag was $N$ ($avg_g$). We can see that there is a wide range of tags that frequently follow nouns. Given a noun, it is indeed hard to predict what the next tag will be. Information, that would for example allow us to predict better which of the three most important tags will come next, would therefore be very useful in improving the model.

From the more detailed analysis based on table 4.4, we can see that the contexts causing the largest fraction of $LTP$ are the ones that occur very frequently or that are often followed by open class words.

We can also see from table 4.3, that the first four tags account for 49% of the $LTP$. In other words, about 49% of the $LTP$ is caused by a fraction of roughly $\frac{4}{42} \approx 0.10$ of the tags. In general, Figure 4.5 shows the relationship between the fraction of $LTP$ and the fraction of tags causing this fraction of $LTP$. The graph shows clearly that a small number of tags causes a large fraction of $LTP$ and that a large number of tags only causes a small fraction of $LTP$. Qualitatively, this kind of graph occurs very often in natural language processing and it is a typical example of a Zipf distribution ([158]). Other quantities having similar distributions can be found at almost all levels of language ranging from the phoneme to the sentence ([3], [4]). One can argue that the Zipfian nature of the graph shows that the tags are not well suited for a language modeling task. There is really not much point in differentiating among most of the 50% of the tags that only account for 10% of the $LTP$.

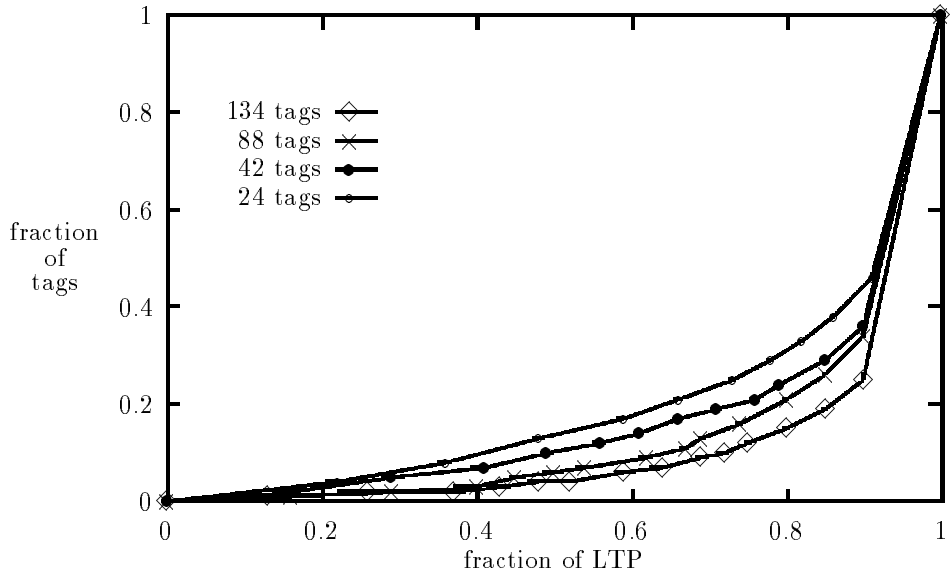

Figure 4.5: fraction of $LTP$ caused by fraction of tags

In this section, we have shown on which contexts we should concentrate our efforts to improve our bi-pos model. These contexts are the points in the text where the preceding word was a noun, an article or a preposition. The advantage of knowing these contexts is that we can now look at each one of them in turn in order to analyze *why* the prediction of the following word is so difficult (see for example 4.5) and we can then try to improve our model on this specific point. It is clear that solving such a specific problem is much easier than trying to somehow improve the language model in general. However, we will not explore this issue any further at this point, mainly because we have already shown the usefulness of identifying weaknesses by showing how it reduces the size of the problem at hand. Moreover, the additional understanding of the model obtained from this detailed analysis also shows the usefulness of the central idea of this thesis, the identification and analysis of weaknesses of language models.

| unknown word model | total $LTP$ | $LTP$(unknown) | $LTP$(unknown)$/LTP$ |
|:---:|:---:|:---:|:---:|
| M1 | -1.30e+05 | -7.78e+03 | 0.0599 |
| M4 | -2.51e+05 | -1.29e+05 | 0.514 |

Table 4.6: $LTP$ caused by unknown words when model M1 and M4 are used

### 4.3.2 Unknown Words

We will now move to the second weakness, the prediction of unknown words.

On page 69, we derive the formula of the bi-pos model we use. Our model gives a constant probability to unknown words and this is different from the rest of the model, which uses the bi-pos formula. This prompted us to measure how much impact this separate part of the model, which is only used for unknown words, has on the overall performance.

In order to measure the impact of unknown words, we group the elements of the sum in equation 3.3, page 36 into two groups, the terms that correspond to unknown words and the rest. We calculate each sum separately and measure the fraction of the $LTP$ caused by unknown words. The result (shown in table 4.6) is that the unknown words account for approximately 6% of the total $LTP$, independent of the tagset used, since the probability given to unknown words does not depend on the number of tags. If we use a different model for the unknown words, namely the model M4 from section 4.2.1, this fraction is as high as 51%. As in the preceding section, we have now identified a specific weakness, the modeling of unknown words. We now proceed with trying to find ways of improving the model with respect to this particular problem.

The current model gives a constant probability to unknown words, independent of the current context or the most likely tag of the next word. However, it is clear that the probability of the next word being unknown depends on the hypothesized tag for the next word. For example, it is clear intuitively that if the next word is an open class word, the word is much more likely to be unknown. Hence, rather than having

$$P(w[i] = w_l | w[i-1]) = d_2 \text{ if } w_l \notin V,$$

we would now like to make $d_2$ depend on the supposed tag $g_j$ of the next word. This leads to a formula similar to the part of the model that deals with words in the vocabulary: first, we predict a likely next tag $(c_1 * f(g(w[i]) = g_j | g(w[i-1])) + c_2)$, then we predict an unknown word give this tag $(d_{g_j})$. This leads to the following formula:

$$P(w[i] = w_l | w[i-1])$$

| number of tags | old model | new model | improvement |
|:---:|:---:|:---:|:---:|
| 24 | 265 | 229 | 0.14 |
| 42 | 259 | 218 | 0.16 |
| 88 | 249 | 196 | 0.21 |
| 134 | 243 | 192 | 0.21 |

Table 4.7: The perplexity of the old and the new model

$$
= \begin{cases}
\sum_{g_j \in G}[(1 - u * d_1 - d_{g_j})(c_1 * f(g(w[i]) = g_j | g(w[i-1]))) + c_2) \\
\quad * f(w[i] = w_l | g(w[i]) = g_j)] & \text{if } w_l \in V \text{ and } w_l \text{ was seen} \\
d_1 & \text{if } w_l \in V \text{ but } w_l \text{ unseen} \\
\sum_{g_j \in G} d_{g_j} * (c_1 * f(g(w[i]) = g_j | g(w[i-1]))) + c_2) & \text{if } w_l \notin V
\end{cases}
$$

We can also explain this modeling of unknown words by saying that the unknown word is a word of the vocabulary that can appear with all tags. We can estimate the values $d_{g_j}$ of formula 4.7 from the training text using the same technique that was used to estimate $d_2$ (see page 61). We used Turing's formula again and estimated $d_{g_j}$ as the number of unique words with tag $g_j$ over the number of words with tag $g_j$. We show in Appendix C that the sum of the probabilities is still equal to one.

The perplexity of the new model is shown in Table 4.7. First, we can see that the improvement is substantial for all sets of tags, ranging between 14% and 21%. Second, the improvement increases when the number of tags increases. This is because for each tag, we have a different distribution for unknown words. As the number of tags increases, the distributions of unknown words can become more and more specific.

In this section, we identified another weakness of our bi-pos model, the modeling of unknown words. We then improved our model with respect to this particular weakness by developing a new modeling for unknown words. In this new model, the probability of seeing an unknown word depends on the context (like the prediction of vocabulary words), in particular, on the hypothesized tag for the next word. This resulted in an improvement in performance ranging between 14% and 21%, depending on the number of tags used. Finding a new modeling of unknown words, which results in a significant improvement, is an important result by itself. But equally important for us is that it shows that the identification of a weakness is, at least in this case, a first step in improving our model. Furthermore, in this concrete example, the identification of a weakness does lead to a subsequent improvement of the model. This shows the usefulness of the central idea of this thesis, the identification

| nb of tags | unknown | fact | word | pos |
|:---:|:---:|:---:|:---:|:---:|
| 24 | 0.06 | 0.01 | 0.58 | 0.35 |
| 42 | 0.06 | 0.01 | 0.53 | 0.40 |
| 88 | 0.06 | 0.01 | 0.45 | 0.48 |
| 134 | 0.06 | 0.01 | 0.43 | 0.50 |

Table 4.8: The $LTP$ caused by different components of the model

and analysis of weaknesses of language models.

### 4.3.3 Different Components

We will now move to the third weakness, the prediction of a word given its tag.

The bi-pos model from page 69 calculates the probabilities of words in a two step process. First, it calculates the probability of a tag; then, given the tag, it calculates the probability of a word. It seems natural to measure how much of the overall $LTP$ is caused by each of these components of the model.

To be more precise, using the method of probability decomposition introduced on page 51, we measure the fraction of $LTP$ caused by unknown words ($d_2$), by the factor for known words ($1 - u * d_1 - d_2$), by the term for predicting the next tag ($p(g(w[i])|g(w[i-1]))$), and by the term for predicting the next words given its tag ($p(w[i] = w_{l(i)}|g(w[i]))$). Recall from page 69 that unseen words are words of the vocabulary that do not occur in the training text. Since we fixed our vocabulary based on the 50,000 words of text used to train the model, all words of the vocabulary do appear in the training text. Hence, there are no unseen words and unseen words do not account for any fraction of $LTP$. We therefore do not have a column for unseen words in table 4.8. However, the fraction of $LTP$ each other component causes is shown in table 4.8.

A first observation is that the $LTP$ tends to shift from the word column to the pos column as the number of tags increases. This is very understandable. If we have only 24 tags, it is much easier to predict which one of them will appear next than if we have 42 tags. By the same token, if we have only 24 tags, more words will belong to each tag and it is harder to predict the word given the tag than if we use 42 tags. In general, as the number of tags increases, the prediction of the next tag will be much harder, but given the tag of the next word, it will be easier to predict the actual word.

A second observation from table 4.8 is that the prediction of the next word, given its tag, is a very important part of the bi-pos model, at least when we consider the amount of $LTP$ caused by each part. Depending on the number of tags used, it accounts for between 35% and 58% of the $LTP$. This is more than the prediction of the next part of speech when 24 and 42 tags were used. Hence, the prediction of the words given the tag is at least as important as predicting the next tag and this is a very important fact to note for future research.

The observation that the prediction of the next word, given its tag, is very important also sheds a different light on using probabilistic context free grammars for language modeling. Recently, a number of researchers have investigated the use of probabilistic context free grammars (PCFG's) or stochastic context free grammars (SCFG's) ([89], [24], [64], [91], [92], [114], [157], [128]) for language models. The use of these grammars can lead to better language modeling by improving the prediction of the next tag. However, the problem of predicting the actual word given its tag will remain. Furthermore, these grammars rely on the same mechanisms for estimating the probabilities of words, given their tags, and there is therefore no reason to believe that these grammars will do better at this prediction than standard N-pos models. Hence, trying to improve the prediction of the word given its tag is an important research area, independent of the kind of model used to predict the tag, e.g. N-pos or PCFG.

How could one go about improving this weakness, the prediction $p(w[i] = w_{l(i)}|g(w[i]) = g_j)$ of the next word given its tag? In order to answer this question, we will first take a closer look at $p(w[i] = w_{l(i)}|g(w[i]) = g_j)$ by examining which tags $g_j$ which account for most of the $LTP$ caused by this factor ($LTP_{word}$). For each tag $g$, we will look at the number $n_g$ of times $g$ is the tag of a vocabulary word [4], the $LTP$ caused by the prediction of these words given that their tag is $g$ ($LTP_g$), the average $LTP$ of these words ($avg_g$), the fraction $f_g$ of $LTP_{word}$ caused by each tag and the fraction $f_g(total)$ of the total $LTP$ caused by each tag. These numbers are given in table 4.9.

We can see that it is a lot harder to predict the word given its tag if the tag is an open class tag (low $avg_g$ for $N, V, JJ$). The only tag that is not an open class tag in the table is the tag $IN$ (preposition). Even though it is relatively easy to predict the preposition (e.g. high $avg_{IN}$), it causes a large fraction of $LTP$ because it occurs more than twice as often as for example the following tag $JJ$ (adjectives).

---

[4]We do not want unknown words to influence the analysis of how difficult it is to predict the word given the tag.

| $g$ | $n_g$ | $LTP_g$ | $avg_g$ | $f_g$ | $f_g$(total) |
|-----|-------|---------|---------|-------|--------------|
| N | 3201 | $-2.4*10^4$ | -7.4 | 0.35 | 0.18 |
| V | 1620 | $-1.1*10^4$ | -6.5 | 0.15 | 0.081 |
| IN | 2406 | $-6.1*10^3$ | -2.5 | 0.09 | 0.047 |
| JJ | 939 | $-6.0*10^3$ | -6.4 | 0.09 | 0.047 |

Table 4.9: A detailed analysis of $p(w[i] = w_{l(i)}|g(w[i]) = g_j)$ for the four tags causing the highest fraction of $LTP$

In the light of these results, how can we improve $p(w[i] = w_{l(i)}|g(w[i]) = g_j)$? The current model has only one distribution for each $g_j$. However, it seems clear from our intuition that this probability depends on context in the same manner as the prediction of $p(g(w[i])|g(w[i-1]))$. As an example, in the context "Peter talked to the NOUN", the overall frequency of nouns is not a good indicator for the likelihood of appearance of a noun in this particular context. Nouns like "money", that people usually do not talk to, are very unlikely to appear. The information useful in predicting $p(w[i] = w_{l(i)}|g(w[i]))$ could be entirely different from the information used for predicting $p(g(w[i])|g(w[i-1]))$. We will try to improve this particular weakness in the next section (section 4.4).

In this section, we used the method of probability decomposition introduced on page 51 to measure the impact of different components of our bi-pos model on the overall performance. We have shown that the prediction of the next word, given its tag is at least as important as the prediction of the next part of speech. Again, this information is helpful for future work because it identifies a weakness of the model. Moreover, this information is not only useful for improving our bi-pos model, but it also shows that the recent interest in probabilistic context free grammars as language models does not address an important part of the model at all. The prediction of the word given its tag therefore is an important area of research, even if entirely different models, e.g probabilistic context free grammars, are used to predict the next tag. These results show again the usefulness of our definition of a weakness and of the central idea of this thesis, the analysis of weaknesses of language models, in general.

## 4.4 The Generalized $N$-pos Model

In the last section (section 4.3.3), we identified the prediction of $p(w[i] = w_{l(i)}|g(w[i]))$ as one of the weaknesses. In this section, we will try to improve this weakness by introducing the generalized $N$-pos model. We will first describe the idea behind the generalized $N$-pos model in section 4.4.1, followed by preliminary experiments that show its usefulness as a framework which can incorporate many kinds of linguistic information.

### 4.4.1 Introducing the Generalized $N$-pos Model

We can introduce a more general $N$-pos model by generalizing the following two aspects of the N-gram and N-pos models. First, rather than having distributions based on the last word or the last tag, we can base these distributions on any information about the words seen so far. We will code this information by a variable $X$. For example, $X$ could stand for the last word, in which case $X$ would range over all possible words of the vocabulary. Or $X$ could stand for the state of a simple parser and we could thus make the distributions depend on that information. The second aspect that is being generalized relates to the N-pos model, in which the probability of the next word is made dependent on the hypothesized tag alone (e.g. $p(w[i] = w_{l(i)}|g(w[i]) = g_j)$). It is clear intuitively that the frequencies of nouns also vary significantly with the context. The example we already used in the previous section (section 4.3.3) is that in the context " Peter talked to the NOUN", the nouns that people usually talk to are more likely to appear. Moreover, out intuition indicates that the immediate context of the NOUN, e.g. the words "to the", do not constrain the choice of possible nouns as much as the fact that NOUN is being talked to by Peter. In other words, even though the immediately preceding word is very useful in predicting the next tag, it seems likely that the information useful in predicting the actual noun is further away from the word to predict. This is for example mentioned in [36, p.187]) and [14]. Our generalized model should therefore allow the prediction of the actual word to depend on contextual information and it should be possible for this information *to be different* from the one used to predict the next tag.

Now that we have seen the intuitive idea behind our generalized model, we will present it in more detail. The probability of a sequence of words $W$ will again be decomposed as a product of probabilities of each word (see equation 2.5):

$$p(W) = \prod_{i=1}^{i=n} p(w[i] = w_{l(i)}|w[1:i-1]).\qquad(4.7)$$

Then the probability of each word will be modeled as

$$p(w[i] = w_{l(i)}|w[1:i-1]) = p(w[i] = w_l|X_1, ..., X_{r+s})$$

$$= \sum_{g_j \in G} p(g(w[i]) = g_j | X_1, ..., X_r) * p(w[i] = w_l | g(w[i]) = g_j, X_{r+1}, ..., X_{r+s})$$

where the $X_j$'s, $1 \leq j \leq r + s$, denote variables coding some information available from the words $w_1, ..., w_{i-1}$ seen so far. It is important to verify that the resulting probabilities constitute a probability distribution. In other words, we have to ensure that

$$\sum_{w \in V} p(w[i] = w | X_1, ..., X_{r+s}) = 1. \tag{4.8}$$

At a given point in the sentence, all the $X_j$'s have a certain fixed value. If the two component probabilities are true probability distributions for all combinations of values of $X_j$ (and this is ensured if they are constructed as usual from frequency counts), we will show in Appendix D that the sum will indeed be one.

The amount of training data needed by the generalized $N$-pos model depends of course on the knowledge encoded by the different variables. We therefore can not make any general statement with respect to the amount of training data needed. But as we will see, the model can reduce to the $N$-gram model or the $N$-pos model and the amount of training data needed in these cases will be similar to the training data required by the $N$-gram or $N$-pos model.

We will now show that the generalized model reduces to N-gram and N-pos model for a particular choice of variables $X_j$. If we chose $s = 0, r = N - 1$ and $X_j = g(w_{i-j}), j = 1, ..., N - 1$, then the generalized N-pos model reduces to the standard N-pos model. As shown in Appendix E, it will also reduce to the N-gram model for $r = N - 1, X_{r+j} = X_j = w_{i-j}, j = 1, ..., N - 1$. For other choices of the variables, we obtain models that can't be constructed from N-gram or N-pos. This shows that it is a true generalization. Furthermore, some of the variables could code linguistically relevant facts extending over a longer distance in the sentence, e.g. the subject of the sentence or the fact whether the verb is transitive. It therefore is a framework that allows more general linguistic knowledge to be captured. These points show the theoretical potential of the generalized N-pos model as a framework.

However, from a practical point of view, the generalized N-pos model is only useful if we find sources of information for the variables $X_j$ that actually help improve the quality of the model significantly. What information should the variables actually capture in order to improve the quality of the model? The lack of an answer to this question corresponds to a lack of knowledge in this area. A lot of work needs to be done in order to find out what kind of information is useful for this purpose. In the next section (section 4.4.2), we present a small step in that direction by describing the information we experimented with so far.

In the next chapter (chapter 5), we discuss what knowledge might be useful for language models in general.

### 4.4.2 Using the Generalized $N$-pos Model

In the usual bi-pos model, the probability distribution for $p(w[i]|g(w[i]) = Noun)$ is constructed by counting the number of occurrences of each noun in the training text. As seen in section 2.4, an estimate for the probability of a particular noun $w$ is then obtained by dividing the number of times $w$ occurred as a noun by the total number of noun occurrences. The probability distribution obtained in this manner can then be used to calculate the probabilities of words in a testing text.

In all the experiments described here, we construct one other distribution on top of the one just mentioned. A variable $X$ with two values ("general" and "specific") is updated as we move through training or testing text. We will see later in exactly which situation we will have $X = general$ and $X = specific$. Intuitively, we will have $X = specific$ only if we are in a very specific context, e.g. if the current noun phrase was introduced with the preposition "during". In all other situations, we will have $X = general$. After having read the training text, we count how often each noun occurred when $X$ had the value "specific". We then construct a specific probability distribution by dividing the number of times each noun occurred with $X = specific$ by the total number of times we had $X = specific$. When we go through the testing text, we use the normal distribution when $X = general$, but the specific distribution when $X = specific$. We thus replace $p(w[i]|g(w[i]) = Noun)$ by $p(w[i]|g(w[i]) = Noun, X)$, where X has two possible values.

Please note that this is not the same as having two separate tags, say Noun-general and Noun-specific. Having two separate tags would also change the factor $p(g(w[i]) = g_j|g(w[i-N+1:i-1]))$, requiring many more separate distributions. Using $p(w[i]|g(w[i]) = Noun, X)$, however, allows us to have a finer distinction for the prediction of the actual word, while still preserving the same level of abstraction for the prediction of the next tag.

Before we describe what information is coded by $X$, we would like to address the issue of smoothing again. The normal noun distribution is a typical example of a Zipf distribution ([158]). A very small number of words occur very often and a very large number of words occur very rarely. As an example, we show the fraction of nouns that occur less than $x, 1 \leq x \leq 20$, times in our 50,000 words of training text in figure 4.6. We can see that more than 50% of the nouns occur only once. If we construct the specific distribution in the manner described above, many words will not have occurred with $X = specific$, and they would get a zero probability. In order to avoid that, we obtain a combined distribution $p_{comb}(w[i]|g(w[i]) = g_j)$ by smoothing the specific distribution with the normal distribution in the following manner

$$p_{comb}(w[i]||g(w[i]) = g_j) = \lambda p(w[i]|g(w[i]) = g_j) + (1 - \lambda)p(w[i]|g(w[i]) = g_j, X = specific).$$

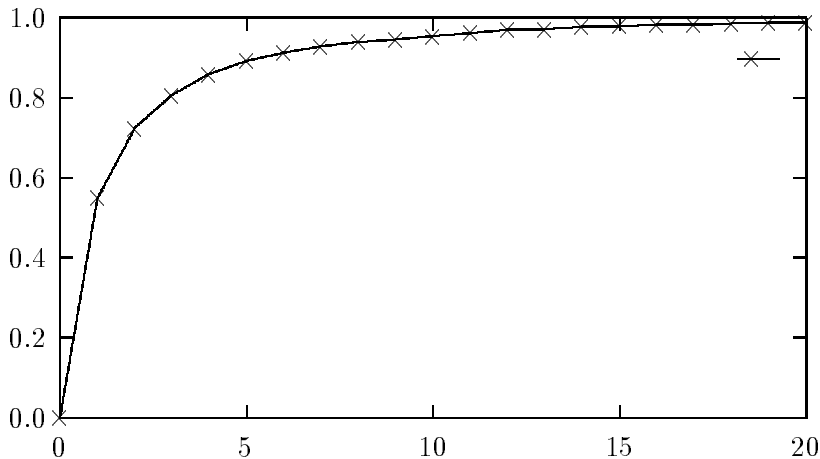

Figure 4.6: Fraction of nouns occurring less than $x$ times in text

Rather than using deleted interpolation ([66]) to determine the best value for $\lambda$, we tried $\lambda = 0.1, 0.2, ..., 0.9$ and used the value of $\lambda$ that gives the best performance on the testing text.

Experiment 1: $X$ indicates whether the noun is likely to be singular. Initially, $X$ is set to "general". If we come across "this", "a" or "an" in the text, the following noun is likely to be singular and $X$ is set to "specific". $X$ retains this value as long as we are likely to not have finished this noun phrase and is then reset to general. We did not have a parsed corpus available and used very primitive techniques to decide what constitutes a noun phrase. Roughly, we considered the noun phrase to be ongoing as long as we encounter adjectives, adverbs or nouns.

Experiment 2: $X$ was used to indicate whether the noun phrase was preceded by the preposition "during". We used the same mechanism as above to determine the boundaries of a noun phrase.

| Exp. | nb. words in S | perpl. on S | specific perpl on S | improv. on S |
|------|----------------|-------------|---------------------|--------------|
| Exp1 | 479 | 1665 | 1428 | 0.178 |
| Exp2 | 16 | 469 | 396 | 0.156 |
| Exp3 | 237 | 1365 | 1394 | -0.021 |

Table 4.10: Results on the subset S of words where X=specific

Experiment 3: With a very low probability, $X$ was randomly set to specific. This experiment was performed to see whether a random choice for the value of $X$ would lead to an improvement.

Experiments one and two make the distribution depend on information that can be several words away, depending on the length of the noun phrase. This shows that the generalized $N$-pos model is a framework that can incorporate linguistically relevant information, independent of its distance to the current word. [5]

The resulting change in perplexity on all the approximately 25,000 words of testing text is 0.031, 0.00012 and -0.0021 for the three experiments. The change is very small and this is due to the fact that the specific distribution was used only a very small number of times (e.g. $X = specific$ in the testing text). We therefore measured the improvement in perplexity only on the words that actually *used* the specific distribution (denoted with S for subset in the table 4.10). We can see that the improvement in these cases is significant, especially when compared to experiment 3 where $X$ was set randomly.

A more detailed analysis of the 479 cases of experiment 1 shows that in 195 cases (approx. 41%), the probability assigned to the word using the specific distribution was actually lower than the one from the general distribution. These were words that did not occur with $X = specific$ during training, but occurred with $X = specific$ during testing. According to the specific distribution, they would get the probability zero, but thanks to the interpolation, they get a probability between zero and the value of the general distribution.

---

[5]However, the linguistic information captured by the variables should depend on the context. If it does not, then the information will be the same in all contexts and it will therefore not change the probability distributions. As an example, take the fact that the subject is often followed by a verb. We could try to use this to increase the probabilities of verbs after we have seen a subject (assuming for the sake of this argument, that we can identify subjects). Even though one could think that this is a context independent statement, it does depend on the context in the sense used here: if we have seen a subject, then we will increase the probability of verbs; if we have not seen a subject, we won't. The fact that we will increase the probability of verbs therefore depends on the current context.

This interpolated value will always be less or equal to that of the general distribution (equal if $\lambda = 1$) and this is why we actually do worse when using the specific distribution for these words.

The critical issue in these experiments is that of capturing semantic restrictions on possible nouns based on context. How much of this can we capture by a purely distributional analysis? Can we somehow estimate a separate distribution for each $X_j$ and then combine them meaningfully so that we will not have to estimate a separate distribution for each possible combination of values of the $X_j$'s? These are important questions that we will address in the next chapter (chapter 5).

As stated in section 4.4.1, we had two reasons for introducing the more general model. One was to improve the quality of the language model, the other to capture linguistically or intuitively more satisfying information. We showed that the generalized N-pos model is a framework that can indeed incorporate linguistic information several words away and we thereby achieve the second goal. The first goal of reducing the perplexity has not been achieved on an overall level. However, the small overall improvement does not reflect the capabilities of the generalized $N$-pos model, but the usefulness of the very simple knowledge source we used. Moreover, the experiments performed show the significant reduction in perplexity in the cases where the generalized model was actually used. Further research needs to find other sources of information that are useful in reducing the perplexity in a larger number of cases and this issue will be addressed in chapter 5. These could then be added to the generalized $N$-pos model in the same manner, leading to a bigger overall improvement. Thus, the existence of knowledge that can reduce the perplexity in a large number of cases will ultimately decide on the practical usefulness of the generalized $N$-pos model. But this applies to any model that tries to incorporate more knowledge. If there is no knowledge that will reduce the perplexity in a large number of cases, then the model will not lead to a significant improvement.

## 4.5 Summary

In this chapter, we applied the central idea of this thesis, our technique of identifying weaknesses of a language model, to a commonly used bi-pos language model, reported the results, and thus showed the usefulness of performing analysis of weaknesses of language models.

We started by choosing the Lancaster-Oslo-Bergen corpus and the bi-pos model for our work and we verified that the 50,000 words of the corpus we use as training data is sufficient to train our model. The fact that the performance of our model initially decreased with an increase in training data prompted a discussion of the issue of a sample space, the set of all possible events considered by a model. We noted that it is not meaningful to compare two language models with the perplexity measure if they differ in their underlying sample spaces. Yet language models are usually compared with the perplexity measure, even though they do sometimes differ in their sample spaces due to different vocabularies or due to different ways of dealing with unknown words. One way to solve this problem is to agree on a common vocabulary. But if this is not possible, we propose to use the adjusted perplexity measure $APP$. It is a flexible way of making the results more comparable, even if the underlying sample spaces are not identical.

We then applied the idea of identifying weaknesses of a language model to our bi-pos model and reported the following results:

1) We identified three weaknesses, the prediction of words in a very small number of contexts, the prediction of unknown words, and the prediction of words given their part of speech. We thus gained considerable additional insight into the model. This insight is helpful in improving our model, but it is also relevant to other language models. The last weakness, in particular, is important with respect to the recent interest in probabilistic context free grammars as language models. Even though probabilistic context free grammars might improve the prediction of the next part of speech, they are unlikely to improve the prediction of the word given the part of speech. They therefore do not address this important weakness at all.

2) We improved one of the weaknesses, the prediction of unknown words, by introducing a new model for unknown words. This lead to an improvement in the model's performance ranging between 14% and 21%.

3) In order to address the third weakness, the prediction of words given their part of speech, we developed the generalized $N$-pos model. It can incorporate linguistic information, even if it extends over many words. Also, the information used for the prediction of the word given the part of speech in this model can be different from the information used to predict the part of speech. It is important that the model allows for this because even though the immediate context (e.g. the two or three preceding words) contains a lot of information about the part of speech of the next word, we can argue that useful semantic information that restricts the actual word may be further away.

With these results, we have shown the usefulness of our definition of weaknesses and of analyzing weaknesses of a language models in general.

# Chapter 5

# Adding Linguistic Knowledge to Language Models

In the previous chapter, one of the weaknesses of our model prompted the development of the generalized $N$-pos model as a framework that can incorporate knowledge into language models. In this chapter, we will address the issue of adding knowledge in more general terms. We begin by giving reasons for wanting to add knowledge to language models (section 5.1). We then develop criteria to select knowledge that will be useful for a language model (section 5.2). We conclude this chapter by reviewing methods for combining different types of knowledge in a language model (section 5.3).

## 5.1   Reasons for Adding Linguistic Knowledge to Language Models

Why do we want to add more knowledge to language models than current models contain? We see the following three reasons for attempting to add knowledge. First, we would like to improve the performance of existing models. Our hope is that adding knowledge will lead to improved performance. Second, adding more knowledge may well become a necessity in the future. Current speech recognizers achieve an acceptable recognition rate partly because they work in a constrained domain with a limited vocabulary. As we move to more general domains with larger vocabularies, the complexity of the recognition task and the number of acoustically confusable alternatives increases. A language model incorporating a broader range of linguistic knowledge sources could rule out many of these hypotheses, thereby helping to cope with the additional complexity. Third, adding knowledge is more satisfying than sticking to existing models on psychological grounds because humans seem to use other knowledge to predict a word than the knowledge used by current models, namely the immediately preceding two or three words. "This does not make sense!" is a reaction we often have when we look at the sentence a speech recognizer deems the most likely to be spoken. Subject and verb do not agree, prepositions are not where they should be, the verb is missing entirely or a certain combination of words is semantically incorrect. Two examples of such obviously incorrect sentences are shown below. They are taken from a recognition session of the SPHINX system (see [95, p.161,p.165]) in which a bi-gram language model is used. These utterances were from the resource management task ([118]), which uses a constrained syntax for inquiring about naval resources. For both examples, we will give the utterance spoken as well as the recognized sentence:

**Correct:** Is the economic speed of apalachicola less than that of the brunswick

**Bi-gram:** Whose economic speech of apalachicola less than that of the brunswick

**Correct:** On what day could dubuque arrive in port at his maximum sustained speed

**Bi-gram:** What would it take dubuque arrive in port at his maximum sustained speed

In these examples and in many other cases, humans seem to notice almost without effort the constraints violated by the proposed output. Adding these constraints to a speech recognizer is therefore a very natural and tempting endeavour.

Once we have decided to add linguistic knowledge to a language model, two questions come to mind. First, what knowledge should we try to add? Second, how can we combine the different types of knowledge in a language model? In the following two sections, we will deal with each question in turn. [1]

## 5.2   What Knowledge Should We Add?

In linguistics, knowledge about language can be divided into phonetics, phonology, prosody, morphology, syntax, semantics and pragmatics. The acoustic model of a speech recognizer captures some of the phonological knowledge and the part unique to speech recognition, the signal processing. Morphology, syntax, semantics and pragmatics could all be included in the language model. Therefore, from an abstract point of view, there is a very wide range

---

[1] The related question of how to obtain and to encode the knowledge is addressed in the first section.

| Type of knowledge | Task-dependent knowledge | Conversation-dependent knowledge | Speaker-dependent knowledge | Analysis-dependent knowledge |
|---|---|---|---|---|
| Pragmatic and Semantic | A priori semantic knowledge about the task domain | Concept subselection based on conversation | Psychological model of the user | Concept subselection based on partial sentence recognition |
| Syntactic | Grammar for the language | Grammar subselection based on topic | Grammar subselection based on speaker | Grammar subselection based on partial phrase recognition |
| Lexical | Size and confusability of the vocabulary | Vocabulary sub-selection based on topic | Vocabulary sub-selection and ordering based on speaker preference | Vocabulary subselection based on segmental features |
| Phonemic and phonetic | Characteristics of phones and phonemes of the language | Contextual variability in phonemic characteristics | Dialectal variations of the speaker | Phonemic subselection based on segmental features |
| Parametric and acoustic | A priori knowledge about the transducer characteristics | Adaptive noise normalization | Variations resulting from the size and shape of vocal tract | Parameter tracking based on previous parameters |

Table 5.1: Different Types of Knowledge (taken from Reddy and Newell)

of knowledge that we could incorporate into a language model. Which knowledge should we add?

For speech recognition, the knowledge has been classified in [120] according to two dimensions: the level described by the knowledge (e.g. parametric, phonemic, lexical, etc.) and its range of validity across different situations (e.g. a priori knowledge, task dependent knowledge, conversation dependent etc.). This classification is shown in table 5.1. One can see that most of the knowledge of the two lower rows (e.g. parametric and phonemic) is captured, at least partially, by the acoustic model (e.g hidden Markov model). But all the other types of knowledge could potentially be useful for a language model. Which knowledge should we try to add?

To address this question, we will first present possible criteria for selecting useful knowledge (section 5.2.1). Then, to give a structure to the space of possibly useful knowledge, we will propose a classification of possibly useful knowledge (section 5.2.2). Finally, we will show for a concrete example how the criteria from section 5.2.1 can be used to decide about its usefulness (section 5.2.3).

## 5.2.1  Criteria for Selecting Useful Knowledge

Rather than giving a necessarily incomplete list of useful knowledge, we will discuss some criteria that we think can be used to identify useful knowledge. For each criterion, we also suggest how we can evaluate a given type of knowledge with respect to the criterion:

1) Restrictions of Possible Choices. The knowledge must frequently be able to restrict the choice of words in a sentence. If this is not the case, it will not improve the quality of the language model, at least not with respect to the standard measure used to evaluate language models (see section 3.2.1). In order to find out whether a given type of knowledge restricts the possible choices of words, we can for example use introspection. Do humans often seem to use the knowledge to restrict the choice of words Does the knowledge create strong expectations about the words to come? However, we have to be careful in using introspection. A point often made by Jelinek and others (see for example [85]) is that intuitive judgments have often been misleading in the area of language models. It is always important to verify these intuitions with real data and to have the parameters of the model be trained rather than determined by hand. We can formalize the restriction of possible choices using information theory. This is for example done in [122] to define the strength of the selection restriction between predicates and arguments in terms of relative entropy. We can use the same method to measure the extent to which a given type of knowledge restricts a word that occurs later.

Let $X$ denote a random variable encoding the knowledge under investigation and ranging over the set $\{x_1, ..., x_m\}$. $X$ can for example encode the fact that the subject of the current sentence is animate or not. Let $Y$ denote another random variable encoding the identity of the word that is being restricted by $X$ and ranging over the set $\{y_1, ..., y_l\}$. $Y$ can for example encode the identity of the verb in the sentence, in which case the set of possible values would be the set of all possible verbs. We can now measure the restriction $X$ imposes on $Y$ by looking at the difference between the *prior* distribution $p(Y)$ and the *posterior* distribution $p(Y|x_i)$. An appropriate way of measuring this difference is to use the relative entropy $D(p(Y|x_i), p(Y))$, which is defined as

$$D(p(Y|x_i), p(Y)) = \sum_{y_j} p(y_j|x_i) log \frac{p(y_j|x_i)}{p(y_j)}. \tag{5.1}$$

If we rewrite equation 5.1 as

$$D(p(Y|x_i), p(Y)) = \sum_{y_j} p(y_j|x_i)[log \frac{1}{p(y_j)} - log \frac{1}{p(y_j|x_i)}], \tag{5.2}$$

we can see that the relative entropy measures in fact the average difference in information (see section 3.2.2) provided by the two distributions $p(Y)$ and $p(Y|x_i)$. In fact, the relative entropy is a measure for the amount of information provided about the random variable $Y$ (the word that will be restricted) by an event $X = x_i$ (the observation of one value for a type of knowledge). As mentioned in [122, p.58], it is discussed in [139] why this measure is the appropriate one to use in this case.

Given the measure of relative entropy, we can now quantify how useful a certain type of knowledge $X$ is for the prediction of some words $Y$. [2] [3]

2) Error Analysis. There is not much point in adding knowledge that can improve the prediction of words that almost never occur or that only account for a tiny fraction of the overall performance measure. In order to find out whether a given type of knowledge has a significant impact on the overall quality, we can perform the following steps. First identify the word whose prediction will be improved by the knowledge. Second, use the technique of error analysis to determine the percentage of the $LTP$ (see section 3.3) these words cause. If this fraction is not significant then there is not much point in adding this knowledge. This point is separate from criterion 1) for the following reason. Even if a type of knowledge $X$ contains a lot of information about $Y$, the overall effect this has on the performance of the language model may be insignificant (see for example our experiments with the generalized $N$-pos model in section 4.4.2). For example, the gender of the subject and object in the previous sentence may have a significant impact on the choice of pronouns in the subsequent sentence (e.g. he, she, it). However, if the $LTP$ caused by all the pronouns is very small then improving the prediction will not lead to a significant overall improvement.

3) Computational Effort. The language models discussed in this work (see section 2.3) are used together with the acoustic model to narrow down the search space. Therefore, many thousands of hypotheses need to be evaluated in a very short time. This places severe computational restrictions on the kind of analysis that can be used. For example, given the time required for parsing unrestricted English sentences using the current technology, it seems unlikely that the language model could use a full parser for unrestricted text. However, it is possible that future work, for example in the area of neural networks or partial parsing, might change this situation (see also p. 104). In order to determine the computational effort required for a given type of knowledge, we can use the standard techniques of analyzing the computational complexity of algorithms (see for example [53, chapter 12]), such as the theoretical worst time complexity of an algorithm. Moreover, if the given type of knowledge is also used in other areas (e.g. natural language processing), we can also use the actual time and space requirements of an algorithm as it is reported in the literature.

---

[2]It would be very interesting to measure, for example, whether the amount of information a grammar provides about the part of speech of the next word is significantly higher than the amount of information provided by the immediately preceding parts of speech. This could help to explain why $N$-pos or $N$-gram models are so very useful for language models and whether a parser has the potential to improve on this.

[3]In order to decide the usefulness of a given type of knowledge, we can also measure the performance of the distribution $p(Y|x_i)$ on a testing text. The difference in the $PP$ or $LTP$ between the distribution $p(Y)$ and $p(Y|x_i)$ will be a quantitative measure for how useful the knowledge $X$ is. However, this means that we actually have to implement the knowledge $X$, but we are looking for criteria to select useful knowledge sources before implementing the more promising ones.

4) Knowledge Acquisition and Coding. For a general purpose language model, it is important that the knowledge can be acquired and coded for the use with unrestricted text. It may be possible to describe some knowledge (e.g. syntax) in terms of rules acquired by hand. For others (e.g. semantics and pragmatics), this might be very difficult.

As an example, consider the restrictions, often called selectional restrictions, a verb imposes on its direct objects, e.g. 'to drink X'. One way of capturing this would be to organize objects in a hierarchy of types. We could imagine a type corresponding to all drinkable objects and we could then have the restriction that the direct object of 'drink' belongs to that type. As pointed out in [19], there are two main problems with respect to the task at hand. First, these type hierarchies are "large, complicated and expensive to acquire by hand". Moreover, attempts at acquiring them automatically have been only partially successful. "Yet without a comprehensive hierarchy, it is difficult to use such classifications in the processing of unrestricted text". Second, even if we had these type hierarchies, this would not be sufficient to predict patterns of usage in many cases. Even though peanuts and potatoes may be very similar and therefore quite close to each other in the hierarchy (both are edible foods that grow underground), one typically 'bakes potatoes' and 'roasts peanuts'. A distribution-based analysis could capture these differences automatically and "promises to do better at least on these two problems".

On the other hand, such a hierarchy can allow generalizations that may be hard to describe with a distributional analysis. In the above example, we might be able to discover from a few data points (e.g. drink cola, drink beer, ...) that all direct objects of 'to drink' belong to the class of liquids or drinkables. We can then generalize the fact that the words actually seen are likely objects of 'to drink' for all elements of this class. With a distributional analysis, many data points would be needed for each element of that class in order to obtain the same effect. This is because it can't perform generalizations based on semantically meaningful classes.

To find out whether a particular type of knowledge can be acquired and used with unrestricted text, we can either look through the existing literature or try to decide on the issue ourselves.

### 5.2.2 Classification of Possibly Useful Knowledge

In the previous section, we saw different criteria for judging the usefulness of a particular type of knowledge. But to which types of knowledge are we going to apply these criteria? To all the potential sources of knowledge, e.g. morphology, syntax etc., mentioned on page 90? To help answer this question, we will construct a classification of all possible sources of knowledge in this section. Given this classification, we can then keep track of where the different types of knowledge fall in the classification, which parts of the classification have been tried already and we can construct a mental image of the space of possible types of knowledge.

According to which measure are we going to classify our possible types of knowledge? The most important criterion that we will use to select useful knowledge is the first one given in section 5.2.1, the restriction of possible choices. If a type of knowledge is not restricting the choice of possible words, it is not going to be useful for our task. Any knowledge that we would want to consider therefore must have the property of restricting the choice of words. Thus, we will classify all possible types of knowledge with respect to this property, namely according to the distance between the origin of the knowledge and the word it restricts. This guarantees that our classification will contain all possibly useful types of knowledge. Moreover, we also find it intuitively appealing. We will now give the resulting classification. For each class, we will describe an example of knowledge falling into the class. Furthermore, we will mention whether this class of knowledge has been used for language modeling:

1) Knowledge whose constraints do not extend across sentences. One example is grammatical knowledge. Current language models almost exclusively use knowledge from this class. Most often, the restriction of grammaticality is approximated by the constraints provided by words from the immediate context, e.g. the two or three preceding words. Judging from the success of techniques using this immediate context , this must provide quite powerful constraints, especially for fixed word order languages like English.

2) Knowledge whose constraints extend across sentences but remain within paragraphs. One example is the knowledge of the gender of a noun phrase. For example, the noun phrase "Mr. Baker" can not be referred to as "she" in a subsequent sentence. Recently used dynamic language models (see section 2.5.5) capture the fact that words are more likely to appear again if they appeared before in the current paragraph.

3) Knowledge whose constraints extend across paragraphs but not across documents. One example is the knowledge of the topic or content of the current paragraph. With respect to uses for language models, the remark of the previous class also applies here.

4) Knowledge whose constraints extend across texts. One example is the knowledge of a certain kind of vocabulary or style of writing. It has been shown (see for example [105], [68], [69]), that the language used in different genres is quite different and this is actually used for a language model in [79].

### 5.2.3 The Usefulness of Collocational Constraints

In the following, we will apply the above five criteria to identify useful knowledge from section 5.2.1 to the knowledge about collocational constraints. We will use the term collocation "quite broadly to include constraints on SVO (subject verb object) triples, phrasal verbs, compound noun phrases, and psycholinguistic notions of words association (e.g. doctor/nurse)" as suggested in [19, p.75]. We begin by reviewing work that suggests that collocational constraints are very frequent and important in language. This is taken as evidence that criteria one and two (see section 5.2.1) are partly satisfied. However, it does not replace a study that actually measures this effect quantitatively as suggested in section 5.2.1.

In [76], G. Kjellmer classifies combinations of words according to how much variability they allow. The spectrum ranges from fossilized phrases (Anno Domini, aurora borealis) to semi-fossilized phrases (by and by, by and large, Achilles heel, Achilles tendon) or idioms (have a weak/soft spot for, do badly/well for) to finally variable phrases (glass of water, go to college, his approach to, to be appointed by). "So anyone who can be said to be proficient in a language has command of a great number of set phrases as well as skill in producing acceptable variants within the limits drawn up by the selectional rules" ([76, p.126]).

In [77], words are analyzed with respect to their tendency to form clusters. "There is a continuum in English words (including names), from those whose contextual company is entirely predictable (Angeles, Fidel) to those whose contextual company is entirely unpredictable (therefore), but the evidence indicates that most of the words are to be found towards the Angeles end of the scale" ([77, p.172]). [4]

In [2], a study of the phraseology of spoken English is motivated and presented. "The native speaker's ability to speak fluently and idiomatically can thus be ascribed to his command of a great number of such preassembled phrases. This means that linguistic competence must include a large and important phraseological component ... which acts as an elastic link between the lexicon and the productive rules of grammar" ([2, p.3]).

In [136], the nature of lexical items and their relation to grammar is examined, and two principles of interpretations are stated in order to explain how meaning arises from text:

1) Principle of Choice. At each completion of a unit, a choice opens up and the only constraint is grammaticality.

2) Idiom Principle. A language user has a large number of semi-preconstructed phrases that constitute a single choice, even though it involves more words.

It is argued that the second principle is at least as important as the first, and one of the reasons for this is the following. It is noted that the more frequent a word is, the less well defined its meaning becomes. For the most frequent words, we are in fact talking about usages rather than meanings. This is called progressive delexicalization. Most normal text is made up of frequent words and of the frequent senses of less frequent words. This shows that normal text is often delexicalized and it shows the use of the idiom principle.

In [121], 'frameworks' are proposed in order to explain language patterning. It is argued that frameworks are very productive and therefore deserve closer examination. The examined frameworks are discontinuous but do not extend over more than three words and can therefore be captured in a traditional N-gram model. Nevertheless, the idea of a framework could be extended to capture constructs that extend across this local context and could therefore not be captured by a $N$-gram approach.

---

[4]Even though the immediate context can be captured well with for example a tri-gram model, it requires enormous amounts of data. Moreover, restrictions that extend over more than the two preceding words are not captured.

In [19], the usefulness of collocational constraints for natural language parsers is examined. The constraints provided by syntax as opposed to collocations are described as follows:

> "Syntactic constraints, by themselves, though are probably not very important. Any psycholinguist knows that the influence of syntax lexical retrieval is so subtle that you have to control very carefully for all the factors that really matter (e.g., word frequency, word association norms, etc.). On the other hand, collocational factors (word associations) dominate syntactic ones so much that you can easily measure the influence of word frequency and word association norms on lexical retrieval without careful controls for syntax" ([19, pp.79-80]). However, syntax may be necessary to capture stronger constraints. "We believe that syntax will ultimately be a very important source of constraint, but in a more indirect way. As we have been suggesting, the real constraints will come from word frequencies and collocational constraints, but these questions will probably need to be broken out by syntactic context. How likely is it for this noun to conjoin with that noun? Is this noun a typical subject of that verb? And so on. In this way, syntax plays a crucial role in providing the relevant representation for expressing these very important constraints, but crucially, it does not provide very much useful constraint (in the information theoretic sense) all by itself." ([19, p.80])

In [122], the notion of selectional restriction (see page 94) is formalized. This is achieved by using an information-theoretic measure and it leads to the following interpretation of selection constraints: "the strength of a predicate's selection for an argument is identified with the quantity of information it carries about that argument" ([122, p.iv]). This allows us to measure quantitatively the strength of a selectional restriction. Using a manually constructed hierarchy of words ([107], [8]), this notion of selection restriction is used to perform syntactic disambiguation of prepositional phases, coordination and nominal compounds. The information in selectional restriction must therefore be strong enough to perform this disambiguation.

Having seen evidence that collocational constraints satisfy the first two criteria of section 5.2.1, we now briefly address the third and fourth criterion.

With respect to the third criterion, computational efficiency, we note that since some of the collocational constraints require the identification of subject, verb and object, a parser seems to be needed. As pointed out in section 5.2.1, it is very unlikely that we will be able to use a full parser for the kind of language model under consideration in this thesis. However, with recent progress in the area of partial parsing (see [141], [50], [51], [26]) it seems possible to get parts of a parse with far less computational effort. These parts of the parse could be sufficient for our purpose. Alternatively, recent work in the area of neural networks (see [59]) might be extended to provide an efficient solution in the future.

The fourth criterion is the acquisition and coding of the knowledge for use with unrestricted text. Collocational constraints also pose serious problems in that respect. Progress, for example, in the automatic acquisition of subcategorization frames (see [12] and [101]) or in the availability of fully parsed corpora (see [102]), could again be sufficient to acquire and code the knowledge of collocational constraints.

In the light of the evidence presented above, we believe that collocational constraints are a good candidate to be included in a language model. However, we suggest a further investigation of the extent to which collocation constraints that can not be captured in the standard $N$-gram model, quantitatively satisfy criteria one and two of section 5.2.1.

## 5.3  How Can We Combine Useful Knowledge in a Language Model?

In the last section, we saw how we could go about selecting useful knowledge. Once we have decided on what knowledge is useful, we have to determine how to produce a probability distribution that depends on the chosen knowledge in a meaningful manner. In section 5.3.1, we will present some traditional approaches to this problem. In section 5.3.2, we will describe a very successful method that was proposed recently.

### 5.3.1  Traditional Approaches for Combining Knowledge in a Language Model

In Hearsay II ([33]), a typical example of knowledge based speech recognition, different types of knowledge are combined using a blackboard, a dynamic global data structure. Different modules, corresponding to different types of knowledge, generate hypotheses, write hypotheses onto the blackboard, and evaluate the plausibility of hypotheses found on the blackboard. This architecture is used to combine types of knowledge at different levels, e.g. phoneme, word and sentence, which are not necessarily represented in terms of probability distributions. However, in the case of the language model, we only need to combine different existing probability distributions and this does not require the complicated asynchronous architecture of a blackboard. In the following, we will present some simple mechanisms to combine different probability distributions.

We will encode the event to be predicted (e.g. the next word) and the knowledge used (e.g. the preceding word, the state of a parser) in terms of variables and their values. Let $Y$ denote the event to be predicted with a set of possible values $\{y_1, ..., y_n\}$. Following the terminology of decision trees (see section 2.5.4) and following the specific/general example of the generalized N-pos model (see section 4.4), we will denote the knowledge by variables $X_i$ with corresponding values $\{x_{i1}, ..., x_{iv_i}\}$. We will focus on the combination of two variables $X_1$ and $X_2$, but the concepts can be extended in a straight forward manner to several variables.

1) Joint distribution. For each pair $(X_1 = x_{1i}, X_2 = x_{2j})$, this methods estimates a separate distribution from frequency data.

2) Decision trees. This method estimates a separate distribution for a pair $(X_1 = x_{1i}, X_2 = x_{2j})$ only if this significantly improves the quality of the model.

3) Deleted interpolation. This method constructs a separate distribution for each pair $(X_1 = x_{1i}, X_2 = x_{2j})$, but not from frequency data of the joint distribution (as method one). Instead, it combines the distributions of each variable according to $\lambda_{ij} p(Y = y_l | X_1 = x_{1i}) + (1 - \lambda_{ij}) p(Y = y_l | X_2 = x_{2j})$. For each pair $(X_1 = x_{1i}, X_2 = x_{2j})$, the $\lambda_{ij}$ is estimated to optimize some criterion.

We will now illustrate these three methods by giving the probability distributions each method would calculate in an example. In this example, the variable $Y$ which we need to predict only has two values $Y \in \{N, R\}$. $Y$ indicates whether the next word is a noun (value $N$) or not (value $R$, $R$ standing for rest). The variable $X_1$ also has two values $X_1 \in \{Adj, R\}$. $X_1$ indicates whether the previous word was an adjective (value $Adj$) or not (value $R$). Similarly, the variable $X_2$ has the two values $X_2 \in \{Art, R\}$ and it indicates whether the second last word was an article (value $Art$) or not (value $R$). The values of the probability distributions in this example were calculated from real data, namely from our training text of 50,000 words. The overall distribution for $Y$, independent of the variables $X_1$ and $X_2$ is $p(Y = N) = 0.28$ and $p(Y = R) = 0.72$. In other words, 28% of the words in the text are nouns.

| $X_1$ | $X_2$ | $p(Y = N)$ | $p(Y = R)$ | distr. name |
|-------|-------|------------|------------|-------------|
| Adj | Art | 0.84 | 0.16 | $p_1$ |
| Adj | R | 0.65 | 0.35 | $p_2$ |
| R | Art | 0.26 | 0.74 | $p_3$ |
| R | R | 0.25 | 0.75 | $p_4$ |

Table 5.2: The joint distribution for $Y$ given $X_1$ and $X_2$

| $X_1$ | $p(Y = N)$ | $p(Y = R)$ |
|-------|------------|------------|
| Adj | 0.72 | 0.28 |
| R | 0.25 | 0.75 |

Table 5.3: The distribution for $Y$ given $X_1$

1) Joint distribution. This method directly samples the joint distribution and thus estimates a separate distribution for the four combination of values of $X_1$ and $X_2$. The resulting distributions are shown in table 5.2.

2) Decision trees. This method will estimate a separate distribution for a pair of values of $X_1$ and $X_2$ only if the resulting distribution is significantly different. As we can see from table 5.2, the value of $X_1$ has a bigger impact on the distribution than the value of $X_2$. Since varying $X_1$ while keeping $X_2$ fixed results in a large variation in probabilities, the decision tree method therefore estimates a separate distribution for the two different values of $X_1$. These two resulting distributions are shown in table 5.3. Furthermore, we can see from table 5.2, that the distribution for $X_1 = R$ does not change significantly for the two values of $X_2$. However, the distribution for $X_1 = Adj$ does change significantly depending on the value of $X_2$. The decision tree method could therefore split the first distribution in table 5.3 into two distributions, leading to a total of three distributions shown in table 5.4.

3) Deleted interpolation. This method combines the two distributions of each variable shown in table 5.5 to get four distributions $p'_1, p'_2, p'_3$ and $p'_4$ that are used as approximations of the four joint distributions shown in table 5.2. The interpolation is linear

| $X_1$ | $X_2$ | $p(Y = N)$ | $p(Y = R)$ |
|-------|-------|-----------|-----------|
| Adj | Art | 0.84 | 0.16 |
| Adj | R | 0.65 | 0.35 |
| R | (R or Art) | 0.25 | 0.75 |

Table 5.4: The distributions for $Y$ given $X_1$ and $X_2$ constructed by the decision tree

| $X_1$ | $p(Y = N)$ | $p(Y = R)$ | distr. name |
|-------|-----------|-----------|-------------|
| Adj | 0.72 | 0.28 | $p_{1a}$ |
| R | 0.25 | 0.75 | $p_{1b}$ |
| $X_2$ | $p(Y = N)$ | $p(Y = R)$ | distr. name |
| Art | 0.41 | 0.59 | $p_{2a}$ |
| R | 0.27 | 0.73 | $p_{2b}$ |

Table 5.5: The distribution for $Y$ given $X_1$ and for $Y$ given $X_2$

according to the four parameters $\lambda_{aa}, \lambda_{ab}, \lambda_{ba}$ and $\lambda_{bb}$:

$$p'_1 \quad = \lambda_{aa} * p_{1a} + (1 - \lambda_{aa}) * p_{2a}$$

$$p'_2 \quad = \lambda_{ab} * p_{1a} + (1 - \lambda_{ab}) * p_{2b}$$

$$p'_3 \quad = \lambda_{ba} * p_{1b} + (1 - \lambda_{ba}) * p_{2a}$$

$$p'_4 \quad = \lambda_{bb} * p_{1b} + (1 - \lambda_{bb}) * p_{2b}$$

Since value of $X_1$ has a bigger impact on the resulting distribution, the values of the $\lambda$'s will tend to be bigger than 0.5. We will use the values $\lambda_{aa} = 1, \lambda_{ab} = 0.77, \lambda_{ba} = 0.5$ and $\lambda_{bb} = 1$ because the resulting distributions shown in table 5.6 are very good approximations to the joint distributions shown in table 5.2. In fact, by comparing table 5.2 and 5.6, we can see that all approximations except for $p'_1$ are identical to the joint distribution.

We will now discuss the advantages and disadvantages of the three methods in general.

| distr. name | $p(Y = N)$ | $p(Y = R)$ |
|-------------|-----------|-----------|
| $p'_1$ | 0.72 | 0.28 |
| $p'_2$ | 0.65 | 0.35 |
| $p'_3$ | 0.26 | 0.74 |
| $p'_4$ | 0.25 | 0.75 |

Table 5.6: The approximations to the joint distributions constructed by deleted interpolation

One advantage of the joint distribution method is its simplicity. It is very easy to implement and it does not require a lot of computational resources to construct the joint distribution. Moreover, it directly samples the joint distribution. This means that the method will usually get closer to the true joint distribution than methods that try to approximate it based on the component distributions (e.g. the deleted interpolation method). One big disadvantage is the amount of data it requires. For each pair $(X_1 = x_{1i}, X_2 = x_{2j})$, it tries to estimate the entire distribution. It therefore has to estimate $|X_1| * |X_2| * |Y|$ probabilities. Especially in the case of language models, where the number of values for $Y$ is very large, this requires an enormous amount of data. Consider for example, the tri-gram language model presented in the review section (section 2.5.2). $Y$ corresponds to the next word, $X_1$ to the last word and $X_2$ to the second last word. It samples the joint distribution directly and may have around $10^{12}$ parameters to estimate. This requires many million words of training text and even then, the tri-gram estimates are combined with bi-gram or uni-gram estimates using the deleted interpolation method. Another disadvantage of the joint distribution method is that it constructs a separate distribution for all pairs, even if some of them will lead to very similar distributions.

The main advantage of the decision tree method is that it only constructs the joint distribution for a pair $(X_1 = x_{1i}, X_2 = x_{2j})$ if this will significantly improve the performance. This will usually result in fewer distributions performing about as well as the joint distribution method. For that reason, the decision tree method can easily be applied to many variables at a time. Whereas method one would not have nearly enough data to sample the overall joint distribution, this method will only construct it at the point where it results in a significant improvement for the given data. The main disadvantage is the computational complexity of the algorithm. It is harder to implement and may require an enormous amount of computation. As stated in [16], the algorithm for finding the optimal partitioning for classification and regression trees is exponential in the number of values of the variables $X$. However, a locally optimal partition can be found in time linear in the number of value for $X$ for each iteration.

One of the advantages of the deleted interpolation method is again its simplicity, making it easy to implement and fast to execute. Furthermore, it does not require a lot of data because it does not attempt to directly sample the joint distribution. Instead, for each pair $(X_1 = x_{1i}, X_2 = x_{2j})$, it approximates the joint distribution by a linear combination of the two components. For each pair, it therefore only has to estimate the one interpolation parameter. Its main disadvantage is that the combination of the two component distributions may not get very close to the joint distributions. In cases where there is enough data to sample the joint distribution directly, this may lead to poor performance.

Adding several knowledge sources to a language model is one instance of the general trend to construct richer probabilistic models of language. As pointed out in [122, p.1], this appears to be a trend in practical and theoretical work on language. One example of this trend is that the Penn Treebank is moving towards the annotation of text with predicate-argument structure, not only with surface linguistic structures (see [102]). Another is the use of probabilistic models for tagging (see for example [18], [28] and [90]), parsing (see for example [9]), and many other applications.

However, one of the main problems with richer probabilistic models is the sparseness of data. This is for example pointed out in [115, p.183]: "It is well known that a simple tabulation of frequencies of certain words participating in certain configurations, for example of frequencies of pairs of a transitive main verb and the head noun of its direct object, can not be reliably used for comparing the likelihoods of different alternative configurations. The problem is that for large enough corpora, the number of possible joint events is much large than the number of event occurrences in the corpus, so many events are seen rarely or never, making their frequency counts unreliable estimates of their probabilities".

From this perspective, it seems unlikely that we will be able to sample joint distributions for many different types of knowledge, even with ever growing corpora. We can therefore expect that approaches that do not require this sampling (like the deleted interpolation method) will find widespread use. One recently proposed method, that can integrate different types of knowledge, will be presented in the next section. A second approach, which has been shown to integrate constraints from different levels on a smaller task (see [72]), is the use of neural networks. Their application to this task could be the subject of a further study.

### 5.3.2 The Maximum Entropy Approach to Combining Knowledge in a Language Model

Trying to use different probability distributions to produce one combined distribution is a problem that appears in many tasks. The maximum entropy principle from the area of statistics (see [25], [61]) is a very general approach to this problem. Recently, this approach has been applied successfully to language modeling (see [93], [94], [126]). A summary of this work is given below.

The maximum entropy approach proposes the following two steps:

1) Rewrite the different probability estimates as constraints on the expectations of various functions, that the combined estimate has to satisfy.

2) From the set of all possible probability distributions satisfying the constraints, choose the one that has the highest entropy.

Suppose we are trying to estimate a (joint) probability function $p(X = x)$, $x = (x_1, ..., x_n)$. Using $k$ constraint functions $f_i(x)$, $1 \leq i \leq k$, we can impose $k$ constraints coming from different types of knowledge on the resulting distribution $p(X = x)$ in the following manner:

$$\sum_x p(X = x)f_i(x) = c_i, 1 \leq i \leq k.$$

As an example, if we choose

$$f_i(x) = \begin{cases} 1 & \text{if } x = x_0 \\ 0 & \text{else} \end{cases},$$

the constraint $f_1$ imposes that the value of $p(x_0)$ is $c_1$. In general, $f_i(x)$ can of course be of a different form thus allowing more complicated constraints for $p(X = x)$. Given such a set of $k$ consistent constraints, we can then use the algorithm of generalized iterative scaling ([25]) to find the $p(X = x)$ satisfying these constraints that has the highest entropy. We can guarantee that a unique function $p(X = x)$ exists and that it is of the form

$$P(X = x) = \prod_i \mu_i^{f(x)},$$

where the $\mu_i$ are constants that have to be found by the generalized iterative scaling algorithm.

For a model that combines the distribution obtained by the maximum entropy approach with a standard tri-gram model, a reduction in perplexity of about 30% compared to the standard tri-gram model was achieved.

The advantages of the maximum entropy approach are

1) The maximum entropy principle is simple, intuitively appealing, imposes all of the given constraints and does not assume anything else.

2) The maximum entropy principle is extremely general. It can be used for any conceivable linguistic or statistical phenomenon.

3) Information captured by traditional language models can be incorporated into the maximum entropy principle.

4) The generalized iterative scaling algorithm can be adapted incrementally, thus allowing the addition of new constraints or the relaxation of old ones.

5) The generalized iterative scaling algorithm is guaranteed to converge to the unique solution under assumptions that can be met in practice.

The weaknesses of the maximum entropy approach are

1) The generalized iterative scaling algorithm is computationally very expensive [5]. For the experiments described in [126], the algorithm ran in parallel on an average of 15 high performance workstations and it took three weeks to complete.

2) Even though the algorithm is guaranteed to converge, we do not have a theoretical bound on its convergence rate.

3) When we add constraints that are not satisfied by the training data – and this is sometimes desirable – the theoretical results guaranteeing existence, uniqueness and convergence of the algorithm may not hold.

Nevertheless, we can see from the results presented in [126] that this is one of the rare times a standard tri-gram model is outperformed significantly and consistently. This approach therefore holds considerable promise for future work.

## 5.4  Summary

In this chapter, we motivated the addition of knowledge to language models, developed different criteria for identifying useful knowledge, and presented methods for combining knowledge in a language model.

We began by pointing out three reasons for wanting to add knowledge to a language model. First, we would like to improve the model's performance. Second, if we apply current speech recognition technology to more complex tasks than the ones tackled today, the number of acoustically confusable hypotheses will increase, and we will need a better language model in order to deal with this ambiguity. Third, adding knowledge is more satisfying than sticking to existing models on psychological grounds because humans seem to use knowledge to predict a word other than the knowledge used in current models, namely the immediately preceding two or three words. Hence, there is clearly a need for a language model which incorporates more linguistic knowledge.

---

[5]Even though only the training part, which can be done off line, is expensive, its computational complexity may prevent the application of the algorithm to very large data sets for reasons of practicality.

Once we had decided to add knowledge to a language model, the following two questions came to mind. First, *what* knowledge should we add, and second, how can we *combine* different types of knowledge in a language model. We addressed both issues in turn.

Rather than trying to give a necessarily incomplete list of types of knowledge that we should add, we presented four criteria that we think should be used to identify useful knowledge. First, the knowledge should restrict the number of possible words, otherwise it is not going to help in solving our task. Second, it should be applicable often enough to be of statistical significance. Third, it should be possible computationally to use this knowledge in real time speech recognition. Finally, we should be able to acquire and code this knowledge for use with unrestricted text. Moreover, we developed a classification of possibly useful knowledge and applied the criteria for identifying useful knowledge to one kind of knowledge, that promises to be useful to improve language models in general.

We then moved on to the issue of combining different knowledge in a language model. We presented three methods for combining knowledge and developed some of the advantages and disadvantages we see in each method. Following that, we concluded that it is very unlikely that we will have enough data to estimate distributions that depend on several knowledge sources directly, even with the availability of increasingly large corpora. Therefore, we think that methods that combine distributions from single knowledge sources in a meaningful fashion will be very useful and require further investigation. One method shown to be very useful in recent work is the maximum entropy principle and it shows great promise for future work.

# Chapter 6

# Summary of Results and Future Work

The main contribution of this thesis is the idea of applying error analysis to language models. We define what we mean by error or weakness of a language model and we show how an analysis of weaknesses is useful in improving a concrete model. Thus, in addition to the concrete results we obtained, we have shown how one can go about improving language models in general. We could therefore call this "meta language modeling". In the remainder of this chapter, we will give a more detailed summary of our results (section 6.1). This will be followed by possible extensions to our work (section 6.2).

## 6.1   Summary of Results

In this thesis, we set out to improve existing language models for speech recognition. Since it is a widely accepted fact that knowing the errors or weaknesses of any kind of model makes it easier to improve the model, we proposed to perform an analysis of the weaknesses of language models. We defined in general terms what we mean by a weakness of a language model and analyzed the weaknesses of a particular, commonly used model. This analysis led, among other things, to an improvement in the model's performance ranging between 14% and 21%. This shows, in a concrete case, the usefulness of performing an analysis of weaknesses of a language model.

In order to analyze the weaknesses of a language model, we first needed to define what we mean by a weakness of a language model. Given a part of the model, we measured its impact on the overall performance of the model in terms of the percentage of the LTP, a measure closely related to the standard perplexity measure used to evaluate the language model performance. We then defined a weakness of a model as a part of the model that has a big impact on the overall performance. Does this definition conform to the intuitions we have about the word weakness? Intuitively, a weakness should be something that needs to be improved. Given our definition, weaknesses cause a considerable fraction of the overall performance measure and this means that improving them is important for the overall performance. Our definition therefore agrees with our intuition. Furthermore, the definition is directly applicable to almost all currently used language models (except probabilistic context free grammars) and the calculations involved in identifying weaknesses are very straight forward.

In order to show the usefulness of our definition of weakness and of the analysis of weaknesses in general, we performed this analysis on a commonly used bi-pos language model. We chose a corpus and a model for our work and verified that our model is well trained with the amount of training data we use. This led to the development of the adjusted perplexity measure *APP*, which gave us a flexible way of making results of models with different sample spaces more comparable. We then analyzed the weaknesses of our model giving the following results:

1) We identified three weaknesses, the prediction of words in a very small number of contexts, the prediction of unknown words, and the prediction of words given their part of speech. We thus gained considerable additional insight into the model. This insight is helpful in improving our model, but it is also relevant to other language models. The last weakness in particular is important with respect to the recent interest in probabilistic context free grammars as language models. Even though probabilistic context free grammars might improve the prediction of the next part of speech, they are unlikely to improve the prediction of the word given the part of speech. They therefore do not address this important weakness at all.

2) We improved one of the weaknesses, the prediction of unknown words, by introducing a new modeling for unknown words. This leads to an improvement of the model's performance ranging between 14% and 21%.

3) In order to address the third weakness, the prediction of words given their part of

speech, we developed the generalized $N$-pos model. It can incorporate linguistic information, even if it extends over many words, and the information used for the prediction of the word given the part of speech in this model can be different from the information used to predict the part of speech. It is important that the model allows for this because even though the immediate context (e.g. the two or three preceding words) contains a lot of information about the part of speech of the next word, we can argue that useful semantic information that restricts the actual word may be further away.

4) Our work, in particular the generalized $N$-pos model, led us to the following questions about language models in general:

    a) what knowledge should we add to language models in order to improve their performance?

    b) how can we combine different types of knowledge in a language model?

To help answer the first question, we developed four criteria useful in deciding whether a given type of knowledge is useful. Rather than having to implement all possible types of knowledge, we can thus select the more promising ones. With respect to the second question, we presented and evaluated some existing techniques that can be used for this task.

## 6.2 Future Work

The most immediate extension to our work is to try to improve the bi-pos language model with respect to the weaknesses identified in section 4.3.1. What information would we need in order to decide on the tag that will follow a noun? Does this knowledge satisfy the criteria set out in section 5.2.1? One possibility might be to divide all the nouns into two groups (e.g. two tags) – the nouns that are often followed by other nouns and the ones that are not.

Another obvious extension is to apply the idea of analyzing weaknesses of a language model to other existing models. This can be done for $N$-gram, $N$-pos, decision tree based or cache based models. Furthermore, the idea is applicable to other languages (e.g. French, German, Japanese etc.) as well as to language models that are not based on the word level (e.g. syllable, phoneme, etc.). For each model, this can show whether their weaknesses are similar to the weaknesses of our model and where the analysis of weaknesses leads to improvement in these models. This extension should be very straight forward and we do not anticipate any difficulties.

Furthermore, we could apply the criteria for judging the usefulness of one type of knowledge to selectional restrictions, a type of knowledge identified as potentially useful for language models. We could then decide whether it is worth incorporating selectional restriction into language models and whether we can expect a significant improvement in performance by doing so. To this end, we might be able to use the same data as in [122]. Alternatively, we could use a stochastic tagger and a very primitive heuristic to identify, for example, verbs and their direct objects. This would allow us to extract the necessary data from any text, rather than being restricted to a fully parsed corpus, and this would make the whole process very flexible.

Moreover, we could perform a systematic study of the usefulness of different types of knowledge (e.g. phonological, prosodic, syntactic, semantic) for a language model. Quantitative results of such a study would be very valuable to the research community because it could help in steering its future research efforts. In order to perform this kind of study, we would require a corpus annotated with many different kinds of information. Depending on the required amount of data, this could be hard to find at the moment, but we think that it will surely become available in the future.

Besides language modeling for speech recognition, we can also apply the idea of analyzing weaknesses of probabilistic language models to other areas that use $N$-gram statistics. One example is handwriting and optical character recognition. Analyzing the weaknesses of the models and improving them afterwards could lead to an improvement in performance in these areas.

# Appendix A

# Sample Text

In this section, we give a sample of section A01 of the LOB corpus. Each item in the text is made up of two parts joined by an underscore ('_'). The first part is the word itself (for example 'a'), and the second part is the tag associated with this occurrence of the word (for example 'AT' for article). Any item that has these two parts is part of the text and needs to be predicted by the language model. These items include quotes, commas, colons, and other punctuation marks. The tags are listed and explained briefly in Appendix B. For more details on the form of the corpus see [71]).

```
A01   2 ^ *'_*' stop_VB electing_VBG life_NN peers_NNS **'_**' ._.
A01   3 ^ by_IN Trevor_NP Williams_NP ._.
A01   4     ^ a_AT move_NN to_TO stop_VB \0Mr_NPT Gaitskell_NP from_IN
A01   4 nominating_VBG any_DTI more_AP labour_NN
A01   5 life_NN peers_NNS is_BEZ to_TO be_BE made_VBN at_IN a_AT meeting_NN
A01   5 of_IN labour_NN \0MPs_NPTS tomorrow_NR ._.
A01   6     ^ \0Mr_NPT Michael_NP Foot_NP has_HVZ put_VBN down_RP a_AT
A01   6 resolution_NN on_IN the_ATI subject_NN and_CC
A01   7 he_PP3A is_BEZ to_TO be_BE backed_VBN by_IN \0Mr_NPT Will_NP
A01   7 Griffiths_NP ,_, \0MP_NPT for_IN Manchester_NP
A01   8 Exchange_NP ._.
A01   9     ^ though_CS they_PP3AS may_MD gather_VB some_DTI left-wing_JJB
A01   9 support_NN ,_, a_AT large_JJ majority_NN
A01  10 of_IN labour_NN \0MPs_NPTS are_BER likely_JJ to_TO turn_VB down_RP
A01  10 the_ATI Foot-Griffiths_NP
A01  11 resolution_NN ._.
A01  12 ^ *'_*' abolish_VB Lords_NPTS **'_**' ._.
A01  13     ^ \0Mr_NPT Foot's_NP$ line_NN will_MD be_BE that_CS as_CS labour_NN
```

```
A01  13  \0MPs_NPTS opposed_VBD the_ATI
A01  14  government_NN bill_NN which_WDTR brought_VBD life_NN peers_NNS into_IN
A01  14  existence_NN ,_, they_PP3AS should_MD
A01  15  not_XNOT now_RN put_VB forward_RB nominees_NNS ._.
A01  16   ^ he_PP3A believes_VBZ that_CS the_ATI House_NPL of_IN Lords_NPTS
A01  16  should_MD be_BE abolished_VBN and_CC that_CS
A01  17  labour_NN should_MD not_XNOT take_VB any_DTI steps_NNS which_WDTR
A01  17  would_MD appear_VB to_TO *'_*' prop_VB up_RP **'_**' an_AT
A01  18  out-dated_JJ institution_NN ._.
A01  19   ^ since_IN 1958_CD ,_, 13_CD labour_NN life_NN peers_NNS and_CC
A01  19  peeresses_NNS have_HV been_BEN created_VBN ._.
A01  20   ^ most_AP labour_NN sentiment_NN would_MD still_RB favour_VB the_ATI
A01  20  abolition_NN of_IN the_ATI
A01  21  House_NPL of_IN Lords_NPTS ,_, but_CC while_CS it_PP3 remains_VBZ
A01  21  labour_NN has_HVZ to_TO have_HV an_AT adequate_JJ
A01  22  number_NN of_IN members_NNS ._.
A01  23  ^ Africans_NNPS drop_VB rivalry_NN to_TO fight_VB Sir_NPT Roy_NP ._.
A01  24  ^ by_IN Dennis_NP Newson_NP ._.
A01  25   ^ the_ATI two_CD rival_JJB African_JNP nationalist_JJ parties_NNS
A01  25  of_IN Northern_NP Rhodesia_NP
A01  26  have_HV agreed_VBN to_TO get_VB together_RB to_TO face_VB the_ATI
A01  26  challenge_NN from_IN Sir_NPT Roy_NP
A01  27  Welensky_NP ,_, the_ATI federal_JJ Premier_NPT ._.
A01  28   ^ delegates_NNS from_IN \0Mr_NPT Kenneth_NP Kaunda's_NP$ united_JJ
A01  28  national_JJ
A01  29  independence_NN party_NN (_( 280,000_CD members_NNS )_) and_CC \0Mr_NPT
A01  29  Harry_NP Nkumbula's_NP$
A01  30  African_JNP national_JJ congress_NN (_( 400,000_CD )_) will_MD meet_VB
A01  30  in_IN London_NP today_NR to_TO
A01  31  discuss_VB a_AT common_JJ course_NN of_IN action_NN ._.
A01  32   ^ Sir_NPT Roy_NP is_BEZ violently_RB opposed_VBN to_IN Africans_NNPS
A01  32  getting_VBG an_AT elected_JJ
A01  33  majority_NN in_IN Northern_NP Rhodesia_NP ,_, but_CC the_ATI
A01  33  colonial_JJ Secretary_NPT ,_, \0Mr_NPT Iain_NP
A01  34  Macleod_NP ,_, is_BEZ insisting_VBG on_IN a_AT policy_NN of_IN
A01  34  change_NN ._.
A01  35   ^ sir_NN Roy's_NP$ united_JJ federal_JJ party_NN is_BEZ
A01  35  boycotting_VBG the_ATI London_NP talks_NNS on_IN
A01  36  the_ATI protectorate's_NN$ future_NN ._.
A01  37   ^ said_VBD \0Mr_NPT Nkumbula_NP last_AP night_NN :_: ^ *'_*'
```

```
A01  37 we_PP1AS want_VB to_TO discuss_VB what_WDT to_TO do_DO
A01  38 if_CS the_ATI British_JNP government_NN gives_VBZ in_RP to_IN Sir_NPT
A01  38 Roy_NP and_CC the_ATI talks_NNS fall_VB
A01  39 through_RP ._. ^ there_EX are_BER bound_VBN to_TO be_BE
A01  39 demonstrations_NNS ._. **'_**'
A01  40 ^ all_ABN revealed_VBN ._.
A01  41   ^ yesterday_NR Sir_NPT Roy's_NP$ chief_JJB aide_NN ,_, \0Mr_NPT
A01  41 Julius_NP Greenfield_NP ,_,
A01  42 telephoned_VBD his_PP$ chief_NN a_AT report_NN on_IN his_PP$ talks_NNS
A01  42 with_IN \0Mr_NPT Macmillan_NP at_IN
A01  43 Chequers_NP ._.
A01  44   ^ \0Mr_NPT Macleod_NP went_VBD on_RP with_IN the_ATI conference_NN
A01  44 at_IN Lancaster_NP House_NPL
A01  45 despite_IN the_ATI crisis_NN which_WDTR had_HVD blown_VBN up_RP ._. ^
A01  45 he_PP3A has_HVZ now_RN revealed_VBN his_PP$ full_JJ
A01  46 plans_NNS to_IN the_ATI Africans_NNPS and_CC liberals_NNS attending_VBG
A01  46 ._.
A01  47   ^ these_DTS plans_NNS do_DO not_XNOT give_VB the_ATI Africans_NNPS
A01  47 the_ATI overall_JJB majority_NN they_PP3AS
A01  48 are_BER seeking_VBG ._. ^ African_JNP delegates_NNS are_BER
A01  48 studying_VBG them_PP3OS today_NR ._.
A01  49   ^ the_ATI conference_NN will_MD meet_VB to_TO discuss_VB the_ATI
A01  49 function_NN of_IN a_AT proposed_JJ
A01  50 House_NPL of_IN Chiefs_NPTS ._.
```

# Appendix B

# The Four Tagsets Used in Our Experiments

The four tag sets used in our experiments are reproduced here. The first column contains the tag set as part of the LOB corpus distribution. The second, third, fourth and fifth column correspond to the tag sets with 135, 88, 42 and 24 tags each. They were mostly produced by joining tags with the same prefix and are thus dependent on the original set.

| LOB | 135 tags | 88 tags | 42 tags | 24 tags | example or explanation |
|-----|----------|---------|---------|---------|------------------------|
| ! | ! | ! | ! | PU | exclamation mark |
| &FO | &FO | &FO | &FO | F | formula |
| &FW | &FW | &FW | &FW | F | foreign word |
| ( | ) | ) | ) | BR | left bracket |
| ) | ( | ( | ( | BR | right bracket |
| ' | *' | *' | *' | BR | begin quote |
| *' | **' | **' | **' | BR | end quote |
| - | *- | *- | *- | PU | dash |
| , | , | , | , | PU | comma |
| —— | —— | —— | —— | PU | dash |
| . | . | . | . | PU | full stop |
| ... | ... | ... | ... | PU | ellipsis |
| : | : | : | : | PU | colon |
| ; | ; | ; | ; | PU | semicolon |
| ? | ? | ? | ? | PU | question mark |
| ABL | ABL | AB | AB | A | pre-qualifier (quite) |
| ABN | ABN | AB | AB | A | pre-quantifier (all) |
| ABX | ABX | AB | AB | A | double conjunction (both) |
| AP | AP | AP | AP | A | post-determiner (few) |
| AP$ | AP$ | AP | AP | A | other's |
| APS | APS | APS | AP | A | others |
| APS$ | APS$ | APS | AP | A | others' |
| AT | AT | AT | AT | AT | singular article (a) |
| ATI | ATI | AT | AT | AT | singular or plural article (the) |
| BE | BE | BE | BE | BE | be |
| BED | BED | BED | BE | BE | were |
| BEDZ | BEDZ | BEDZ | BE | BE | was |
| BEG | BEG | BEG | BE | BE | being |
| BEM | BM | BEM | BE | BE | am |
| BEN | BEN | BEN | BE | BE | been |
| BER | BER | BER | BE | BE | are |
| BEZ | BEZ | BEZ | BE | BE | is |
| CC | CC | CC | CC | C | co-ordinating conjunction (and) |
| CD | CD | CD | CD | CD | cardinal (2) |
| CD$ | CD$ | CD | CD | CD | cardinal with genitive |
| CD-CD | CD-CD | CD | CD | CD | hyphenated pair of cardinals |
| CD1 | CD1 | CD1 | CD | CD | one |
| CD1$ | CD1$ | CD1 | CD | CD | one's |
| CD1S | CD1S | CD1 | CD | CD | ones |
| CDS | CDS | CD | CD | CD | cardinal with plural (tens) |
| CS | CS | CS | CS | C | subordinating conjunction (after) |
| DO | DO | DO | DO | DO | do |
| DOD | DOD | DOD | DO | DO | did |
| DOZ | DOZ | DOZ | DO | DO | does |
| DT | DT | DT | DT | DT | singular determiner (another) |
| DT$ | DT$ | DT | DT | DT | singular determiner with genitive |
| DTI | DTI | DTI | DT | DT | singular or plural determiner (any) |
| DTS | DTS | DTS | DT | DT | plural determiner (these) |
| DTX | DTX | DTX | DT | DT | double conjunction (either) |
| EX | EX | EX | EX | EX | existential there |
| HV | HV | HV | HV | HV | have |
| HVD | HVD | HVD | HV | HV | had |
| HVG | HVG | HVG | HV | HV | having |
| HVN | HVN | HVN | HV | HV | past participle (had) |
| HVZ | HVZ | HVZ | HV | HV | has |
| IN | IN | IN | IN | IN | preposition (about) |
| JJ | JJ | JJ | JJ | JJ | adjective |
| JJB | JJB | JJB | JJ | JJ | attributive-only adjective (chief) |
| JJR | JJR | JJR | JJ | JJ | comparative adjective |
| JJT | JJT | JJT | JJ | JJ | superlative adjective |
| JNP | JNP | JJP | JJ | JJ | adjective with word-initial capital (English) |
| MD | MD | MD | MD | MD | modal auxiliary (can) |
| NC | NC | NC | NC | F | cited word |
| NN | NN | NN | N | N | singular common noun |
| NN$ | NN$ | NN | N | N | singular common noun with genitive |
| NNP | NNP | NNP | N | N | sing. c. noun w. word-initial capital (English) |
| NNP$ | NNP$ | NNP | N | N | same as above with genitive |
| NNPS | NNPS | NNP | N | N | plural common noun with word-initial capital |
| NNPS$ | NNPS$ | NNP | N | N | same as above with genitive |
| NNS | NNS | NNS | N | N | plural common noun |
| NNS$ | NNS$ | NNS | N | N | same as above with genitive |
| NNU | NNU | NNU | N | N | abbr. unit of meas. unmarked for number (hr) |
| NNUS | NNUS | NNU | N | N | abbreviated unit of measurement |
| NP | NP | NP | NP | N | singular proper noun |
| NP$ | NP$ | NP | NP | N | same as above with genitive |
| NPL | NPL | NPL | NP | N | sing. locative noun w. word-initial cap. (Abbey) |
| NPL$ | NPL$ | NPL | NP | N | same as above with genitive |
| NPLS | NPLS | NPL | NP | N | plural locative noun with word-initial capital |
| NPLS$ | NPLS$ | NPL | NP | N | same as above with genitive |
| NPS | NPS | NPS | NP | N | plural proper noun |
| NPS$ | NPS$ | NPS | NP | N | same as above with genitive |
| NPT | NPT | NPT | NP | N | sing. titular noun w. word-initial cap. (Archbishop) |
| NPT$ | NPT$ | NPT | NP | N | same as above with genitive |
| NPTS | NPTS | NPT | NP | N | plural titular noun with word-initial capital |
| NPTS$ | NPTS$ | NPT | NP | N | same as above with genitive |
| NR | NR | NR | NR | N | singular adverbial noun (January) |
| NR$ | NR$ | NR | NR | N | same as above with genitive |
| NRS | NRS | NRS | NR | N | plural adverbial noun |
| NRS$ | NRS$ | NRS | NR | N | same as above with genitive |
| OD | OD | OD | OD | OD | ordinal (1st) |
| OD$ | OD$ | OD | OD | OD | same as above with genitive |
| PN | PN | PN | P | P | nominal pronoun (anybody) |
| PN$ | PN$ | PN | P | P | same as above with genitive |
| PP$ | PP$ | PP | P | P | possessive determiner (my) |
| PP$$ | PP$$ | PP | P | P | possessive pronoun (mine) |
| PP1A | PP1A | PP1 | P | P | personal pronoun (I) |
| PP1AS | PP1AS | PP1 | P | P | personal pronoun (we) |
| PP1O | PP1O | PP1 | P | P | personal pronoun (me) |
| PP1OS | PP1OS | PP1 | P | P | personal pronoun (us) |
| PP2 | PP2 | PP | P | P | personal pronoun (you) |
| PP3 | PP3 | PP3 | P | P | personal pronoun (it) |
| PP3A | PP3A | PP3 | P | P | personal pronoun (she) |
| PP3AS | PP3AS | PP3 | P | P | personal pronoun (they) |
| PP3O | PP3O | PP3 | P | P | personal pronoun (him) |
| PP3OS | PP3OS | PP3 | P | P | personal pronoun (them) |
| PPL | PPL | PPL | P | P | singular reflexive pronoun |
| PPLS | PPLS | PPL | P | P | plural reflexive pronoun |
| QL | QL | QL | QL | QL | qualifier (as) |
| QLP | QLP | QL | QL | QL | post-qualifier (enough) |
| RB | RB | RB | R | R | adverb |
| RB$ | RB$ | RB$ | R | R | same as above with genitive |
| RBR | RBR | RBR | R | R | comparative adverb |
| RBT | RBT | RBT | R | R | superlative adverb |
| RI | RI | RI | R | R | adverb (homograph of preposition: below) |
| RN | RN | RN | R | R | nominal adverb |
| RP | RP | RP | R | R | adverbial particle (back) |
| TO | TO | TO | TO | TO | infinitival to |
| UH | UH | UH | UH | UH | interjection |
| VB | VB | VB | V | V | base form of berb |
| VBD | VBD | VBD | V | V | past tense of verb |
| VBG | VBG | VBG | V | V | present participle, gerund |
| VBN | VBN | VBN | V | V | past participle |
| VBZ | VBZ | VBZ | V | V | 3rd person singular of verb |
| WDT | WDT | WD | P | P | WH-determiner (what) |
| WDTR | WDTR | WD | P | P | WH-determiner, relative (which) |
| WP | WP | WP | P | P | WH-pronoun (who) |
| WP$ | WP$ | WP | P | P | WH-pronoun (whose) |
| WP$R | WP$R | WP | P | P | WH-pronoun, relative (whose) |
| WPA | WPA | WP | P | P | WH-pronoun (whosoever) |
| WPO | WPO | WP | P | P | WH-pronoun (whom) |
| WPOR | WPOR | WP | P | P | WH-pronoun, relative (whom) |
| WPR | WPR | WP | P | P | WH-pronoun (that) |
| WRB | WRB | WR | P | P | WH-adverb (how) |
| XNOT | XNOT | XNOT | XNOT | XNOT | not |
| ZZ | ZZ | ZZ | ZZ | F | letter of alphabet (e) |

# Appendix C

# The New Model of Unknown Words

In this appendix, we will show that the new modeling of unknown words, as proposed in section 4.3.2, ensures that the sum of probabilities of all possible words is 1.

As stated in section 4.3.2, the probability of the word $w_l$, given the tag $g(w[i-1])$ of the last word is

$$P(w[i] = w_l | w[i-1])$$

$$= \begin{cases} \sum_{g_j \in G}[(1 - u * d_1 - d_{g_j})(c_1 * f(g(w[i]) = g_j | g(w[i-1])) + c_2) \\ \quad * f(w[i] = w_l | g(w[i]) = g_j)] & \text{if } w_l \in V \text{ and } w_l \text{ was seen} \\ d_1 & \text{if } w_l \in V \text{ but } w_l \text{ unseen} \\ \sum_{g_j \in G} d_{g_j} * (c_1 * f(g(w[i]) = g_j | g(w[i-1])) + c_2) & \text{if } w_l \notin V \end{cases}$$

with $c_1 = 1 - |G| * c_2$. The following calculation shows that the sum of the probabilities of all the words in the vocabulary plus the probability of the generic unknown word 'UNKNOWN' is equal to one.

$$S = \sum_{w \text{ is unseen}} P(w) + P('UNKNOWN') + \sum_{w \in V \text{ and seen}} P(w)$$

$$= A + B + C$$

$$A = u * d_1.$$

$$B = \sum_{g_j \in G} d_{g_j} * (c_1 * f(g(w[i]) = g_j | g(w[i-1])) + c_2).$$

$$C = \sum_{w \in V \text{ and } seen} P(w[i] = w | g(w[i-1]))$$

$$= \sum_{w \in V \text{ and } seen} \sum_{g_j \in G} [(1 - u * d_1 - d_{g_j})(c_1 * f(g(w[i]) = g_j | g(w[i-1]))) + c_2)$$

$$* f(w[i] = w | g(w[i] = g_j))]$$

$$= \sum_{g_j \in G} (1 - u * d_1 - d_{g_j}) * (c_1 * f(g(w[i]) = g_j | g(w[i-1]))) + c_2)$$

$$* \sum_{w \in V \text{ and } seen} P(w[i] = w | g(w[i-1]))$$

$$= \sum_{g_j \in G} (1 - u * d_1 - d_{g_j}) * (c_1 * f(g(w[i]) = g_j | g(w[i-1]))) + c_2).$$

$B+C$

$$= \sum_{g_j \in G} d_{g_j} * (c_1 * f(g(w[i]) = g_j | g(w[i-1]))) + c_2)$$

$$+ \sum_{g_j \in G} (1 - u * d_1 - d_{g_j}) * (c_1 * f(g(w[i]) = g_j | g(w[i-1]))) + c_2)$$

$$= \sum_{g_j \in G} (1 - u * d_1) * (c_1 * f(g(w[i]) = g_j | g(w[i-1]))) + c_2)$$

$$= (1 - u * d_1) * \left( c_1 * \sum_{g_j \in G} f(g(w[i]) = g_j | g(w[i-1]))) + |G|c_2 \right)$$

$$= (1 - u * d_1) * (c_1 + |G|c_2)$$

$$= (1 - u * d_1).$$

$$S = A + B + C$$

$$= u * d_1 + (1 - u * d_1) = 1.$$

# Appendix D

# The Generalized $N$-pos Model $-$ Part I

In this appendix, we will show that the generalized $N$-pos model introduced in section 4.4.1 ensures that the sum of the probabilities of all the words is one.

As stated in section 4.4.1, the probability of the word $w_l$, given the variables $X_1, ..., X_{r+s}$ is

$$
\begin{aligned}
& p(w[i] = w_l | X_1, ..., X_{r+s}) \\
& = \sum_{g_j \in G} p(g(w[i]) = g_j | X_1, ..., X_r) * p(w[i] = w_l | g(w[i]) = g_j, X_{r+1}, ..., X_{r+s})
\end{aligned}
$$

In the following, we will use $p_1(g_j | X_r)$ as a short hand for $p(g(w[i]) = g_j | X_1, ..., X_r)$ and $p_2(w_l | g_j, X_{r+s})$ as a short hand for $p(w[i] = w_l | g(w[i]) = g_j, X_{r+1}, ..., X_{r+s})$.

We assume that $p_1(g_j | X_r)$ and $p_2(w_l | g_j, X_{r+s})$ are probability distributions for all combinations of values of variables $X_k, 1 \leq k \leq r + s$ and tags $g_j, 1 \leq j \leq t$. In other words:

$$
\sum_{j=1}^{j=t} p_1(g_j | X_r) = 1
$$

and

$$
\sum_{l=1}^{l=m} p_2(w_l | g_j, X_{r+s}) = 1.
$$

This is for example the case if $p_1(g_j | X_r)$ and $p_2(w_l | g_j, X_{r+s})$ are constructed from frequency data and it is in general a reasonable assumption.

We can now show that the sum $S$ of the probabilities of all the words in the vocabulary is one:

$$
\begin{aligned}
S & = \sum_{w_l \in V} \sum_{g_j \in G} p_1(g_j|X_r) * p_2(w_l|g_j, X_{r+s}) \\
& = \sum_{g_j \in G} p_1(g_j|X_r) \sum_{w_l \in V} p_2(w_l|g_j, X_{r+s}) \\
& = \sum_{g_j \in G} p_1(g_j|X_r) \\
& = 1.
\end{aligned}
$$

# Appendix E

# The Generalized $N$-pos Model $-$ Part II

In this appendix, we want to show that the generalized $N$-pos model reduces to the $N$-gram model. For ease of reference, we repeat the formula for the generalized model:

$$p(w[i] = w_k | X_1, ..., X_m) =$$
$$= \sum_{g_j \in G} p(g(w[i]) = g_j | X_1, ..., X_n) * p(w[i] = w_k | g(w[i]) = g_j, X_{n+1}, ..., X_{n+m}).$$

If we chose $n = N - 1, X_{n+l} = X_l = w_{i-l}, l = 1, ..., N - 1$, we obtain

$$p(w[i] = w_k | X_1, ..., X_m) =$$
$$= \sum_{g_j \in G} p(g(w[i]) = g_j | w[i - N + 1 : i - 1]) * p(w[i] = w_k | g(w[i]) = g_j, w[i - N + 1 : i - 1]).$$

If we further assume that the probabilities are estimated from frequency counts denoted by the function $f()$, we have

$$p(w[i] = w_k | X_1, ..., X_m) =$$
$$= \sum_{g_j \in G} p(g(w[i]) = g_j | w[i - N + 1 : i - 1]) * p(w[i] = w_k | g(w[i]) = g_j, w[i - N + 1 : i - 1]) =$$
$$= \sum_{g_j \in G} \frac{f(g(w[i]) = g_j, w[i - N + 1 : i - 1])}{f(w[i - N + 1 : i - 1])} * \frac{f(w[i] = w_k, g(w[i]) = g_j, w[i - N + 1 : i - 1])}{f(g(w[i]) = g_j, w[i - N + 1 : i - 1])} =$$
$$= \sum_{g_j \in G} \frac{f(w[i] = w_k, g(w[i]) = g_j, w[i - N + 1 : i - 1])}{f(w[i - N + 1 : i - 1])} =$$
$$= \frac{f(w[i] = w_k, w[i - N + 1 : i - 1])}{f(w[i - N + 1 : i - 1])} =$$
$$= p(w[i] = w_k | w[i - N + 1 : i - 1]).$$

Since the frequency counts are usually smoothed before they are used to estimate the probabilities, the generalized $N$-pos model will not be exactly the same as the $N$-gram model. However, as these calculations show, it will be an approximation of it, based on the same dependencies.